\pdfoutput=1

\documentclass[11pt]{article}

\makeatletter
\def\@fnsymbol#1{\ensuremath{\ifcase#1\or\dagger\else\@arabic{#1}\fi}}
\makeatother

\usepackage[final]{acl}

\usepackage{times}
\usepackage{latexsym}

\usepackage[T1]{fontenc}

\usepackage[utf8]{inputenc}

\usepackage{microtype}

\usepackage{inconsolata}

\usepackage{graphicx}

\usepackage{amsmath,amsfonts,bm}

\def\eqref#1{equation~\ref{#1}}

\def\1{\bm{1}}

\DeclareMathAlphabet{\mathsfit}{\encodingdefault}{\sfdefault}{m}{sl}
\SetMathAlphabet{\mathsfit}{bold}{\encodingdefault}{\sfdefault}{bx}{n}

\usepackage{hyperref}
\usepackage{url}

\usepackage{tabularx}
\usepackage{booktabs}
\usepackage{caption}
\usepackage{array}
\usepackage{longtable}
\usepackage{multirow}
\usepackage{colortbl}
\usepackage{makecell}
\usepackage{fvextra}
\usepackage{xspace}
\usepackage{xcolor}
\usepackage{tablefootnote}
\usepackage{fontawesome5}

\usepackage{pifont}
\newcommand{\cmark}{\ding{51}}
\newcommand{\xmark}{\ding{55}}

\definecolor{darkgreen}{HTML}{1E7F2A}
\newcommand{\greencmark}{\textcolor{darkgreen}{\cmark}}

\definecolor{darkred}{HTML}{DA2C43}
\newcommand{\redxmark}{\textcolor{darkred}{\xmark}}

\definecolor{rankFirst}{RGB}{220, 53, 69}   
\colorlet{rankSecond}{rankFirst!50}         
\colorlet{rankThird}{rankFirst!20}          

\newcommand{\first}[1]{\cellcolor{rankFirst}{\textcolor{white}{#1}}}
\newcommand{\second}[1]{\cellcolor{rankSecond}{#1}}
\newcommand{\third}[1]{\cellcolor{rankThird}{#1}}

\newcommand{\ModelIconSize}{1.1em}
\newcommand{\icon}[1]{%
  \raisebox{-0.15\height}{\includegraphics[height=\ModelIconSize]{icons/#1}}%
}

\DeclareRobustCommand{\anthropic}{\icon{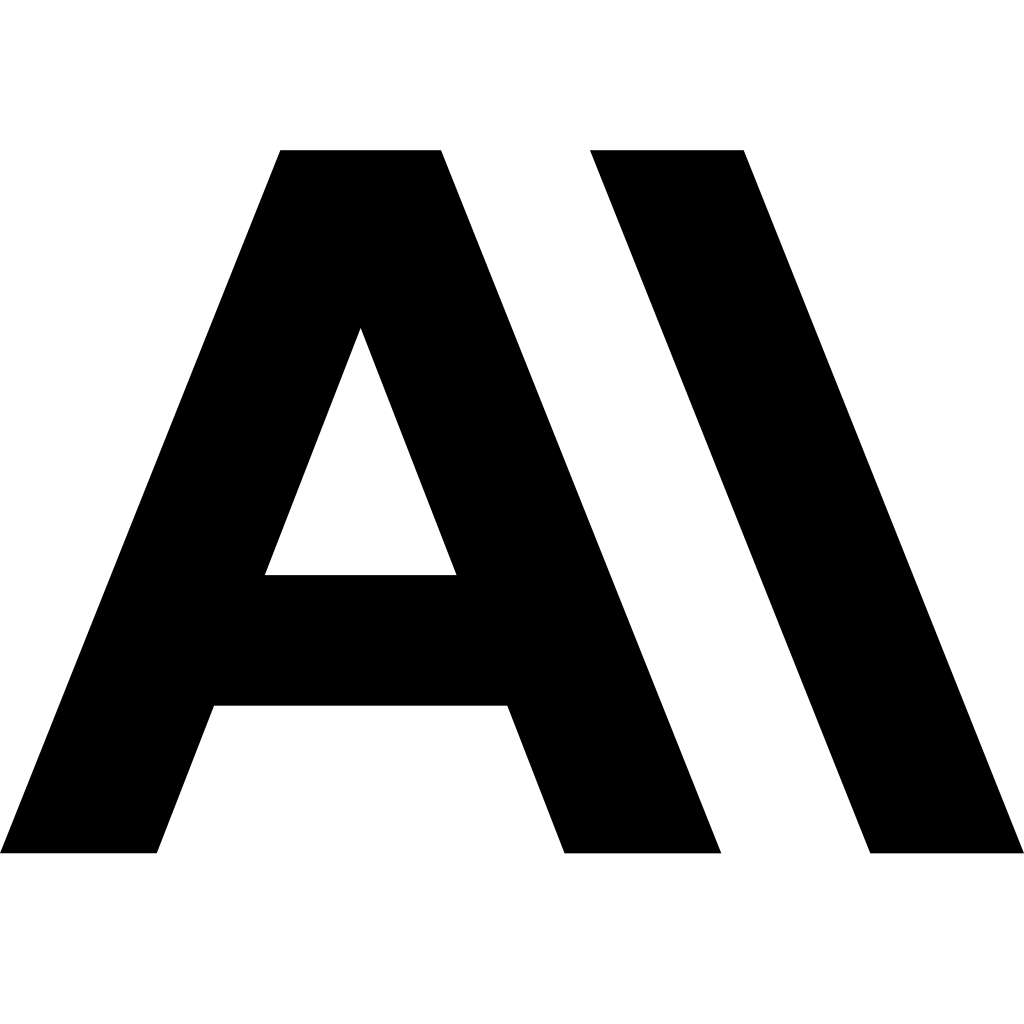}\xspace} 
\DeclareRobustCommand{\openai}{\icon{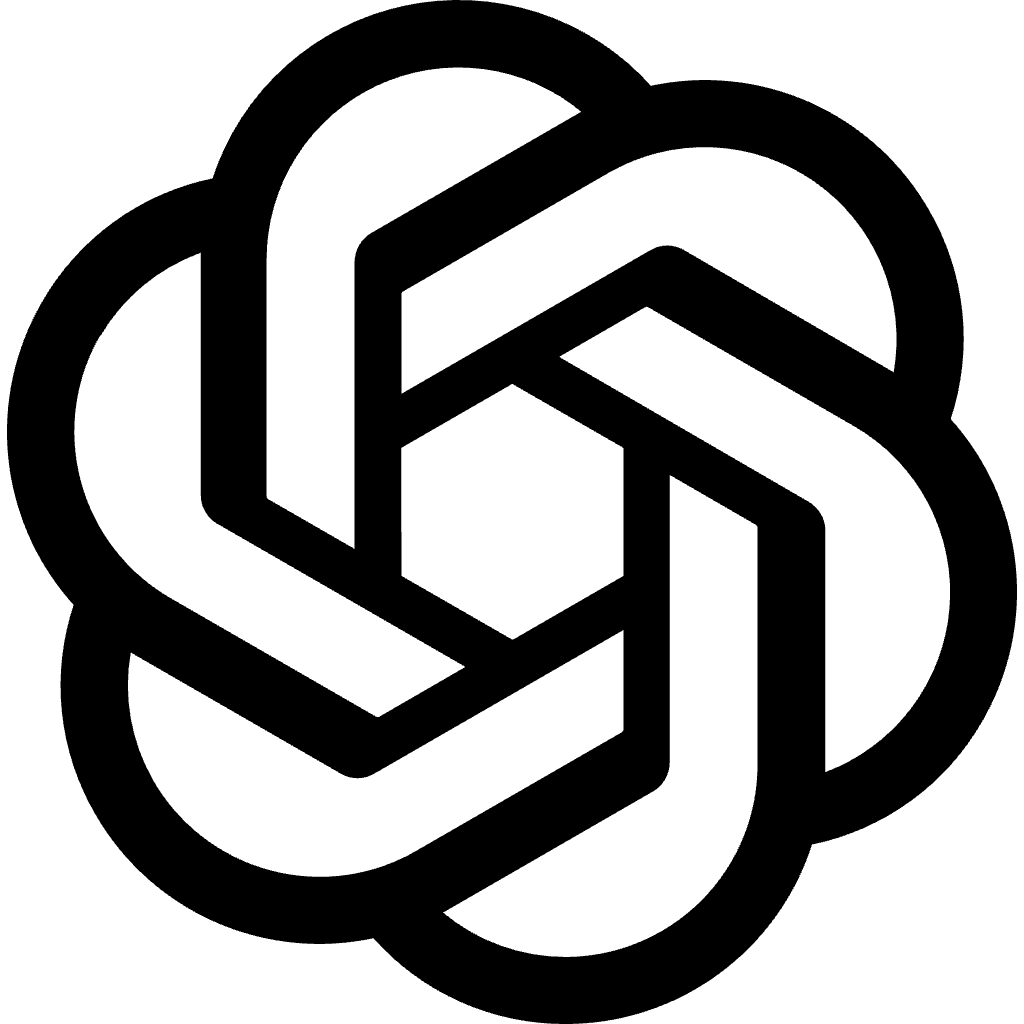}\xspace}    
\DeclareRobustCommand{\llama}{\icon{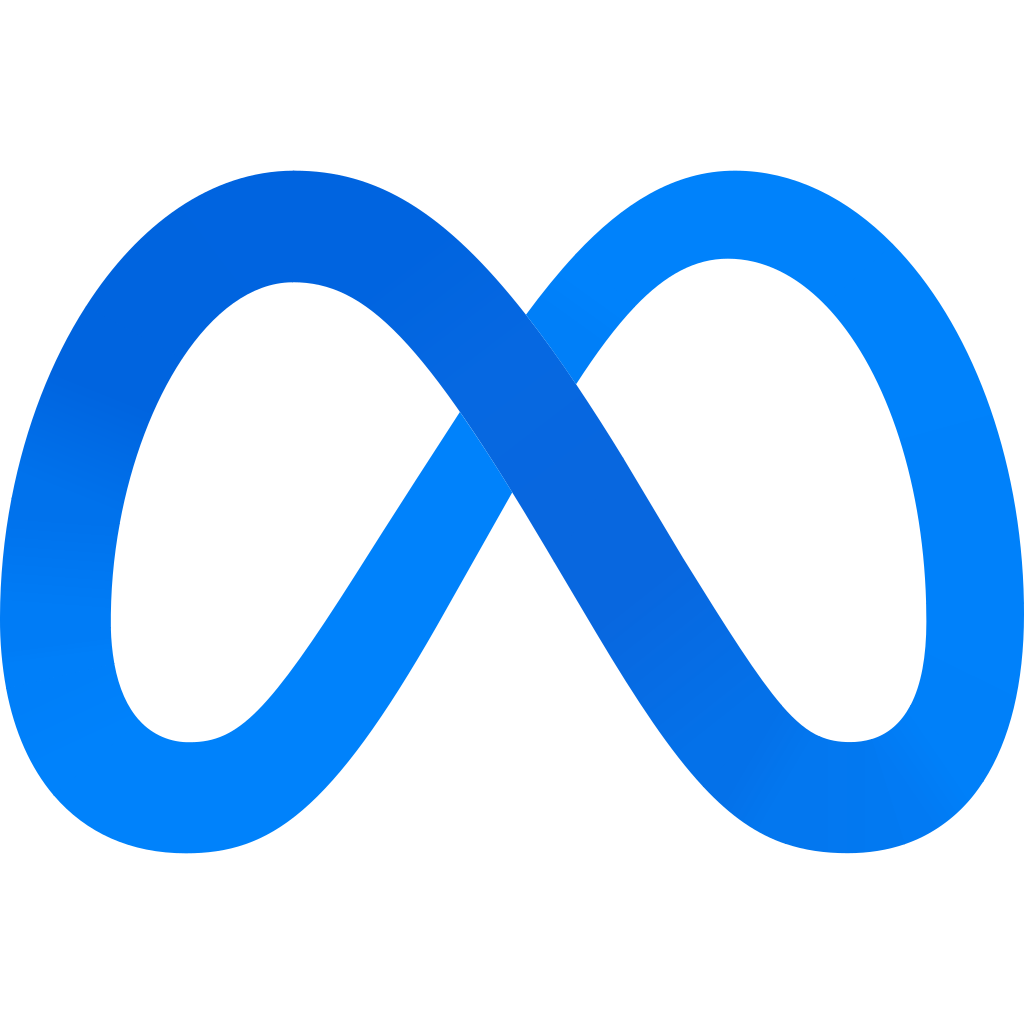}\xspace}       
\DeclareRobustCommand{\microsoft}{\icon{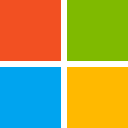}\xspace}
\DeclareRobustCommand{\gemini}{\icon{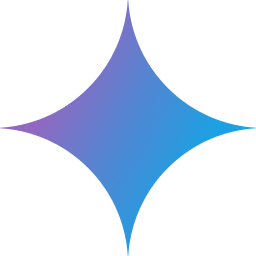}\xspace}
\DeclareRobustCommand{\mistral}{\icon{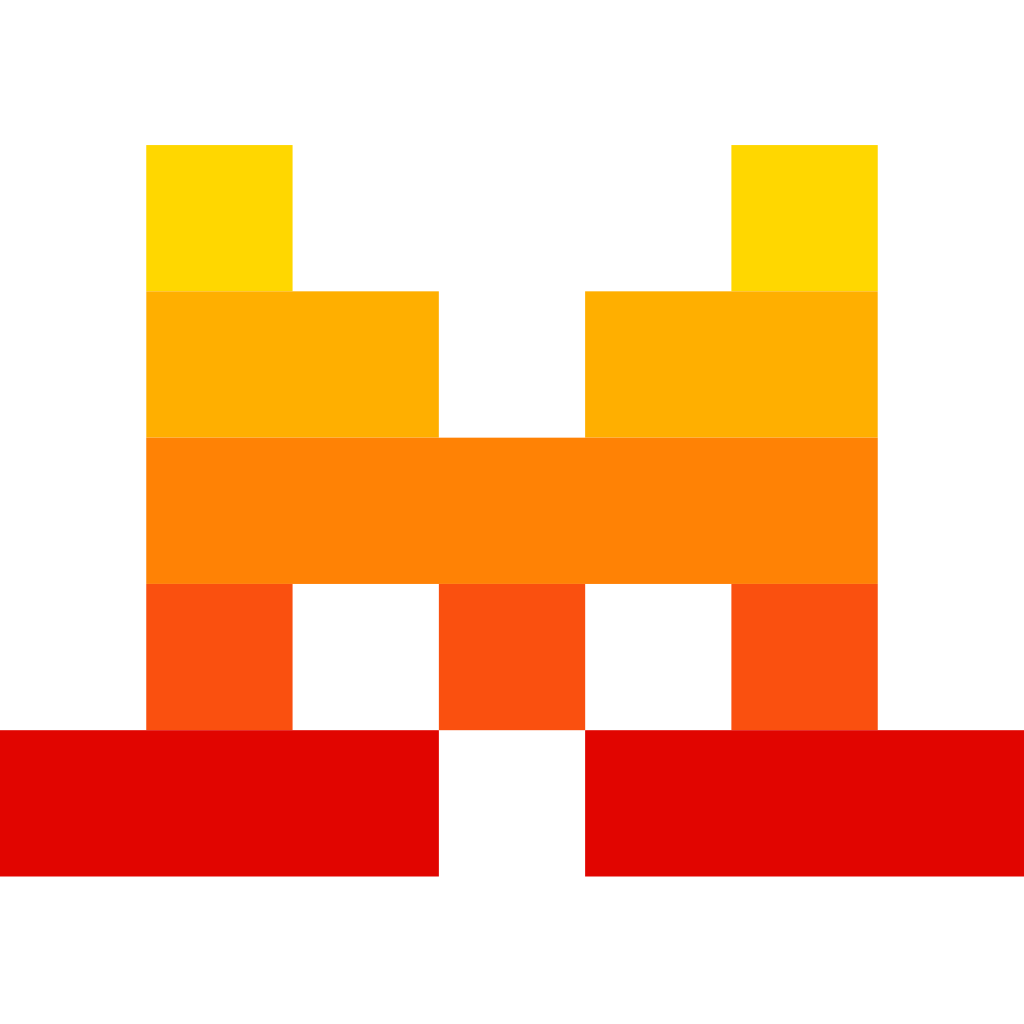}\xspace}

\usepackage{enumitem}

\definecolor{project_url_color}{HTML}{E87B00}

\title{\bn: Are User Simulators Reliable Proxies for Multi-Turn Evaluation of AI Assistants?}

\author{
  {\bfseries
    Yao~Dou}$^{1}$ \quad
  {\bfseries
    Michel~Galley}$^{2}$ \quad
  {\bfseries
    Baolin~Peng}$^{2}$ \quad
  {\bfseries
    Chris~Kedzie}$^{2}$\textsuperscript{†} \quad
  {\bfseries
    Weixin~Cai}$^{2}$ \\
  {\bfseries
    Alan~Ritter}$^{1}$ \quad
  {\bfseries
    Chris~Quirk}$^{2}$ \quad
  {\bfseries
    Wei~Xu}$^{1}$ \quad
  {\bfseries
    Jianfeng~Gao}$^{2}$ \\[4pt]
  $^{1}$Georgia Institute of Technology \qquad
  $^{2}$Microsoft 
  \\ \\
  \faIcon[regular]{envelope}\,
  \texttt{\href{mailto:douy@gatech.edu}{douy@gatech.edu}},
  \texttt{\href{mailto:mgalley@microsoft.com}{mgalley@microsoft.com}} \quad
  \faGlobe\,%
  \begingroup
    \hypersetup{urlcolor=project_url_color}
    \href{https://aka.ms/SimulatorArena}{aka.ms/SimulatorArena}
  \endgroup
}

\usepackage{listings}
\usepackage{tikz}
\usetikzlibrary{shapes,calc,positioning}

\global

\usepackage[most]{tcolorbox}
\tcbset{
  aibox/.style={
    width=\textwidth,
    top=10pt,
    colback=white,
    colframe=black,
    colbacktitle=black,
    enhanced,
    center,
    attach boxed title to top left={yshift=-0.1in,xshift=0.15in},
    boxed title style={boxrule=0pt,colframe=white,},
  }
}
\newtcolorbox{AIbox}[2][]{aibox,title=#2,#1}

\newcommand{\bn}{SimulatorArena\xspace}

\usepackage[textsize=scriptsize]{todonotes}

\definecolor{darkorange}{HTML}{aa5e02} 
\definecolor{lightred}{HTML}{ad0303} 
\definecolor{lightpurple}{HTML}{B446B4}
\definecolor{tabletitlebrown}{HTML}{9F6C0D}
\definecolor{tabletitlegreen}{HTML}{0A8016}
\definecolor{tabletitlepurple}{HTML}{430D9F}
\definecolor{tabletitleblue}{HTML}{0C1A97}
\definecolor{baseline}{HTML}{4d4d4d}  
\definecolor{length}{HTML}{2563eb}    
\definecolor{profile}{HTML}{b91c1c}   
\definecolor{both}{HTML}{7e22ce}      

\begin{document}

\maketitle

\renewcommand{\thefootnote}{\relax}
\footnotetext{\hspace*{-1.8ex}\textsuperscript{†} Now at Scaled Cognition.}

\setcounter{footnote}{0}
\renewcommand{\thefootnote}{\arabic{footnote}}

\begin{abstract}

Large language models (LLMs) are increasingly used in interactive applications, and human evaluation remains the gold standard for assessing their performance in multi-turn conversations. 
Since human studies are costly, time-consuming, and hard to reproduce, recent work explores using LLMs to simulate users for automatic assistant evaluation.
However, there is no benchmark or systematic study to evaluate whether these simulated users are reliable stand-ins for real users.
To address this, we introduce \bn, a benchmark of 909 annotated human–LLM conversations on two interactive tasks---math tutoring and document creation.
\bn evaluates simulators based on how closely their messages match human behavior and how well their assistant ratings align with human judgments.
Experiments on various simulator methods show that simulators conditioned on user profiles, capturing traits like background and message styles, align closely with human judgments. They reach Spearman's $\rho$ of 0.7 on both tasks, providing a practical, scalable alternative to human evaluation.
Using the best simulator for each task, we benchmark 18 assistants, including the latest LLMs such as GPT-5, Claude 4.1 Opus, and Gemini 2.5 Pro.

\end{abstract}
\section{Introduction}

\begin{figure}[!t]
  \centering
    \includegraphics[width=0.99\linewidth]{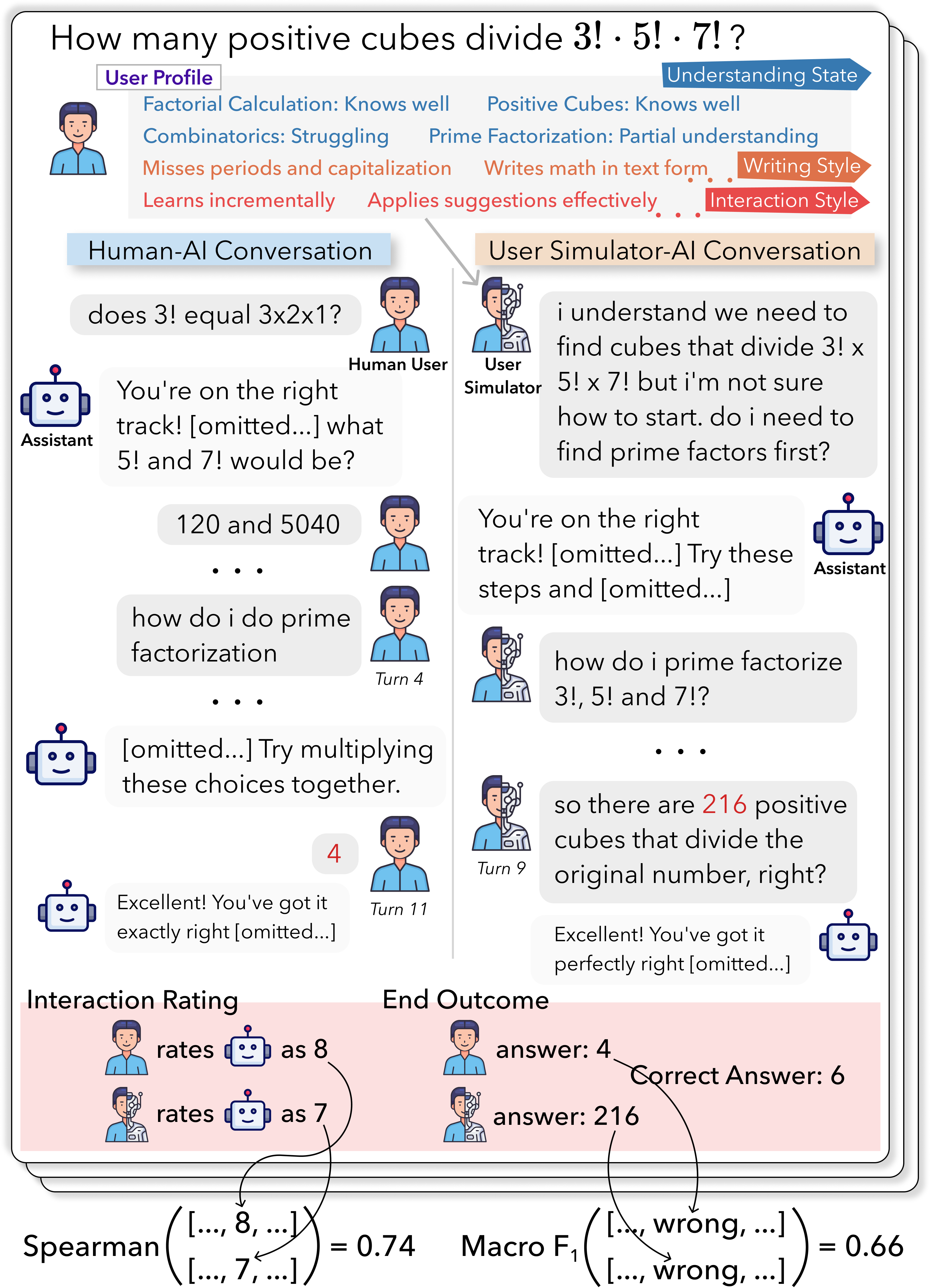}
    \vspace{-19pt}
 \caption{\bn systematically evaluates user simulators by comparing their behavior to humans'. User profiles improve simulator quality, offering an efficient, scalable alternative to human evaluation.}
    \label{fig:figure1}
    \vspace{-16pt}
\end{figure}

\begin{table*}
\centering
\setlength{\tabcolsep}{4pt}
  \renewcommand{\arraystretch}{1}
  \resizebox{0.99\linewidth}{!}{
\begin{tabular}{lccccccccc}
\toprule
\textbf{Dimension} & \makecell[c]{MT-Bench\\{\footnotesize\citeyear{zheng2023judging}}} &
\makecell[c]{MT-Eval\\{\footnotesize\citeyear{kwan2024mt}}} & 
\makecell[c]{MT-Bench-101\\{\footnotesize\citeyear{bai2024mt}}} & 
\makecell[c]{MINT\\{\footnotesize\citeyear{wang2023mint}}} & 
\makecell[c]{$\tau$-bench\\{\footnotesize\citeyear{yao2024tau}}} & 
\makecell[c]{MediQ\\{\footnotesize\citeyear{li2024mediq}}} & 
\makecell[c]{Anthropomorphism\\{\footnotesize\citeyear{ibrahim2025multi}}} & 
\makecell[c]{LostInConv.\\{\footnotesize\citeyear{laban2025llms}}} & 
\textbf{Ours} \\
\midrule
Dynamic Interaction & \redxmark & \redxmark & \redxmark & \greencmark  & \greencmark & \greencmark & \greencmark & \greencmark & \greencmark \\
Multi-Dynamic Tasks & \redxmark & \redxmark & \redxmark & \redxmark & \redxmark & \redxmark & \redxmark & \redxmark & \greencmark \\
Fine-grained User Profile & \redxmark & \redxmark & \redxmark & \redxmark & \redxmark & \redxmark & \redxmark & \redxmark & \greencmark \\
Systematic Sim. Eval. & \redxmark & \redxmark & \redxmark & \redxmark & \redxmark & \redxmark & \redxmark & \redxmark & \greencmark  \\
\bottomrule
\end{tabular}}
\vspace{-3pt}
\caption{Comparison of related work and our work across key  dimensions.
\textit{Dynamic Interaction:} conducts multi-turn engaged interactions instead of pre-defined questions.
\textit{Multi-Dynamic Tasks:} includes tasks with varied interaction dynamics. 
\textit{Fine-grained User Profile:} simulates users with diverse message style and inherent knowledge.
\textit{Systematic Sim. Eval.:} provides a comprehensive automatic framework for evaluating the quality of user simulators.}
\label{tab:related_work}
\vspace{-12pt}
\end{table*}

Large language models (LLMs) \cite{achiam2023gpt} have emerged as transformative tools, 
with interactive systems being one of their most influential applications.
From education \cite{wang2024tutor} to creative collaboration \cite{dhillon2024shaping}, LLMs are increasingly expected to engage in meaningful, multi-turn conversations that adapt to users with various needs and capabilities.

To evaluate these multi-turn interactions, human studies are the gold standard \cite{alexa,dstc,ji2022achieving,collins2024evaluating}. However, recruiting diverse users for long, realistic conversations with LLMs is costly and time-consuming, often limiting such evaluations to only a few models \cite{jurenka2024towards}. Human evaluations are also difficult to replicate and not directly comparable across different rounds due to the challenge of maintaining a consistent user pool. To address these limitations, recent work has explored using LLMs to simulate users for automated evaluation. Yet, no established benchmarks or systematic studies currently exist to assess the quality of these simulators, and it remains unclear whether they can accurately approximate real user behavior or serve as reliable substitutes for human evaluation.

In this paper, we introduce \bn, a benchmark for evaluating user simulators. It contains 909 authentic, annotated human–LLM conversations, collected from 107 participants chatting with 9 LLMs on two interactive tasks: math tutoring and document creation. Each conversation averages over 7 turns and lasts more than 20 minutes. \bn~enables automatic evaluation of (1) how closely simulator messages resemble real user messages, using metrics such as LLM-judged Turing tests, and (2) how well simulators align with human evaluations of assistants, measured by the correlation between the assistant ratings from the simulator and human users (see Figure \ref{fig:figure1}).

\begin{figure*}[!t]
  \centering
    \includegraphics[width=0.99\linewidth]{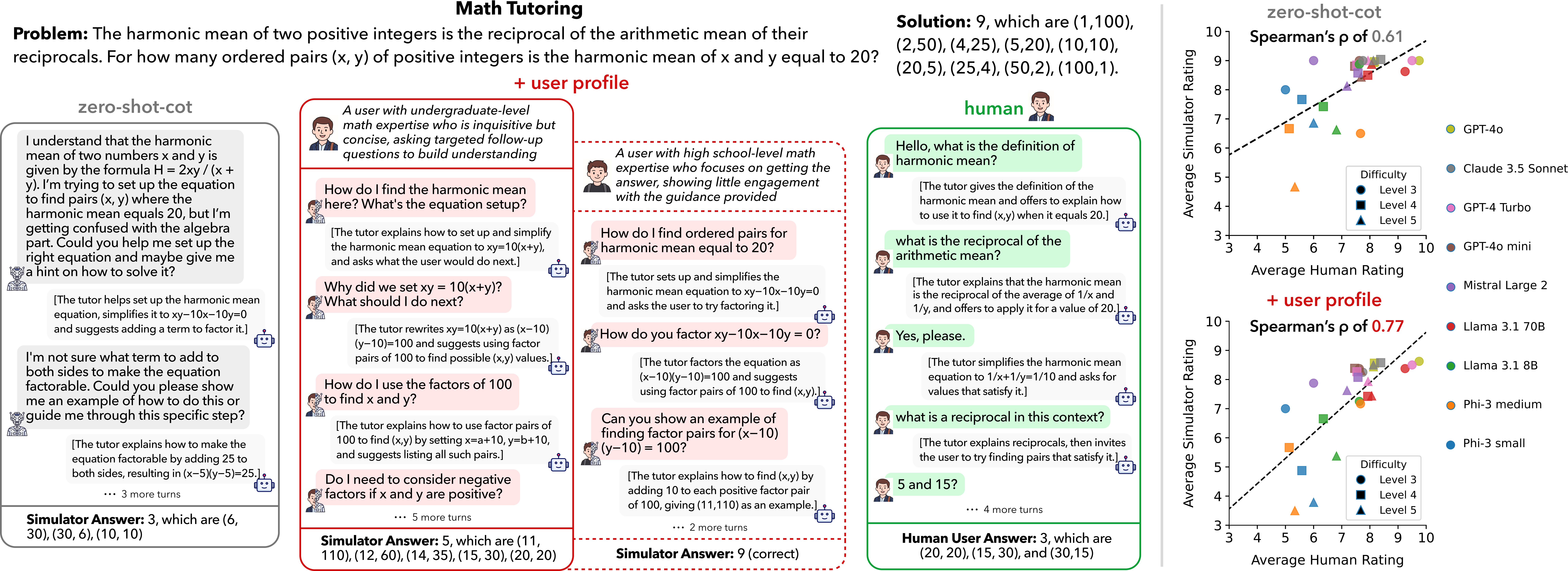}
    \vspace{-3pt}
 \caption{\textbf{Left:} Example math tutoring conversations with a zero-shot-cot simulator, two profile-based simulators with different user profiles, and a real human user. Assistant responses are summarized for space.
 \textit{\textbf{Takeaways:}} (1) The profile-based simulator produces messages that better resemble human users than the zero-shot-cot baseline; (2) Different user profiles lead to different conversation flows and outcomes. \textbf{Right:} Correlation between simulator and human ratings of assistant performance, computed over 27 grouped data points (model × difficulty level), shown as scatter plots.
\textit{\textbf{Takeaway:}} User profile-based simulator significantly improves correlation with human judgments from 0.61 Spearman’s $\rho$ to 0.77. Full conversations and document creation examples are in Appendix \ref{app:examples}.}
    \label{fig:examples_and_scatter}
    \vspace{-12pt}
\end{figure*}

With \bn, we first evaluate vanilla zero-shot and CoT prompting methods as user simulators.
We find that even when prompted to act as the user, LLM-simulated users differ noticeably from real users, often producing overly verbose and polite responses.
To capture the rich variation in user behavior, we introduce detailed user profiles---including document preferences, math expertise, and communication style with over 25 fine-grained attributes like grammar usage, message length, and feedback style. Compared to vanilla prompting, user profile-based simulators not only produce more realistic user behavior, but also significantly improve alignment with human evaluations, raising Spearman correlation from 0.61 to 0.77 in math tutoring and from 0.55 to 0.70 in document creation, as shown in Figure~\ref{fig:examples_and_scatter}.

Regarding task choices in \bn, we select math tutoring and document creation because they exhibit highly varied interaction patterns, making them ideal for testing simulators across scenarios. For example, a student might respond with minimal input, avoid formal math notation, ask for repetition, and push back on the tutor's suggestions.
These tasks also represent two of the most common LLM use cases---education and content generation \cite{tamkin2024clio}---and differ in interaction dynamics: in tutoring, the assistant provides domain knowledge; in document creation, the user brings background information and content preferences.

Our experiments show that the optimal user-profile configuration varies by task. In math tutoring, interaction style improves the human correlation the most---attributes like message length, clarification seeking, and feedback style largely shape the problem-solving dialogue. In document creation, a full profile works best: background information, document preferences, and communication style jointly shape the conversation. We also conduct a fine-grained analysis to identify specific user attributes that simulators struggle to mimic, such as avoiding LaTeX notation or introducing grammar mistakes. Finally, we use the best simulator for each task to evaluate 18 models including the most recent LLMs such as GPT-5 and Claude 4.1 Opus. To support future research, we release all data, annotation tools, and code at \href{https://aka.ms/SimulatorArena}{this link}.

In summary, our contributions are as follows:
\begin{enumerate}[leftmargin=*,itemsep=-0.3em,topsep=-0.3em]
\item We introduce \bn, a benchmark for evaluating user simulators in multi-turn conversations on two tasks: math tutoring and document creation. It contains 909 real human–LLM dialogues covering nine state‑of‑the‑art models.

\item \bn~provides automatic evaluation that measures both message realism and a simulator's ability to benchmark assistant models.

\item We develop user profile-based simulators that capture latent knowledge and message styles, improving correlation with human judgments by 26\% over vanilla prompts at less than 3\% of the cost of human evaluation.

\item Using the best simulators, we benchmark 18 LLMs on the two tasks and find that GPT-5 achieves the strongest performance across both.
\end{enumerate}

\section{LLM as User Simulator for Evaluation}

\subsection{Problem Formulation}
We formalize the evaluation of LLM assistants using user simulators as follows:
Let $\pi_u$ be the user simulator, $\pi_a$ the assistant, and $\pi_r$ the rater.
The user simulator $\pi_u$ engages in a multi-turn conversation with the assistant $\pi_a$. At turn $t$, the simulator generates an utterance $y_u^t \sim \pi_u(\cdot \mid I_u, S_u, H_{t-1})$, conditioned on user information $I_u$, message style $S_u$, and conversation history $H_{t-1}$.
The assistant then replies with $y_a^t \sim \pi_a(\cdot \mid I_a, H_{t-1}, y_u^t)$. The full conversation history up to turn $t$ is denoted by $H_t = {(y_u^1, y_a^1), \ldots, (y_u^t, y_a^t)}$.

A key aspect of this interaction is information asymmetry--- the simulator and assistant have access to different information. This asymmetry is essential for meaningful interaction; otherwise, the assistant could complete tasks directly without engaging with the user. It may arise from differences in knowledge, goals, or user preferences that the assistant must infer. We denote the simulator's information as $I_u$ and the assistant's as $I_a$. Crucially, providing all of this information upfront is unrealistic, as real users tend to communicate iteratively, and many instructions naturally emerge over the course of a conversation.

After the conversation ends at turn $T$, the rater $\pi_r$ evaluates the assistant’s performance, either based on the full interaction history $H_T$ or on the quality of an extracted outcome $o$ from $H_T$. The rater can be a language model or a metric, e.g., $F_1$.

\begin{figure*}[!t]
  \centering
    \includegraphics[width=0.99\linewidth]{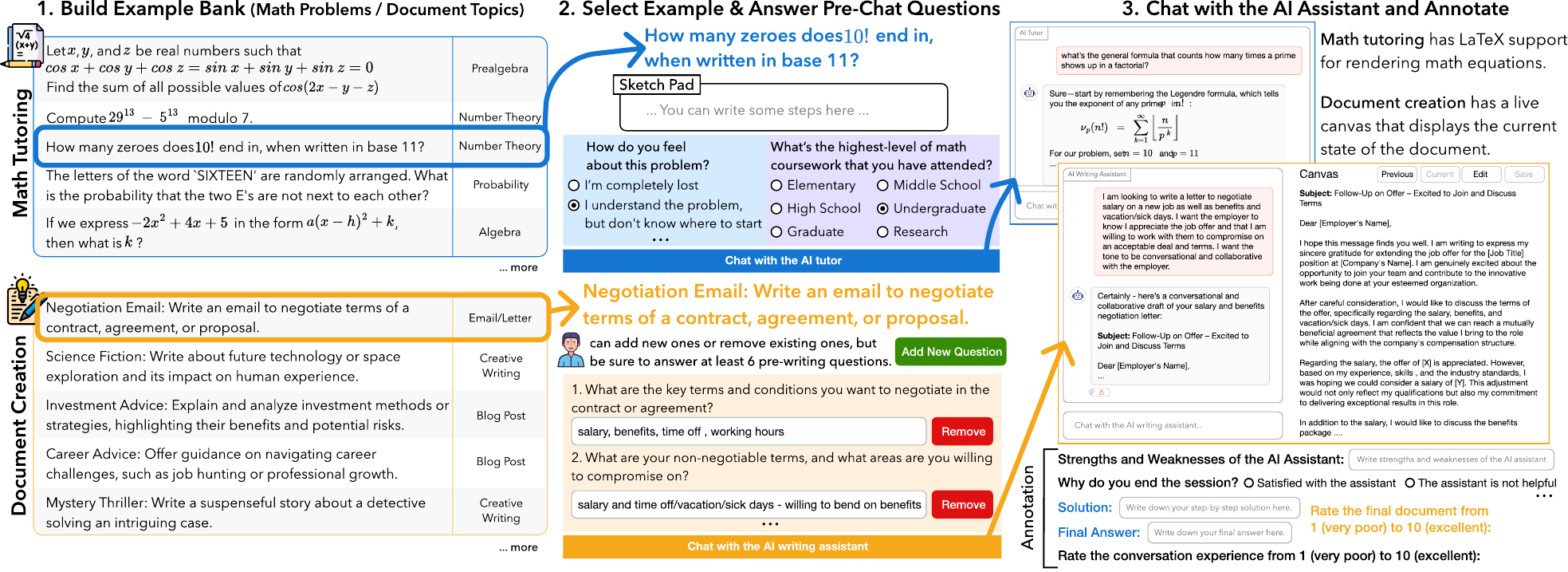}
    \vspace{-5pt}
 \caption{\textbf{Annotation Process.} To elicit authentic human-AI conversations, our interface follows a three‑step workflow: (1) We curate a bank of hundreds of math problems and common document topics.
(2) Annotators select the problem or topic that interests them and answer brief pre‑chat questions  to familiarize themselves with the task, an especially important step for document creation, where initial content ideas and context guide writing.
(3) They then converse with the AI assistant and, upon finishing the dialogue, evaluate its performance.}
    \label{fig:annotation_process}
    \vspace{-12pt}
\end{figure*}

\subsection{User Simulator Methods}

We conduct extensive experiments with three prompting methods with ten configurations.
We choose zero-shot prompting over few-shot as it transfers easily to new tasks and collecting real user conversations is resource-intensive.
We describe each method below.

\noindent \textbf{Zero-shot.} 
The simulator $\pi_u$ generates the user message $y_u^t$ based on user information $I_u$ and conversation history $H_{t-1}$, without CoT. The input $I_u$ includes the user’s intent and relevant background (e.g., the math problem or email recipient).

\noindent \textbf{Zero-shot CoT.}
The simulator first generates a thought process based on the conversation context, followed by the user message $y_u^t$ \cite{wei2022chain}.

\noindent \textbf{Zero-shot CoT with User Profile.}
As shown in Figure~\ref{fig:examples_and_scatter}, the zero-shot CoT simulator generates overly verbose and polite messages that fully articulate the reasoning process---unlike real user behavior.
This mismatch arises because the LLM is originally trained to act as an assistant, not a user.
To address this, we propose incorporating a comprehensive user profile into the simulator's prompt.
Rather than relying on message-level content, which is often too specific and difficult to generalize, the profile captures high-level user attributes derived from the conversation. These include enhanced user information $I_u$, which contains \textit{inherent knowledge} such as the user’s understanding state (in math tutoring) or document preferences (in document creation), and \textit{message style} $S_u$, which describes how the user communicates, including writing habits and interaction patterns.

\noindent \textit{\textcolor{darkorange}{Inherent Knowledge.}}
We enrich $I_u$ with the user's understanding state (for math tutoring) or document preferences (for document creation) beyond basic intent or background.
We develop an automatic and generalizable pipeline with \texttt{GPT-4o} to extract this information from the real user-LLM conversation. 
For understanding state, we first extract concepts needed to solve the problem based on its solution, and then categorize the user's knowledge for each concept into four levels: \textit{Knows well, Partial understanding, Struggling}, and \textit{Not introduced} based on their conversation.
For document preferences, we begin by identifying relevant preference attributes (e.g., Formality Level, Tone) for each document type. For example, emails may include attributes such as Greeting and Sign-off Style. We extract these attribute sets based on five conversations per document type. Once the attribute list is finalized, we extract the user’s preference for each attribute from their conversation.

\noindent \textit{\textcolor{lightred}{User Message Style.}} User message style $S_u$ captures how a user writes and interacts. We start with 11 general\textit{ writing style} attributes and 10 \textit{interaction style} attributes that apply across tasks (e.g., grammar usage, answer thoroughness).
Given different tasks have unique user behavior---for example, math tutoring often involves math notations---we automatically expand the attribute sets for each task in a resource-efficient way.
Specifically, we sample 10 real user–LLM conversations and their corresponding zero-shot CoT simulations. Using a contrastive prompting approach, we ask \texttt{GPT-4o} to identify attributes that distinguish real users from simulators.
After 10 iterations, the model generates over 60 attributes, which are then finalized into a concise list: 12 writing + 17 interaction attributes for tutoring, and 15 writing + 13 interaction attributes for document creation.

Since message length is a key difference between real and simulated users, we introduce a \textbf{length-controlled variant} that adds a note in the prompt specifying a target range. We use a range rather than exact lengths, as exact values are too specific and hard to enforce. The range is derived from the minimum and maximum lengths in the human conversation, then widened by rounding the minimum down and the maximum up to the nearest multiple of five (e.g., 3--43 becomes 1--45).
We treat message length as an interaction-style attribute, and it can be added to any simulator prompt.

We use \texttt{GPT-4o} as our user simulator $\pi_u$ throughout our experiments. All user simulation and data processing prompts are presented in Appendix \ref{app:prompts}.

\section{\bn: A Collection of Real User and LLM Multi-Turn Conversations}

To evaluate user simulators against real users, we curate \bn, a benchmark that contains 909 human–LLM conversations with annotations, including 450 in math tutoring and 459 in document creation. The conversations average 7.8 and 6.9 turns respectively and take more than 20 minutes, showing depth and quality of the interactions.

\subsection{Collecting Human-LLM Conversations}
Collecting authentic human–AI conversations is non-trivial, as simply presenting users with random problems or document topics results in shallow, generic interactions.
We thus design task-specific interfaces that follow a three-step data collection workflow, described below and  illustrated in Fig.~\ref{fig:annotation_process}.

\noindent \textbf{Step 1. Construct Example Bank.}
Instead of assigning users random examples, we build an example bank from which they can select a problem or document topic. For \textit{math tutoring}, we use 1,000 problems from the MATH dataset \citep{hendrycks2021measuring} with difficulty levels 3-5 to better engage adult users, unlike prior work that uses basic arithmetic problems for K-12 learners \citep{wang2024tutor}. For \textit{document creation}, we focus on three common types: email/letter, creative writing, and blog posts. For each type, we manually create a list of example topics,  though users may also choose to write about their own topic (e.g., “Biscuits recipe: Making grandma’s homemade biscuits”).

\noindent \textbf{Step 2. Select an Example and Prepare Pre-writing Materials.}
For \textit{math tutoring}, users select a challenging problem of interest and may jot down initial steps on a scratch pad before interacting with the AI tutor. For \textit{document creation}, users choose a document topic or enter their own. Since people often have a rough idea before writing, we mimic this process by using GPT-4o to generate 10 pre-writing questions to help users brainstorm the core elements of the document. Users can add or remove questions but are required to answer at least six.

\noindent \textbf{Step 3. Chat with AI Assistant and Annotation.}
Users then interact with the AI assistant through task-specific interfaces. The math tutoring interface features a standard chat window that renders LaTeX, and the document creation interface mimics OpenAI Canvas,\footnote{\url{https://openai.com/index/introducing-canvas/}} with \texttt{GPT-4o-mini} updating the document after each turn. After the conversation, users rate the assistant's interaction quality on a 1–10 scale (see definitions in  App.~\ref{app:rating_definition}). In math tutoring, they also submit a final solution; in document creation, they also provide a 1–10 rating for the final document. Additionally, users can give a thumbs up or down for each assistant response.

\subsection{Annotation Details}

We use Amazon Mechanical Turk (AMT) to collect the human–LLM conversations, with 107 unique workers---66 for math tutoring and 75 for document creation---interacting with 9 LLMs. Each worker is paid \$20 USD per hour. The total annotation cost is around \$10,000 USD.

\noindent \textbf{Assistant Model Coverage.}
To evaluate how effectively user simulators can evaluate a diverse range of LLMs, we consider the following nine LLMs as assistants covering both open- and closed-source models:  \texttt{GPT-4o} (2024-05-13), \texttt{GPT-4o mini}, \texttt{GPT-4 Turbo} \cite{achiam2023gpt}, \texttt{Mistral Large 2} (2407) \cite{jiang2023mistral}, \texttt{Claude 3.5 Sonnet} (20240620) \cite{TheC3}, \texttt{Llama 3.1 8B} and \texttt{70B} (Instruct versions) \cite{dubey2024llama}, and \texttt{Phi-3 small} and \texttt{medium} (128k Instruct versions) \cite{abdin2024phi}.

\noindent \textbf{Human Evaluation Results.}
Table \ref{tab:model_evaluation} in Appendix \ref{app:data_stats} shows the statistics and human ratings for each assistant model. In math tutoring, \texttt{GPT-4o} outperforms the others. In document creation, all models receive similar ratings for both interaction and document quality, except for the \texttt{Phi-3} models.
 
\noindent \textbf{User Experience Diversity.}  
We observe that users' experiences with the same assistant on the same topic or math problem often vary widely.
As crowdworkers selected their preferred topics or problems and were randomly assigned assistant models, we obtained 186 annotator pairs for document creation and 34 for math tutoring.
In document creation, 78.5\% of annotator pairs gave different interaction ratings, with 41.9\% differing by more than one point; for document quality, 74.2\% of pairs disagreed, with 42.5\% differing by more than one point. In math tutoring, 85.3\% of pairs differed on interaction ratings (55.9\% by more than one point), and in 35.3\% of cases, one annotator reached the correct final answer while the other did not.
These results demonstrate the diversity of user experiences and the importance of modeling users individually through detailed profiles.

Additional details such as quality control, cost, and interface screenshots are provided in App.~\ref{app:annotation_details}.

\section{Evaluation of User Simulator}
\bn~evaluates user simulators $\pi_u$ along two key dimensions: how closely their messages resemble those of real users, and how well they elicit similar assistant behavior as in human-AI interactions. The latter is measured by comparing human ratings of assistant performance in human-AI interactions with ratings from a rater $\pi_r$ of assistant performance in simulator-AI interactions.

\subsection{Similarity Between Simulated and Real User Messages}
To assess message similarity, we use two complementary evaluation methods with \texttt{GPT-4o} as judge:

\noindent \textbf{Likert Scale.} We rate message similarity on a 1–5 Likert scale across two aspects: writing style and interaction style, with 5 indicating nearly indistinguishable from real user messages.

\noindent \textbf{Turing Test.}  We present two conversations to the evaluator, one between a real user and the AI, and one between a simulated user and the AI, and ask them to identify which involved the real user. An ideal result is a 50\% accuracy (i.e.,random guess), indicating the evaluator cannot distinguish between the two.  Accuracy significantly above or below 50\% suggests noticeable differences. To reduce position bias in the LLM judge \cite{zheng2023judging}, we run each evaluation twice with the conversation order swapped and collect a confidence score for each. If the same position is chosen both times, we use the confidence scores to break the tie.

\subsection{Alignment Between User Simulators and Human Evaluation of AI Assistants}

When evaluating assistants, we focus on two key aspects: \textbf{interaction quality}, which reflects how the conversation feels, and \textbf{outcome quality}, which measures how well the assistant completes the user’s task.
To test the effectiveness of user simulators for evaluation, we compare the interaction ratings that assistants receive when engaging with simulators versus real users, and examine whether they achieve similar outcomes in both settings.

\noindent \textbf{Interaction Aspect.}
A rater LLM $\pi_r$ rates assistant performance in each conversation on a 1–10 scale using the same criteria as human users (1=very poor, 10=excellent).
To reduce annotator bias, we apply z-score normalization to human ratings. For annotators with only one or two conversations, ratings are grouped and normalized using the group's mean and standard deviation.
We compute Spearman’s $\rho$ (primary metric), Pearson’s $r$, and Kendall’s $\tau$; the latter two are reported in App.~\ref{app:further_results}.
Correlations are calculated at three levels: \textit{instance} (each user-assistant conversation), \textit{intermediate} (27 groupings (model $\times$ difficulty for math, model $\times$ document type for writing), and \textit{system} (one score per model). We focus on intermediate-level, as it smooths out instance-level noise while providing a finer-grained view than system-level.

\begin{figure}[!t]
  \centering
    \includegraphics[width=0.9\linewidth]{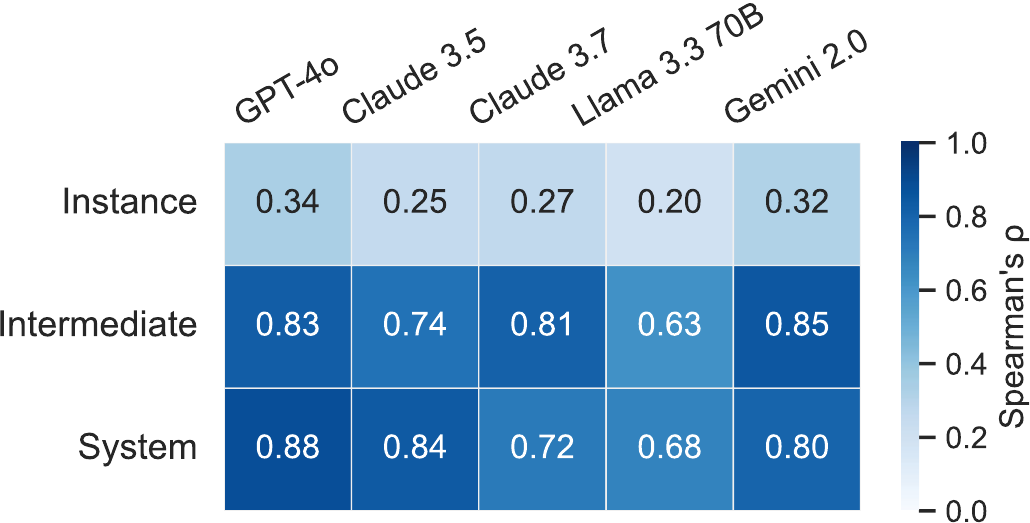}
    \vspace{-5pt}
 \caption{Evaluation of different LLMs as raters $\pi_r$ based on their alignment with human ratings for final documents in the document creation task.}
    \label{fig:rater_evaluation_doc_doc_corr}
    \vspace{-6pt}
\end{figure}

\begin{figure}[!t]
  \centering
    \includegraphics[width=0.90\linewidth]{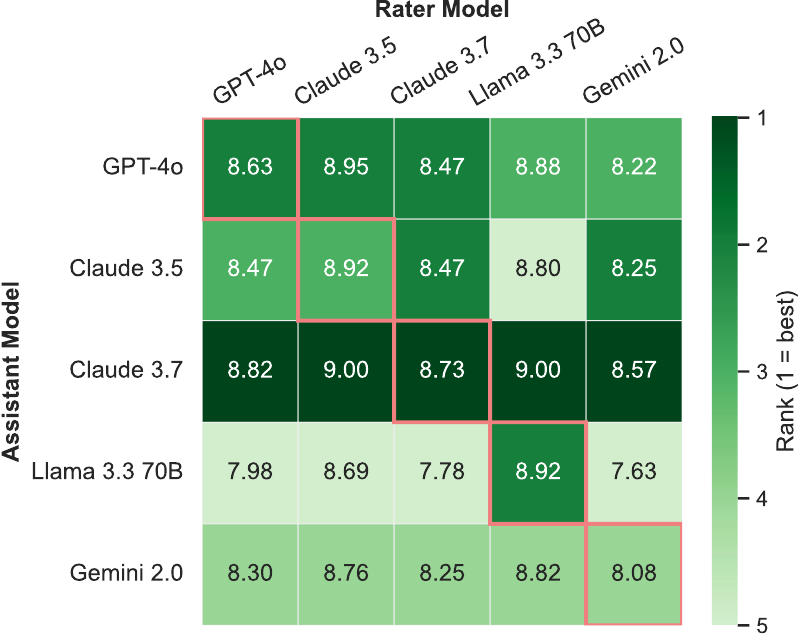}
    \vspace{-5pt}
 \caption{Evaluation of different LLMs as raters $\pi_r$ for self-bias. All models show no evidence of self-bias.}
    \label{fig:rater_evaluation_doc_doc_bias}
    \vspace{-12pt}
\end{figure}

\noindent \textbf{End Outcome Aspect.}
We evaluate whether assistants produce similar outcomes when interacting with simulators versus real users. For math tutoring, we compare final answer correctness---i.e., if the human user answers correctly, does the simulator as well---and report Macro $F_1$, the average of $F_1$ scores on correct and incorrect classes. For document creation, we compute the correlation between rater and human ratings on the final documents.

We evaluate five LLMs as candidates for rater $\pi_r$ by measuring how well their ratings of assistants in human–AI conversations align with human ratings and checking for self-bias, whether they rate their own assistant's outputs more favorably. Results for document creation outcome aspect are shown in Figures~\ref{fig:rater_evaluation_doc_doc_corr} and \ref{fig:rater_evaluation_doc_doc_bias}; results for interaction aspect and math tutoring appear in Appendix~\ref{app:further_results}.
We select GPT-4o as $\pi_r$ for all later experiments, as it achieves top correlations and shows no self-bias. Its intermediate correlations are 0.83 for math tutoring interaction, 0.89 for document creation interaction, and 0.83 for document creation outcome.
Evaluation prompts are listed in Appendix \ref{app:evaluation_prompts}.

\begin{figure*}[!t]
  \centering
    \includegraphics[width=0.99\linewidth]{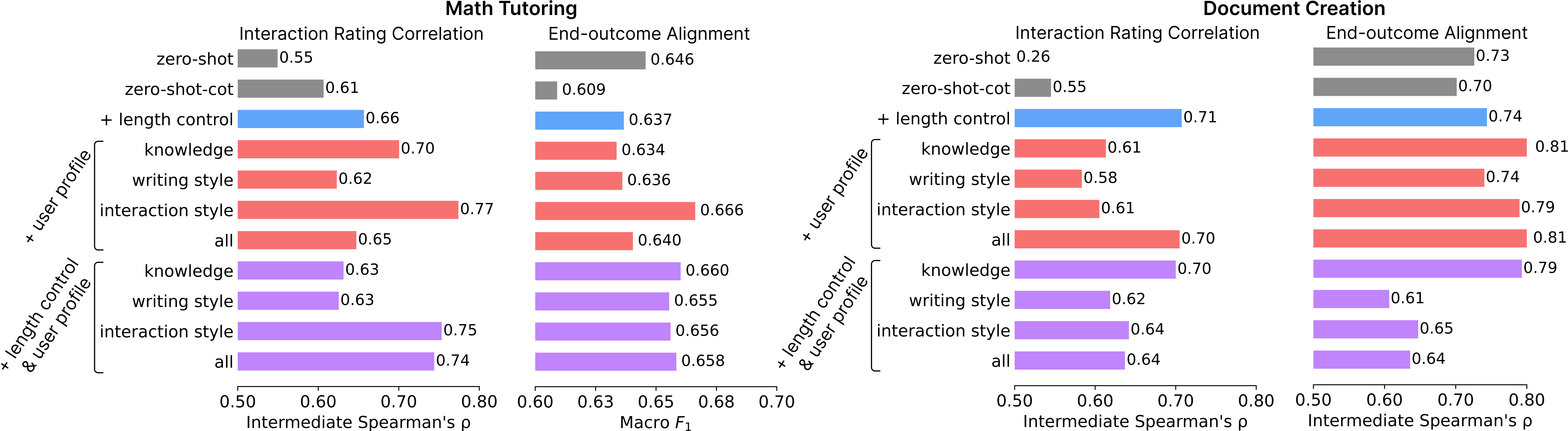}
    \vspace{-5pt}
 \caption{Alignment between user simulator and human evaluations of assistant performance. Simulators with user profiles outperform zero-shot and CoT baselines on both tasks. Interestingly, the most effective user-profile configuration varies by task: for \textbf{math tutoring}, using only interaction style works best, as the goal of task is more closed and interaction style strongly shapes assistant behavior. For \textbf{document creation}, the full user profile is most effective, as users provide richer input to guide the assistant in this more open-ended task.}
    \label{fig:extrinsic_evaluation}
    \vspace{-5pt}
\end{figure*}

\begin{table}[t]
\centering
\setlength{\tabcolsep}{4pt}
\renewcommand{\arraystretch}{1}
\resizebox{\columnwidth}{!}{%
\begin{tabular}{lrrrrrr}
\toprule
&
\multicolumn{3}{c}{\textbf{Math Tutoring}} &
\multicolumn{3}{c}{\textbf{Document Creation}}\\
\cmidrule(lr){2-4}\cmidrule(lr){5-7}
\textbf{Method} & \textit{\#Turns} & \textit{Length} & \textit{Cost (\$)} &
\textit{\#Turns} & \textit{Length} & \textit{Cost (\$)}\\
\midrule
zero-shot          & 5.8 & 87.6 & 0.02 & 6.0 & 77.5  & 0.05\\
zero-shot-cot      & 7.3 & 89.5 & 0.06 & 8.8 & 123.9 & 0.14\\
+ length control   & 9.7 & 21.4 & 0.06 & 11.9 & 30.5 & 0.17\\
+ user profile     & 8.2 & 23.5 & 0.09 & 7.1 & 58.2 & 0.10\\
\midrule
human & 7.8 & 15.5 & 5.33 & 6.9 & 32.6 & 6.50\\
\bottomrule
\end{tabular}}
\vspace{-3pt}
\caption{Conversation statistics for user simulators and humans. \textit{\#Turns} and \textit{Cost} are per conversation; \textit{Length} is average user message length in words.}
\label{tab:user_simulator_stats_short}
\vspace{-12pt}
\end{table}

\section{Results}
\begin{figure}[!t]
  \centering
    \includegraphics[width=0.99\linewidth]{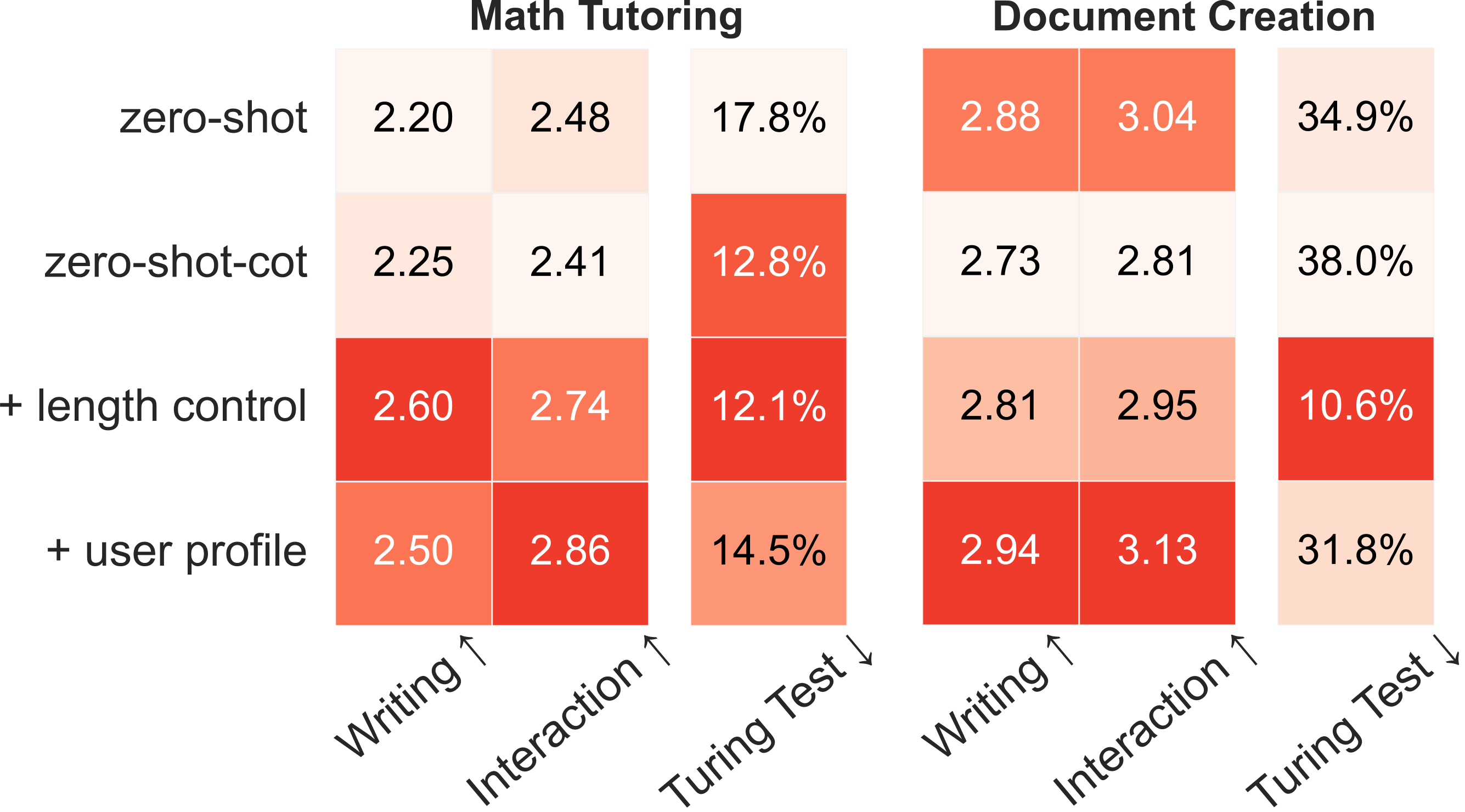}
    \vspace{-20pt}
 \caption{Similarity between simulated and real user messages on both tasks, evaluated with three metrics: \textit{Writing Style} and \textit{Interaction Style} are rated on a 5-point Likert scale (1=least similar, 5=most similar). \textit{Turing Test} is measured as $|p - 50|\%$, where $p$ is the LLM judge’s accuracy in identifying the real user.}
    \label{fig:similatiry}
    \vspace{-12pt}
\end{figure}

\begin{figure*}[!t]
  \centering
    \includegraphics[width=0.99\linewidth]{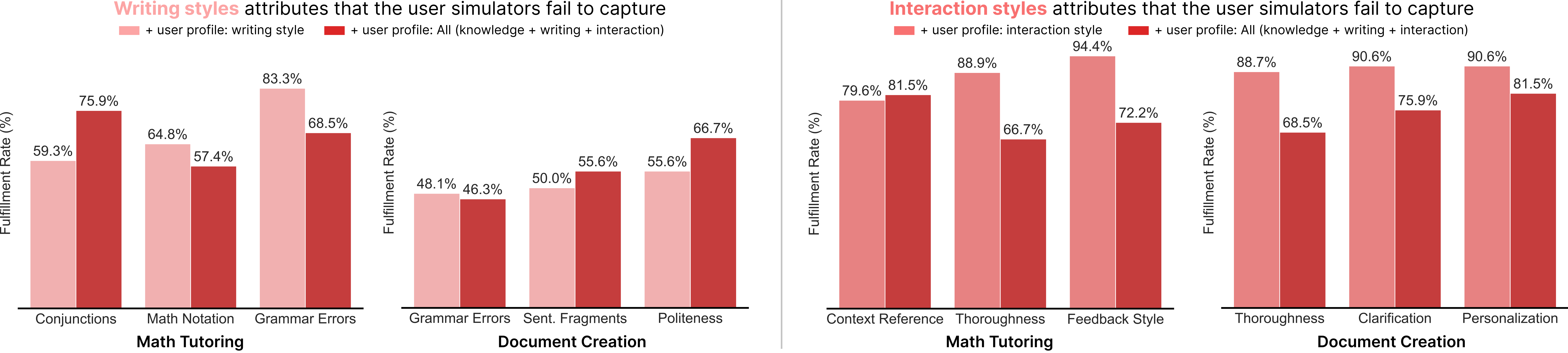}
    \vspace{-7pt}
 \caption{Attributes that profile-based simulators fail to capture, measured by prompting GPT-4o to assess whether each attribute is fulfilled in each conversation. Richer user profiles reduce fulfillment of interaction style attributes.}
    \label{fig:feature_fulfillment_analysis}
    \vspace{-10pt}
\end{figure*}

\begin{figure}[!t]
  \centering
    \includegraphics[width=0.99\linewidth]{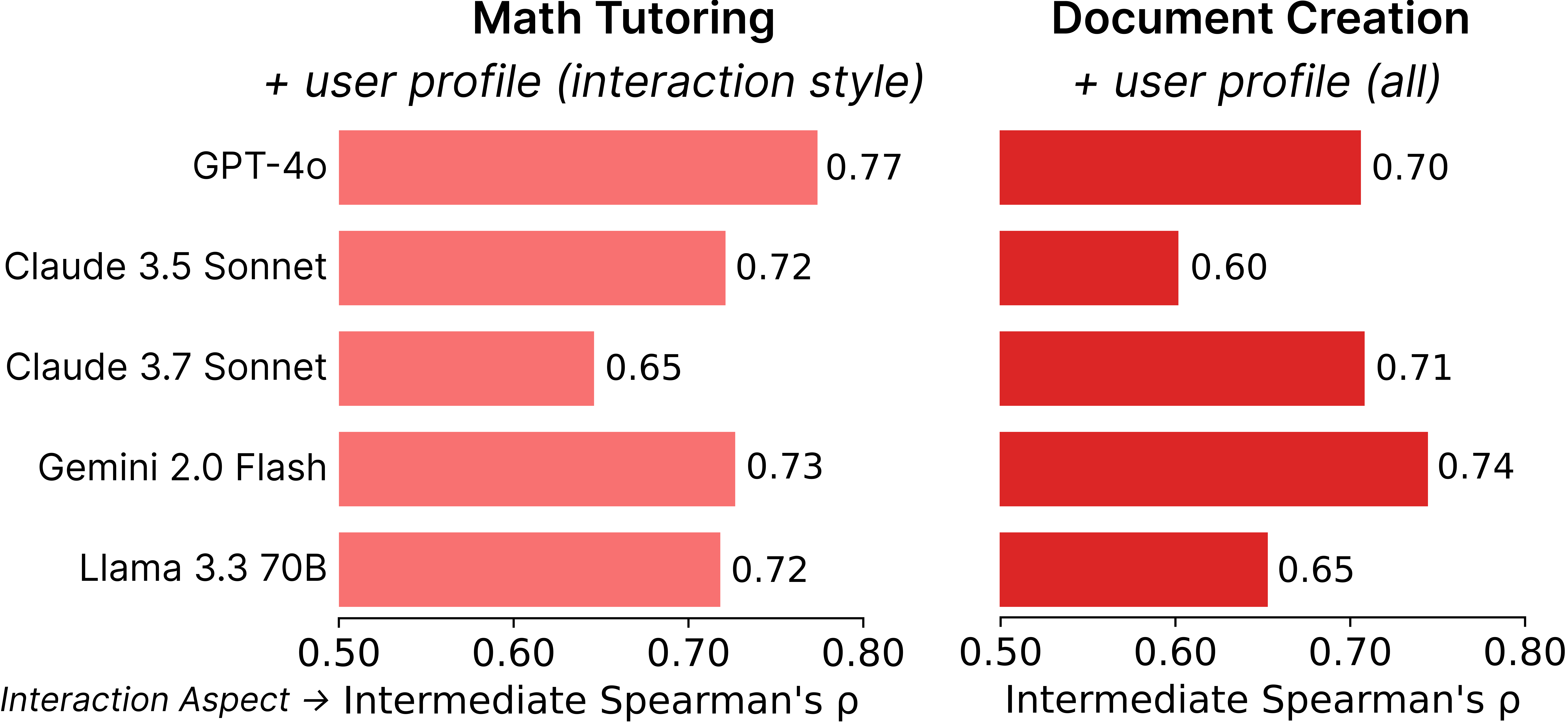}
    \vspace{-18pt}
 \caption{Comparison of different LLMs as user simulators. GPT-4o leads on math tutoring, while Gemini 2.0 Flash performs best on document creation.}
    \label{fig:extrinsic_eval_user_sim_model}
    \vspace{-12pt}
\end{figure}

We present key findings from our evaluation of user simulator methods.
Figure~\ref{fig:extrinsic_evaluation} shows how well each simulator aligns with human evaluations of assistant performance.
Table~\ref{tab:user_simulator_stats_short} summarizes statistics for four methods: zero-shot, zero-shot CoT, length-controlled, and the best-performing profile-based simulator based on alignment with human evaluations, and Figure~\ref{fig:similatiry} reports message similarity between simulated and real users for the same four methods.
Figure~\ref{fig:feature_fulfillment_analysis} highlights representative writing and interaction attributes that simulators fail to capture.
Figure~\ref{fig:extrinsic_eval_user_sim_model} evaluates different LLMs as simulators.
Detailed results are listed in App. \ref{app:further_results}.

\noindent \textbf{User simulators with user profiles produce more authentic interactions and more reliable assistant evaluations (Figure \ref{fig:extrinsic_evaluation}, \ref{fig:similatiry}).}
In both tasks, user profiles improve similarity to human messages in writing and interaction style, making simulated messages harder to distinguish from real ones compared to zero-shot and zero-shot CoT baselines. They also significantly improve alignment with human evaluations on both interaction and outcome. 
We conduct significance tests comparing the zero-shot-CoT baseline with the best user-profile variant for each task: for interaction and document-creation outcome (1-10 ratings), we use Williams’ test to compare dependent correlations; for math tutoring outcome (binary correctness), we apply McNemar’s exact test on paired predictions. All four comparisons were significant (three at $p < 0.01$ and one at $p < 0.05$), confirming the robustness of these improvements.

\noindent \textbf{The best user profile configuration differs by task: interaction style is most effective for math tutoring, while a full profile works best for document creation (Figure~\ref{fig:extrinsic_evaluation}).}
In math tutoring, the interaction-style profile shows the highest alignment with human ratings for both interaction quality and end outcome. As the task has a constrained goal, solving specific math problems, interaction style plays a key role in shaping assistant behavior.
In contrast, the open-ended nature of document creation benefits more from a full profile including preferences, writing style, and interaction style. Writing style alone has limited impact, suggesting that modern LLMs are robust to surface-level syntax variation.
Length control further improves richer profiles in math tutoring but has little effect in document creation except when paired with inherent knowledge. We posit this is because math tutoring messages are shorter, averaging 16 words vs. 33 in document creation, so each word carries more weight in shaping interaction dynamics.

\noindent \textbf{User simulators struggle to fulfill certain writing attributes, and interaction style attributes become less fulfilled with more detailed user profiles (Figure \ref{fig:feature_fulfillment_analysis}).}
For writing style, the most frequently missed attributes are conjunctions and math notation in math tutoring, and grammar errors and sentence fragments in document creation. Attributes like conjunctions and sentence fragments show improved fulfillment when interaction style is included, because they are influenced by interaction traits such as thoroughness.
For interaction style, however, adding more profile information leads to lower fulfillment across attributes, suggesting that current LLMs struggle to satisfy all behavioral constraints when too many are specified.

\noindent \textbf{GPT-4o achieves the highest human correlation in math tutoring, while Gemini 2.0 Flash performs best in document creation (Figure~\ref{fig:extrinsic_eval_user_sim_model}).}
We evaluate five LLMs as user simulators using the best profile for each task: interaction style for math tutoring and the full profile for document creation.
In math tutoring, GPT-4o achieves the highest correlation with human ratings (0.77) on interaction aspect, with three others closely behind at $\sim$0.73. In document creation, Gemini 2.0 Flash leads with 0.74, followed by GPT-4o and Claude 3.7 at $\sim$0.71.

\section{\bn~on Evaluating Assistant}
\label{sec:assistant_benchmark}
Using the best user simulators---GPT-4o with an interaction-style profile for math tutoring and Gemini 2.0 Flash with a full profile for document creation---we benchmark 18 assistants, including the most recent LLMs such as GPT-5, Claude 4.1 Opus, and Gemini 2.5 Pro, on 50 math problems (8 level‑3, 25 level‑4, 17 level‑5) and 51 document topics (17 × 3 types). Simulators use a fixed set of user profiles sampled from real human–AI conversations, ensuring all assistants are evaluated on the same tasks and interaction contexts.
 
We evaluate all models in non-thinking mode. For GPT-5, we set the reasoning effort to minimal; for Gemini 2.5 Pro, we use a thinking budget of 128 (the minimum allowed). Results are shown in Table~\ref{tab:model_evaluation_simulator}. GPT-5 performs the best on both tasks with the highest interaction rating (8.89) and accuracy (90\%) for math tutoring, and the second highest interaction (9.08) and highest document scores (8.96) for document creation.
Claude 3.7 Sonnet and Claude 4.1 Opus follow closely. Among open-source models, Phi-4 tops Llama 3.3 70B.

\begin{table}[t]
\centering
\setlength{\tabcolsep}{4pt}
\renewcommand{\arraystretch}{1}
\resizebox{\columnwidth}{!}{%
\begin{tabular}{lrrrr}
\toprule
& \multicolumn{2}{c}{\textbf{Math Tutoring}} & \multicolumn{2}{c}{\textbf{Document Creation}}\\
\cmidrule(lr){2-3}\cmidrule(lr){4-5}
\multirow{2}{*}{\textbf{Model}} &
  \makecell{Interaction\\Rating} &
  \makecell{Correct\\Rate (\%)} &
  \makecell{Interaction\\Rating} &
  \makecell{Document\\Rating} \\
\midrule
\openai~GPT-5\tablefootnote{As OpenAI’s reasoning models don’t support temperature changes, their temperature is fixed at 1.0. 
All other models are evaluated with temperature = 0.} & \first{8.89} & \first{90.0} & \second{9.08} & \first{8.96} \\
\anthropic~Claude 3.7 Sonnet & 8.70 & \first{90.0} & \first{9.10} & 8.73 \\
\anthropic~Claude 4.1 Opus & 8.71 & \third{82.0} & \first{9.10} & \second{8.90} \\
\openai~GPT-4 Turbo & 8.60 & \second{84.0} & 9.04 & 8.50 \\
\openai~GPT-4o & \third{8.84} & 76.0 & 9.02 & 8.59 \\
\anthropic~Claude 4 Sonnet & 8.74 & 70.0 & \third{9.07} & \third{8.80} \\
\openai~GPT-4.1 & \second{8.87} & 76.0 & \second{9.08} & 8.47 \\
\microsoft~Phi-4 & 8.66 & \second{84.0} & 8.96 & 8.39 \\
\anthropic~Claude 3.5 Sonnet & 8.66 & 76.0 & 9.06 & 8.41 \\
\openai~GPT-4o mini & 8.56 & 76.0 & 8.98 & 7.98 \\
\gemini~Gemini 2.5 Flash & 8.38 & 52.0 & 9.04 & 8.70 \\
\gemini~Gemini 2.5 Pro & 8.36 & 48.0 & 9.02 & 8.66 \\
\gemini~Gemini 2.0 Flash & 8.36 & 58.0 & 8.94 & 8.36 \\
\mistral~Mistral Large 2 & 8.08 & 64.0 & 8.98 & 8.25 \\
\llama~Llama 3.3 70B & 8.26 & 68.0 & 8.88 & 7.92 \\
\llama~Llama 3.1 70B & 7.70 & 70.0 & 8.86 & 8.00 \\
\llama~Llama 3.1 8B & 6.48 & 46.0 & 8.82 & 7.53 \\
\microsoft~Phi-3 Medium & 6.35 & 51.0 & 5.57 & 7.50 \\
\bottomrule
\end{tabular}}
\caption{Performance of 18 assistant models, evaluated by our best user simulators on the two tasks. Models are sorted in descending order by the mean z-score across four metrics. Top 3 models per metric are highlighted.}
\label{tab:model_evaluation_simulator}
\vspace{-12pt}
\end{table}

\section{Related Work}

\noindent \textbf{LLM Multi-Turn Evaluation.} Before LLMs, automatic conversation evaluation relied on static setups, where a model was given a human-written history and assessed on single responses \cite{vinyals2015neural,rastogi2020towards}, which failed to capture conversation-level performance \cite{mehri2020unsupervised}.
Recent work uses LLMs as judges. Benchmarks like MT-Bench and its variants \cite{zheng2023judging,kwan2024mt,bai2024mt,sun2024parrot} assess multi-turn ability but rely on pre-written user messages, leading to unnatural, shallow conversations. To evaluate multi-turn capabilities in task-oriented settings, where each turn builds on the last, prior work has used real users to interact and rate LLMs \cite{mehri2022interactive,Shen2023ParachuteEI,collins2024evaluating,ibrahim2024beyond}. While this remains the gold standard, it is costly. Our work shows that user simulators can be an effective, low-cost substitute.

\noindent \textbf{LLM as User Simulator.}
LLMs have been used as user simulators in various settings, such as simulating social interactions \cite{park2023generative,horton2023large}, economic experiments \cite{tang2024gensim}, and debates \cite{du2023improving,khan2024debating}. They have also been applied to collaborative multi-agent tasks \cite{li2023camel,guo2024embodied,tran2025multi}. Another line of work uses simulators to synthesize multi-turn training data for improving LLM downstream performance \cite{ding2023enhancing,ou2024inductive,sun2024parrot} and encouraging more proactive, collaborative behavior \cite{collabllm2025}.
Our work provides a comprehensive analysis of how to design and evaluate LLM simulators for accurately and cost-effectively assessing the assistant's interactions.

\noindent \textbf{User Simulator for Evaluation.}
\citet{ghandeharioun2019approximating} use self-play for chit-chat evaluation. With LLMs, researchers started using user simulators for evaluating interactive tasks like tool use, flight booking, and patient simulations \cite{wang2023mint,chen2024chatshop,sekulic2024reliable,yao2024tau,zhang2024usimagent,li2024mediq,li2024leveraging}. Simulators have also been used to evaluate broader assistant behaviors, including anthropomorphism \cite{ibrahim2025multi} and general multi-turn capabilities \cite{laban2025llms}.
However, no benchmark exists to systematically evaluate whether these simulators are good proxies for human evaluation.
Our work addresses this by introducing \bn and developing reliable user simulators conditioned on diverse user profiles.

More related work is presented in Appendix \ref{app:additional_related_work}.

\section{Conclusion}
We present \bn, a benchmark for evaluating user simulators in multi-turn conversations on math tutoring and document creation. Built on 909 real human-LLM dialogues, \bn enables automatic assessment of simulator quality through message realism and alignment with human ratings. Our results show that user profile-based simulators substantially improve alignment, achieving over 0.7 Spearman correlation---a 26\% gain over vanilla prompting---at just 3\% the cost of human evaluation. This highlights user simulators as an efficient, scalable path toward more reliable and human-aligned assistant evaluation.
Using the best user simulators, we further benchmark 18 state-of-the-art LLMs, including GPT-5, Claude 4.1 Opus, and Gemini 2.5 Pro, as math tutors and document writing assistants.

\section*{Limitations}
Our evaluations currently focus on a single conversation session. Future studies could investigate how the user simulator performs across multiple sessions.
Another limitation is that we mainly rely on prompting LLMs as user simulators, due to their strong generalization capabilities. However, since model distillation can significantly improve smaller models, future work could explore distilling our released conversations into more efficient user simulators.
Although our primary focus is on using simulators to evaluate LLM assistants, the simulated data could also be used to train LLM assistants for improved alignment and personalization. Investigating these training approaches would be an exciting avenue for future research.
Finally, our current evaluation centers on conversation-level metrics for both the user simulator and the assistant. Future work could look into more fine-grained, turn-level analyses to gain deeper insights into how to further improve user simulators.

\subsubsection*{Acknowledgments}
We thank Shirley Wu for helpful discussions and the three anonymous reviewers for their feedback.
We also thank Jonathan Zheng for beta testing our publicly released code.
Yao Dou, Wei Xu, and Alan Ritter are partially supported by the NSF CAREER Awards IIS-2144493 and IIS-2052498.
Any opinions, findings, and conclusions or recommendations expressed in this material are those of the authors and do not necessarily reflect the views of the National Science Foundation.

\bibliography{custom}

\newpage
\appendix
\section{Additional Related Work}
\label{app:additional_related_work}

\paragraph{Evaluating LLMs on Math Tutoring.} Existing work use human evaluation to assess LLM-based tutoring in math. \citet{kumar2023math} show that LLM-based explanations improve student’s learning of high school-level math. \citet{jurenka2024towards} collaborate with learners and educators to develop LearnLM-Tutor, which is then tested with human participants. \citet{wang2024tutor} demonstrate that LLMs can scale human tutors’ expertise for K–12 students from historically underserved communities. \citet{collins2024evaluating} investigate how LLMs perform as math assistants through interactions with undergraduates and professors. Additionally, \citet{markel2023gpteach} and \citet{macina2023mathdial} use LLMs to simulate student dialogues, helping teachers practice and refine their instructional strategies.

\paragraph{Evaluating LLMs on Document Creation.} Some existing works evaluate LLM-generated or edited writing pieces, such as short stories, with professional writers \cite{chakrabarty2024art,subbiah2024reading,marco2024small,chakrabarty2024can}. Beyond single-piece writing, other research studies human–AI collaborative writing. Specifically, \citet{lee2022coauthor} present the first human study on how LLM assistance affects human writing processes, \citet{mcguire2024establishing} show how AI influences users’ creativity during co-creation, \citet{yeh2024ghostwriter} focus on user personalization and control in AI writing systems, and \citet{wan2024felt} assess the role of LLMs during brainstorming phases. \citet{shahid2024examining} analyze the impact of AI-generated or co-written comments in online discussions.

In comparison, our work studies whether LLMs can simulate users effectively for evaluating LLMs. We demonstrate initial success by incorporating user profiles and introduce a benchmark for future research to develop better user simulators.
\section{Annotation Details}
\label{app:annotation_details}

We collect real human and AI conversations using Amazon Mechanical Turk (AMT). We refine our interface design through multiple rounds of pilot studies based on user feedback.

\subsection{Quality Control}
To ensure high-quality work, we require our annotators to meet strict qualifications:
\begin{itemize}[leftmargin=*,itemsep=-0.3em,topsep=-0.2em]
    \item \textbf{Master Worker Status:} Annotators must be Amazon Mechanical Turk Master Workers.
    \item \textbf{Location:} Annotators must be based in the United States, Great Britain, or Australia.
    \item \textbf{Experience:} Annotators must have completed at least 1,000 HITs.
    \item \textbf{Approval Rate:} Annotators must maintain an approval rate of at least 98\%.
\end{itemize}
\vspace{0.7em}
During the annotation process, we release tasks in batches. After each batch, we use GPT-4o to check for low-quality submissions, and the first author reviews the rest annotations to catch any spammers that might have been missed. Any annotations that do not meet our standards are removed, and the workers responsible are barred from future tasks. Overall, only about 6\% of the workers are marked as low quality under these strict criteria.
Each annotator is paid \$7.5 per math tutoring task and \$6.5 per document creation task, which is around \$20 per hour. The total annotation cost is around \$10,000, including pilot study, bonus, and AMT commissions.
All the annotations are collected with consent from the crowdworkers, and no personal information is collected in the data.

\subsection{Interface} Our annotation interfaces are built using Gradio. The math tutoring interface uses Gradio version 4.41.0, and the document creation interface uses version 5.9.1.
We provide screenshots of our interfaces. Figures \ref{fig:math_tutoring_interface_step_1} to \ref{fig:math_tutoring_interface_step_3_1} show the math tutoring interface, while Figures \ref{fig:document_creation_interface_step_1} to \ref{fig:document_creation_interface_step_3_1} show the document creation interface. Both interfaces are hosted on Heroku.

\begin{table*}
\centering
\setlength{\tabcolsep}{2pt}
  \renewcommand{\arraystretch}{1}
  \resizebox{0.95\textwidth}{!}{
\begin{tabular*}{\textwidth}{@{\extracolsep{\fill}}lrrrrrrrr}
\toprule
& \multicolumn{4}{c}{\textbf{Math Tutoring}} & \multicolumn{4}{c}{\textbf{Document Creation}} \\
\cmidrule(lr){2-5}
\cmidrule(lr){6-9}
& \multicolumn{2}{c}{\textit{Statistics}} & \multicolumn{2}{c}{\textit{Evaluation}} & \multicolumn{2}{c}{\textit{Statistics}} & \multicolumn{2}{c}{\textit{Evaluation}} \\
\cmidrule(lr){2-3}
\cmidrule(lr){4-5}
\cmidrule(lr){6-7}
\cmidrule(lr){8-9}
\textbf{Model} & \#Inst. &  \#Turns & Interaction & Correct Rate & \#Inst. & \#Turns & Interaction & Document \\
\midrule
Phi-3 small & 46 & 6.8 & 5.61 & 60.87 & 51 & 7.6 & 7.45 & 7.63 \\
Phi-3 medium & 33 & 4.1 & 5.67 & \textbf{78.79} &  51 & 6.8 & 5.92 & 6.14 \\
Llama 3.1 8B &  50 & 8.2 & 6.70 & 50.00 & 51 & 7.1 & 8.67 & 8.20 \\
Llama 3.1 70B  & 50 & 6.7 & 8.18 & 64.00 & 51 & 6.8 & \textbf{8.82} & 8.37  \\
Mistral Large 2 & 50 & 10.0 & 7.20 & 70.00 & 51 & 6.3 & 8.49 & 8.24 \\
Claude 3.5 Sonnet & 50 & 9.3 & 8.20 & 66.00 & 51 & 7.3 & 8.47 & 8.45 \\
GPT-4o mini & 50 & 10.1 & 7.56 & 64.00 &  51 & 6.8 & 8.63 & 8.41 \\
GPT-4 Turbo & 49 & 6.1 & 7.98 & 65.31 & 51 & 6.5 & 8.53 & \textbf{8.53} \\
GPT-4o  & 49 & 7.0 & \textbf{8.41} & 77.55 & 51 & 6.9 & 8.45 & 8.29 \\
\bottomrule
\end{tabular*}}
\caption{Statistics and human evaluation of various LMs on math tutoring and document creation tasks. For math tutoring, we exclude single-turn conversations, mostly from the Phi-3 models, as they sometimes fail to follow the system prompt and provide solutions on the first turn. \textit{Correct Rate} measures percentage of human users get correct answer after tutoring. Other evaluation ratings are average 1-10 human rating, with 10 being the highest quality.}
\label{tab:model_evaluation}
\vspace{-12pt}
\end{table*}

\subsection{Rating Definitions}
\label{app:rating_definition}

\begin{itemize}[leftmargin=*,itemsep=-0.3em,topsep=-0.3em]
\item \textbf{Score 1–2 (Very Poor):} The assistant's performance is very poor—responses lack clarity and coherence, failing to help the user achieve the intended goal. 
\item \textbf{Score 3–4 (Poor):} The assistant's performance is poor, offering minimal support and contributing little toward the user's goal. 
\item \textbf{Score 5–6 (Average):} The assistant's performance is average; it may contain errors or omit important details, resulting in only partial progress. 
\item \textbf{Score 7–8 (Good):} The assistant's performance is good, providing useful and relevant support that effectively aids the user's task. 
\item \textbf{Score 9–10 (Very Good):} The assistant's performance is very good, delivering clear, comprehensive, and insightful guidance that significantly advances the task.
\end{itemize}
\vspace{6pt}

\subsection{Data Statistics}
\label{app:data_stats}

Table \ref{tab:model_evaluation} displays the data stats for \bn.
We collect 450 math tutoring conversations (50 per model) and 459 document creation conversations (51 per model). For math tutoring, we exclude single-turn conversations where the tutor failed to follow the system prompt by giving away the solution in the first turn. This occurred most frequently with Phi-3 models. After filtering, we have 427 multi-turn math conversations for experiments.
Since users chose the math problems or document topics they wanted to work on, models did not interact with identical sets of tasks. However, we ensured a balanced distribution across scenarios. For math tutoring, problems span 8 level-3, 26 level-4, and 16 level-5 difficulty levels, with a focus on higher-difficulty problems to better engage adult users. For document creation, we include 17 topics for each of the three types: email/letter, blog post, and creative writing.
Note that this dataset is used to evaluate user simulators, not assistant models. For assistant benchmarking, see Section~\ref{sec:assistant_benchmark}, where we use the best user simulator to evaluate 14 assistant models on the same set of problems, document topics, and user profiles.

\section{Implementation Details}

\subsection{Sampling Parameters} For user simulation, we set the temperature to $0.7$, while the AI assistants use a temperature of $0$. For all other tasks, we use a temperature of $0.7$. In every case, the top-p parameter is fixed at $1.0$.

\subsection{API}
For Claude models, we use the 2024-06-20 version of Claude 3.5 Sonnet and the 2025-02-19 version of Claude 3.7 Sonnet.
For Mistral, we use the 2024-07 version of Mistral Large v2.
For Gemini models, we use the 001 version of Gemini 2.0 Flash and the preview-04-17 version of Gemini 2.5 Flash.
For GPT-4o, we use the 2024-05-13 version as the simulator, and the 2024-11-24 version as the rater, for data processing, and as the evaluated assistant in Table~\ref{tab:model_evaluation_simulator}.
Across all experiments---including calls to GPT-4o, Claude, Mistral, Gemini, and other LLMs---the total API cost is around \$50,000 USD.
\begin{figure}[t]
  \centering
    \includegraphics[width=0.90\linewidth]{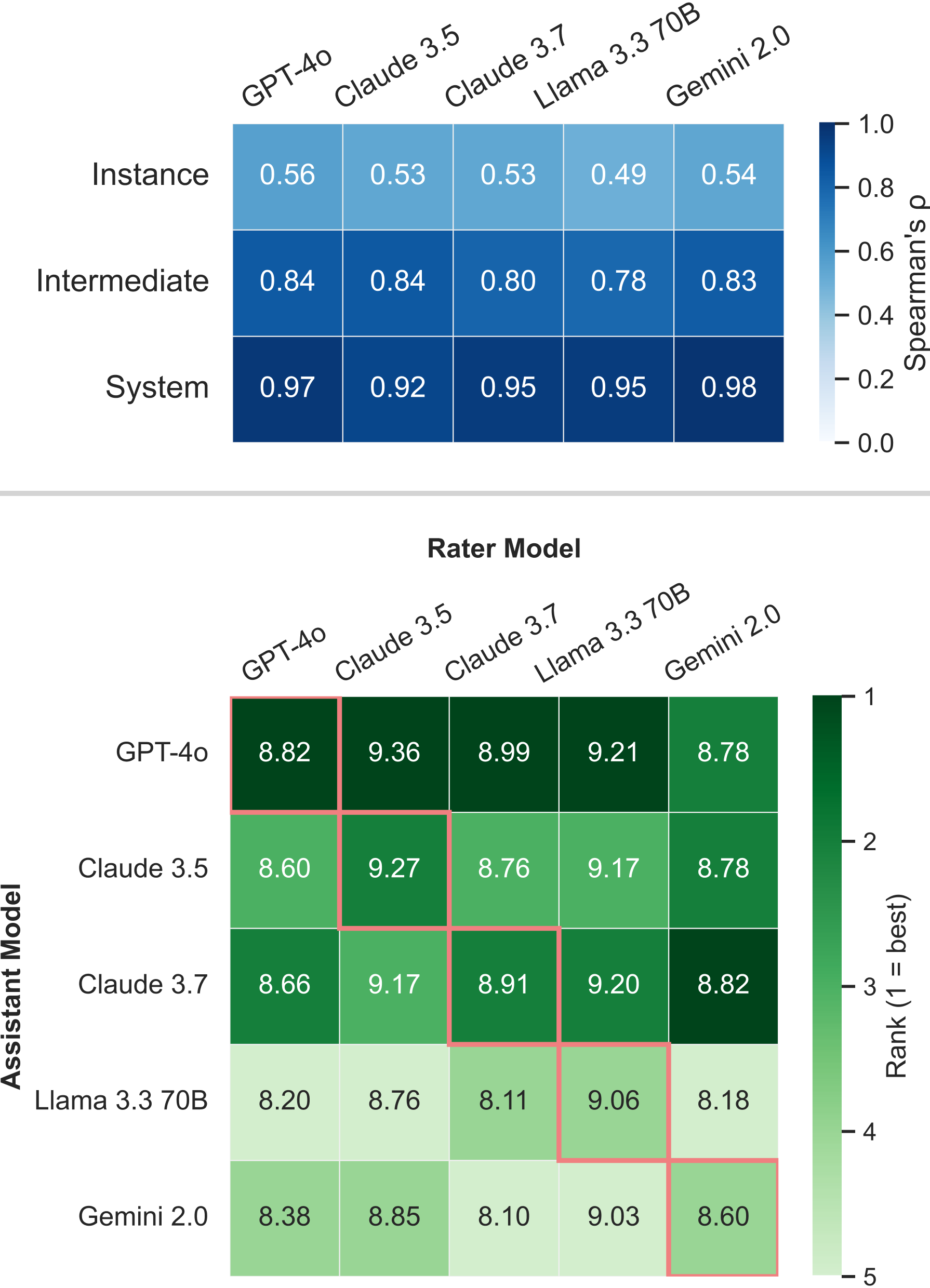}
    \vspace{-5pt}
 \caption{Evaluation of different LLMs as raters ($\pi_r$) for interaction quality in the math tutoring task. The \textbf{top} section shows alignment with human ratings, and the \textbf{bottom} reports average scores to assess self-bias. GPT-4o achieves the highest intermediate-level correlation. Although GPT-4o rates its own assistant the highest, three other raters do the same—indicating that GPT-4o's assistant genuinely performs best. Thus, there is no evidence of self-bias in any of the rater models.}
    \label{fig:rater_evaluation_math_interaction}
    \vspace{-12pt}
\end{figure}

\begin{figure}[t]
  \centering
    \includegraphics[width=0.90\linewidth]{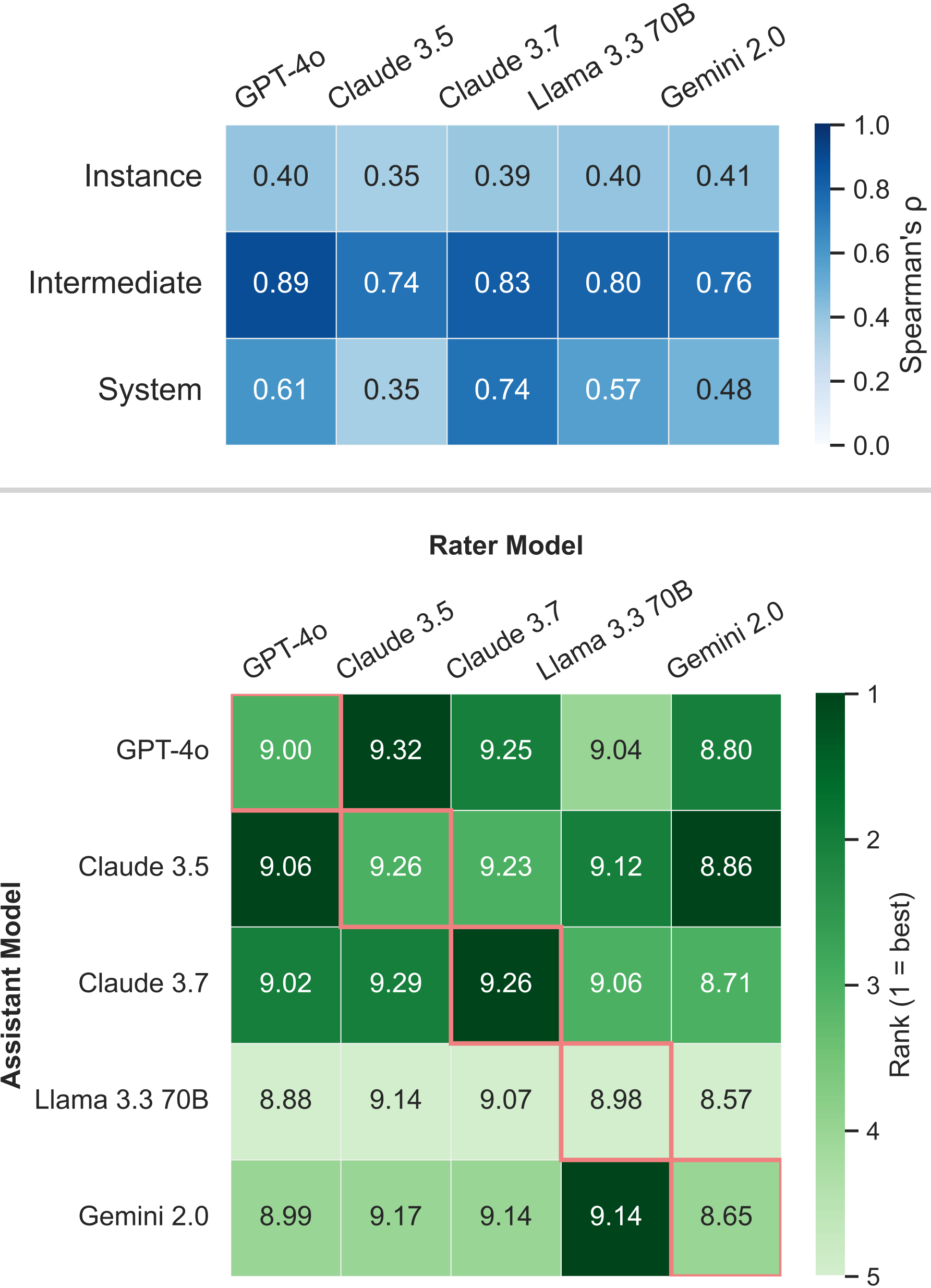}
    \vspace{-5pt}
 \caption{Evaluation of different LLMs as raters ($\pi_r$) for interaction quality in the document creation task. The \textbf{top} section shows alignment with human ratings, and the \textbf{bottom} reports average scores to assess self-bias. GPT-4o achieves the highest intermediate-level correlation, and none of the models show evidence of self-bias.}
    \label{fig:rater_evaluation_doc_interaction}
    \vspace{-12pt}
\end{figure}

\section{Detailed Results}
\label{app:further_results}
\begin{figure*}[t]
  \centering
    \includegraphics[width=0.99\linewidth]{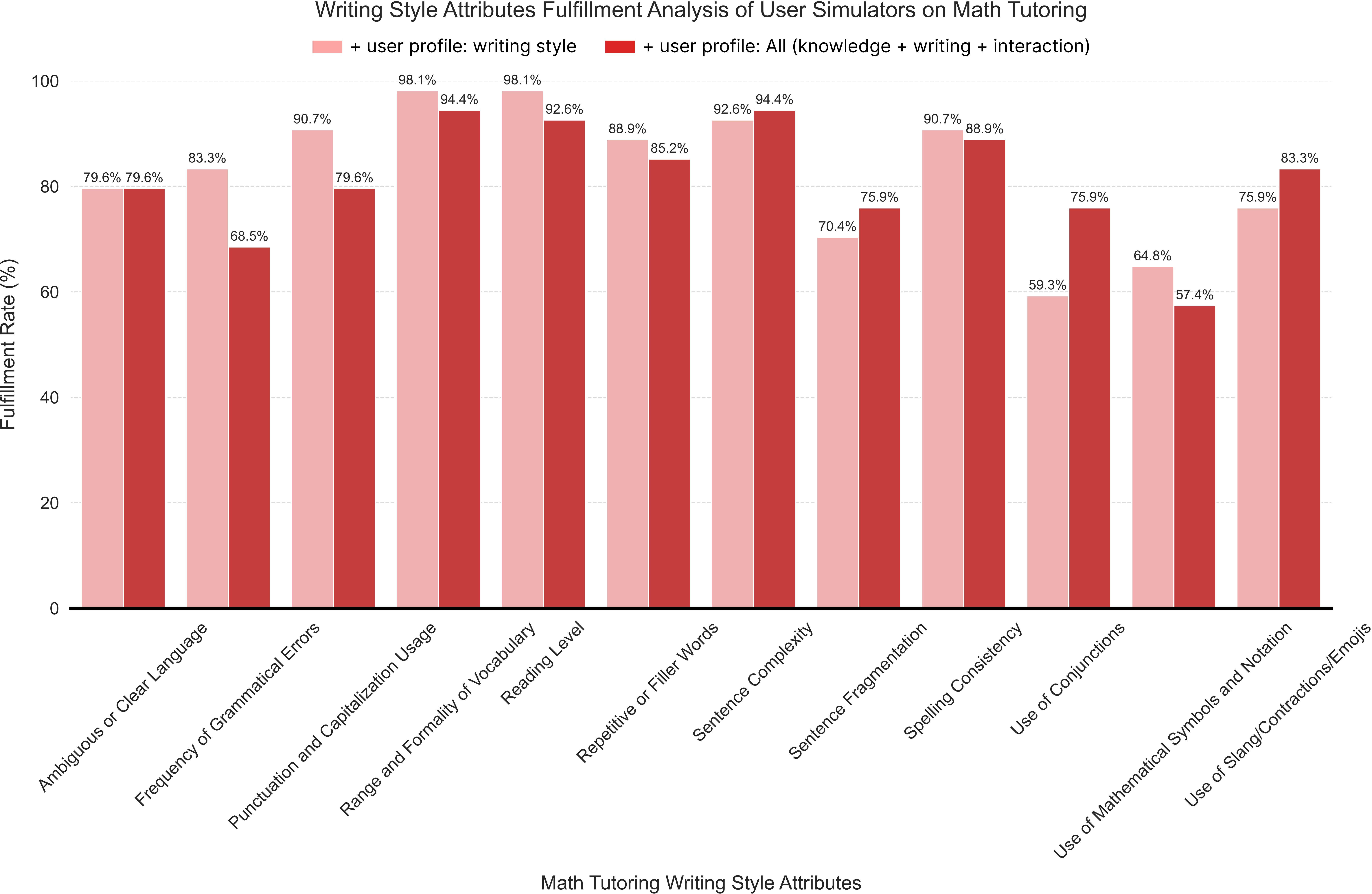}
 \caption{Writing style attributes fulfillment of user simulators on math tutoring.}
    \label{fig:feature_fulfillment_math_writing}
    \vspace{-5pt}
\end{figure*}

\begin{figure*}[t]
  \centering
    \includegraphics[width=0.99\linewidth]{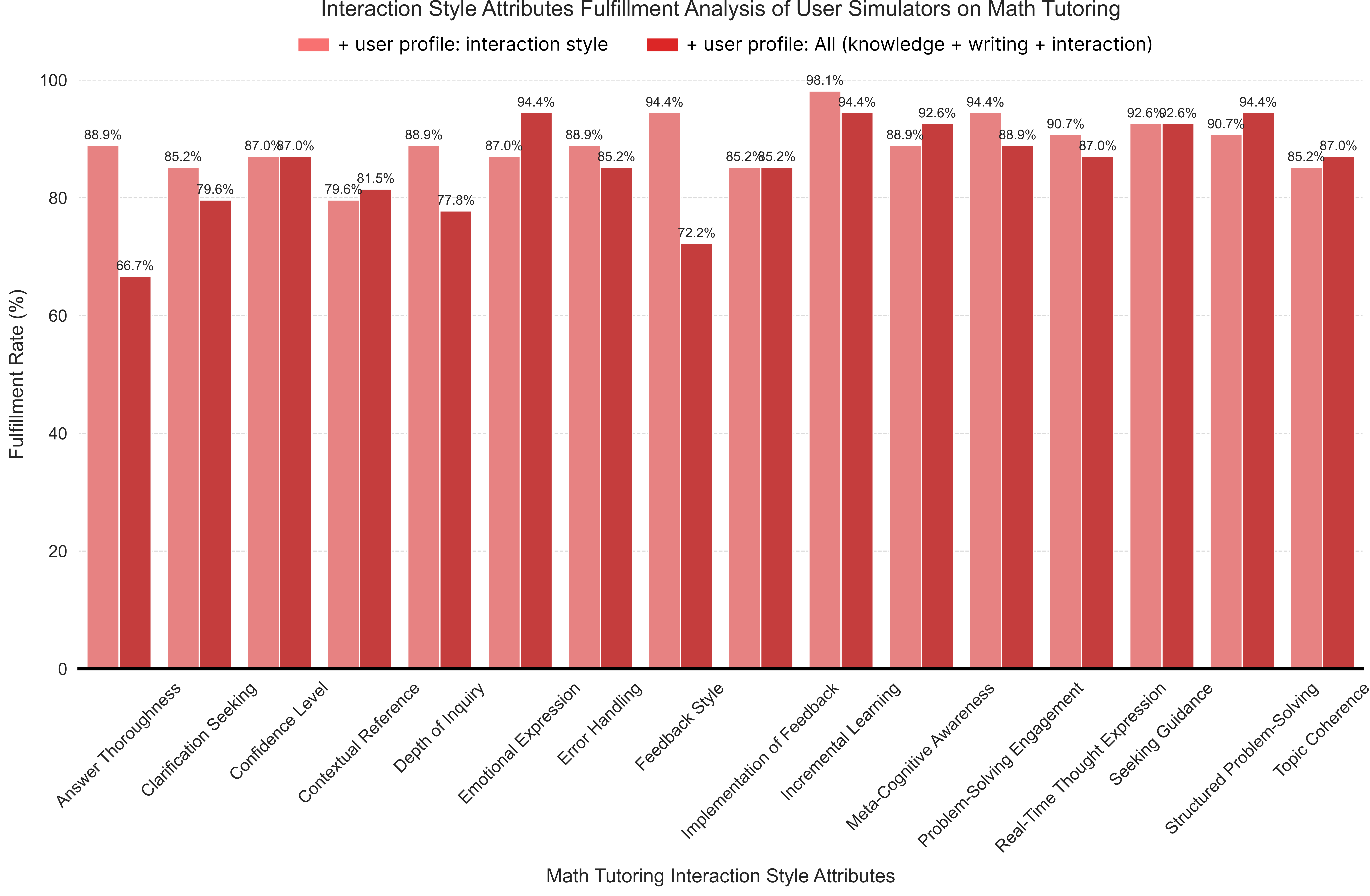}
 \caption{Interaction style attributes fulfillment of user simulators on math tutoring.}
    \label{fig:feature_fulfillment_math_interaction}
    \vspace{-5pt}
\end{figure*}

\begin{figure*}[t]
  \centering
    \includegraphics[width=0.99\linewidth]{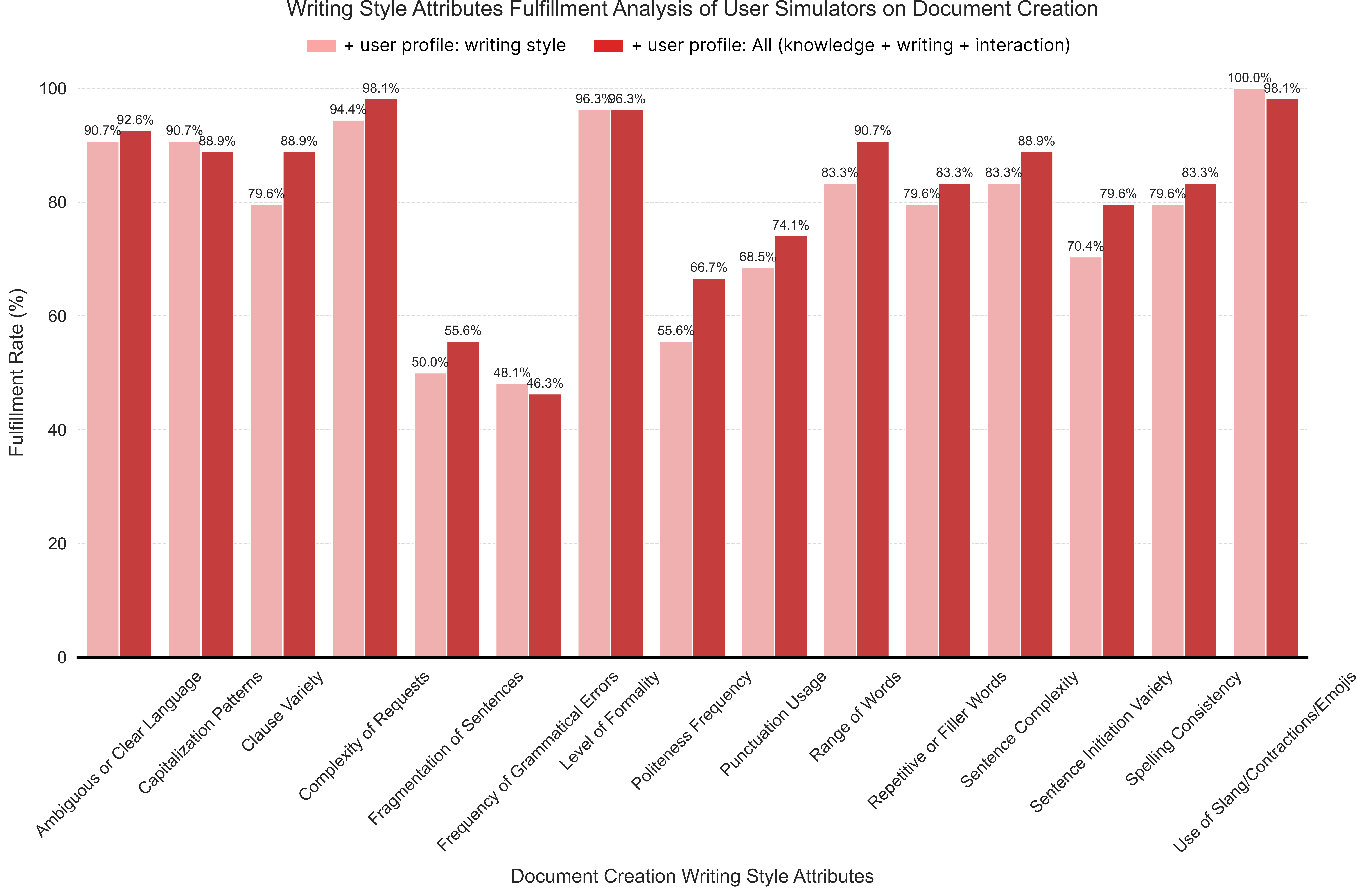}
 \caption{Writing style attributes fulfillment of user simulators on document creation.}
    \label{fig:feature_fulfillment_doc_writing}
    \vspace{-5pt}
\end{figure*}

\begin{figure*}[t]
  \centering
    \includegraphics[width=0.99\linewidth]{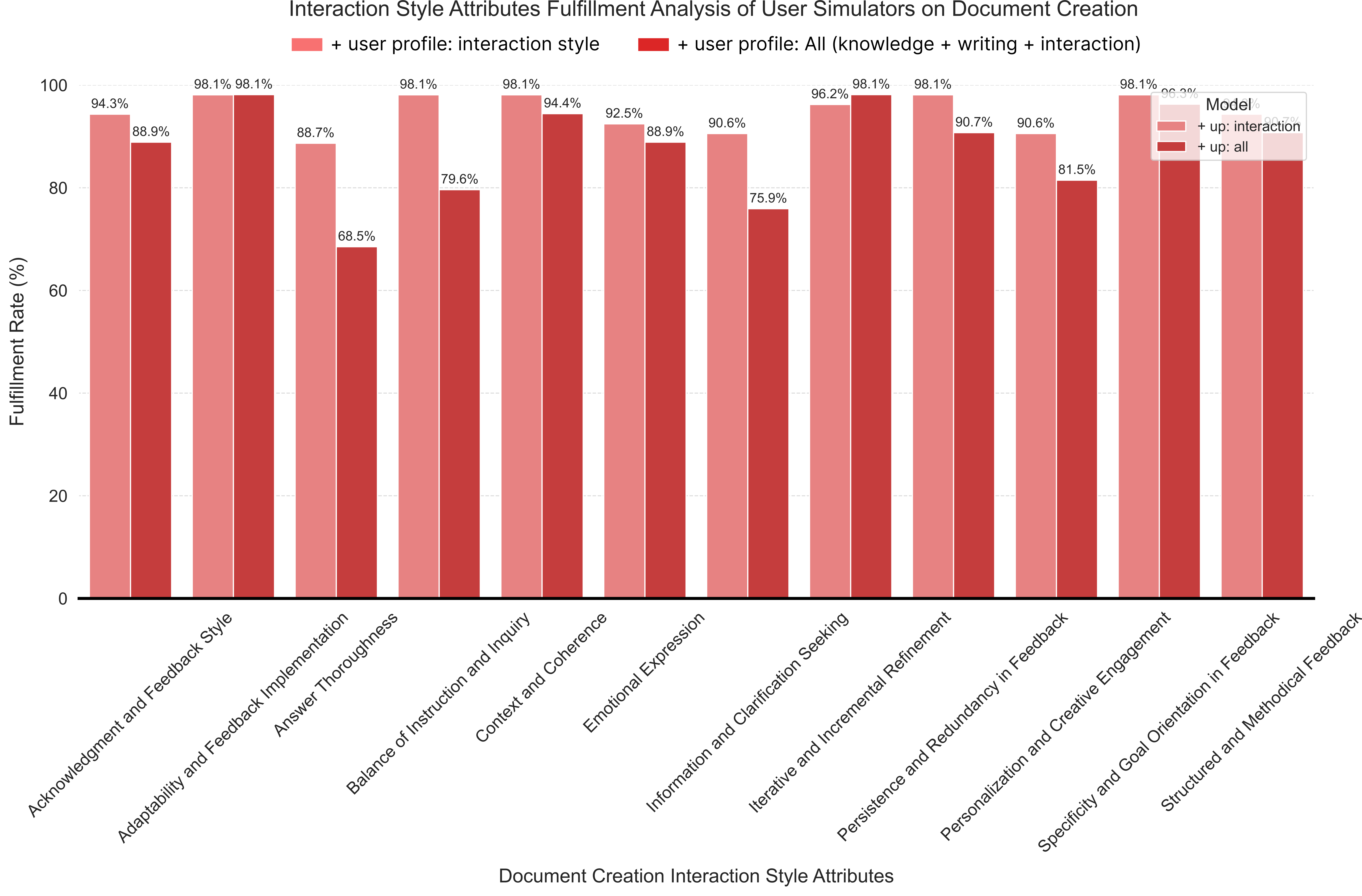}
 \caption{Interaction style attributes fulfillment of user simulators on document creation.}
    \label{fig:feature_fulfillment_doc_interaction}
    \vspace{-5pt}
\end{figure*}

Table~\ref{tab:user_simulator_stats} reports detailed statistics for all user simulator methods, extending Table~\ref{tab:user_simulator_stats_short} from the main text.
Tables~\ref{tab:spearman_correlation}, \ref{tab:pearson_correlation}, and \ref{tab:kendall_correlation} present alignment between user simulator and human ratings on interaction quality across three levels---instance, intermediate, and system---using Spearman’s $\rho$, Pearson’s $r$, and Kendall’s $\tau$, respectively.
Table~\ref{tab:end_outcome_alignment} shows alignment on end outcomes for both tasks. Overall, simulators that use user profiles produce ratings that align more closely with human evaluations than those without.
Table~\ref{tab:simulator_with_different_llms} compares different LLMs---GPT-4o, Claude 3.5 Sonnet, Claude 3.7 Sonnet, Llama 3.3 70B, and Gemini 2.0 Flash---as user simulators. This table extends the analysis from Figure~\ref{fig:extrinsic_eval_user_sim_model}.
Figures~\ref{fig:rater_evaluation_math_interaction} and \ref{fig:rater_evaluation_doc_interaction} present evaluations of different LLMs as raters on the math tutoring and document creation interaction aspects, respectively.
Table~\ref{tab:intrinsic_evaluation} shows message similarity between simulators and real users. The best results are achieved by combining user profiles with length control. Interestingly, higher similarity does not always lead to better alignment with human ratings.
Figures~\ref{fig:feature_fulfillment_math_writing} through \ref{fig:feature_fulfillment_doc_interaction} provide fine-grained analysis of message style attribute fulfillment rates of user simulators, extending Figure~\ref{fig:feature_fulfillment_analysis} from the main paper.
 
\begin{table*}[t]
\centering
\setlength{\tabcolsep}{4pt} 
\renewcommand{\arraystretch}{1.1}
\resizebox{0.95\linewidth}{!}{
\begin{tabular}{lrrrrrrrr}
\toprule
& \multicolumn{4}{c}{\textbf{Math Tutoring}} & \multicolumn{4}{c}{\textbf{Document Creation}} \\
 \cmidrule(lr){2-5}
 \cmidrule(lr){6-9}
& \textit{\#Turns} & \multicolumn{2}{c}{\textit{Message Length}} & \textit{Cost (\$)} & \textit{\#Turns} & \multicolumn{2}{c}{\textit{Message Length}}  & \textit{Cost (\$)}
\\
\cmidrule(lr){3-4}
\cmidrule(lr){7-8}
\textbf{Method}
 &
Per Conv. & User
  & Assistant  & Per Conv. & Per Conv. 
 & User
  & Assistant &  Per Conv. \\
\midrule
\textbf{\textcolor{baseline}{Zero-shot}} &  5.8 & 87.6 & 193.9 & 0.02 & 6.0 & 77.5 & 591.8 & 0.05 \\
\textbf{\textcolor{baseline}{Zero-shot-CoT}} & 7.3 & 89.5 & 293.0 & 0.06 & 8.8 & 123.9 & 761.0 & 0.14 \\
\textbf{\textcolor{length}{+ Length Control}} & 9.7 & 21.4 & 207.2 & 0.06 & 11.9 & 30.5 & 728.7 & 0.17  \\
\textbf{\textcolor{profile}{+ User Profile}} \\
\quad Inherent Knowledge & 9.9 & 96.9 & 377.6 & 0.14 & 8.5 & 137.5 & 751.0 & 0.14 \\
\quad Writing Style &  7.5 & 51.4 & 332.4 & 0.08 &  7.0 & 83.4 & 648.3 & 0.10 \\
\quad Interaction Style & 8.2 & 23.5 & 315.5 & 0.09 & 7.5 & 41.7 & 610.9 & 0.09 \\ 
\quad All & 8.0 & 34.7 & 294.3 & 0.10 & 7.1 & 58.2 & 600.5 & 0.10 \\
\textbf{\textcolor{both}{+ Length Control \& User Profile}} \\
\quad Inherent Knowledge & 10.5 & 21.4 & 280.8 & 0.09 & 11.8 & 34.5 & 750.4 & 0.17
 \\
\quad Writing Style & 9.3 & 18.3 & 297.9 & 0.09 & 9.8 & 27.4 & 614.7 & 0.13  \\
\quad Interaction Style & 8.5 & 18.2 & 317.4 & 0.09 & 8.9 & 28.8 & 609.1 & 0.11 \\ 
\quad All & 9.1 & 18.5 & 291.6 & 0.11 &  8.7 & 28.8 & 622.1 & 0.13 \\
\midrule
\textbf{Human} & 7.8 & 15.5 & 169.6 & 5.33 & 6.9 & 32.6 & 488.0 & 6.50 \\
\bottomrule
\end{tabular}
}
\caption{Conversation statistics for user simulator methods and real human users in math tutoring and document creation tasks. \textit{\#Turns} and \textit{Cost} are averaged per conversation, and \textit{Message Length} per message.  Even the most expensive user simulator method costs less than 3\% of the human cost, and 
 prompt caching can further reduce it.}
\label{tab:user_simulator_stats}
\vspace{-12pt}
\end{table*}

\begin{table*}
\centering
\setlength{\tabcolsep}{4pt}
  \renewcommand{\arraystretch}{1}
  \resizebox{0.90\textwidth}{!}{
\begin{tabular}{@{\extracolsep{\fill}}lcccccc}
\toprule
 & \multicolumn{3}{c}{\textbf{Math Tutoring}}
& \multicolumn{3}{c}{\textbf{Document Creation}}  \\
\cmidrule(lr){2-4} \cmidrule(lr){5-7}
\textbf{Method} & \makecell[c]{\textit{n=427}\\Instance} & \makecell[c]{\textit{n=27}\\Intermediate} & \makecell[c]{\textit{n=9}\\System} & \makecell[c]{\textit{n=459}\\Instance} & \makecell[c]{\textit{n=27}\\Intermediate} & \makecell[c]{\textit{n=9}\\System} \\
\midrule
\textbf{\textcolor{baseline}{Zero-shot}} & 0.238 & 0.550 & 0.900 & 0.250 & 0.263 & \third{0.267}  \\
\textbf{\textcolor{baseline}{Zero-shot-CoT}}  & 0.266 & 0.607 & \first{0.967} & 0.311 & 0.545 & 0.217 \\
\textbf{\textcolor{length}{+ Length Control}} &  0.263 & 0.657 & 0.850 & 0.291 & \first{0.707} & \second{0.367} \\
\textbf{\textcolor{profile}{+ User Profile}} \\
\quad Inherent Knowledge & 0.297 & 0.700 & 0.900 &  \first{0.340} & 0.613 & 0.183 \\
\quad Writing Style & 0.277 & 0.623 & 0.917 & 0.288 & 0.583 & 0.183 \\
\quad Interaction Style & \third{0.358} & \first{0.774} & 0.883 & \second{0.329} & 0.605 & 0.167 \\ 
\quad All & 0.371 & 0.647 &\third{0.933 }& \third{0.326} & \second{0.704} & \second{0.367} \\
\textbf{\textcolor{both}{+ Length Control \& User Profile}} \\
\quad Inherent Knowledge & 0.333 & 0.631 & \second{0.950} &  0.305 & \third{0.700} & \third{0.267} \\
\quad Writing Style & 0.309 & 0.625 & 0.900 & 0.325 & 0.619 & 0.243 \\
\quad Interaction Style  & \second{0.373} & \second{0.753} & 0.900 & 0.308 & 0.642 & \first{0.471} \\
\quad All & \first{0.383} & \third{0.744} & 0.917 &  0.275 & 0.637 & 0.237 \\
\bottomrule
\end{tabular}}
\caption{\textbf{Spearman} correlation between user-simulator and human ratings of AI assistant performance on interaction aspect in math tutoring and document creation tasks at instance, intermediate, and system levels.}
\label{tab:spearman_correlation}
\vspace{-12pt}
\end{table*}

\begin{table}[t]
\centering
\setlength{\tabcolsep}{4pt}
\renewcommand{\arraystretch}{1}
\resizebox{\columnwidth}{!}{%
\begin{tabular}{lrrrr}
\toprule
& \multicolumn{2}{c}{\textbf{Math Tutoring}} & \multicolumn{2}{c}{\textbf{Document Creation}}\\
\cmidrule(lr){2-3}\cmidrule(lr){4-5}
\multirow{2}{*}{\textbf{Simulator LLM}} &
  \makecell{Interaction\\Correlation} &
  \makecell{Outcome\\Alignment} &
  \makecell{Interaction\\Corrleation} &
  \makecell{Outcome\\Alignment} \\
\midrule
GPT-4o               & \first{0.774} & \second{0.666} & 0.704 & \first{0.807}  \\
Claude 3.5 Sonnet    & 0.721 & \first{0.690} & 0.600 & 0.730  \\
Claude 3.7 Sonnet    & 0.646 & 0.655 & \second{0.707} & 0.762 \\
Gemini 2.0 Flash     &  \second{0.727} & 0.647 & \first{0.736} & \second{0.798} \\
Llama 3.3 70B        &  0.718 & 0.652 & 0.640 & 0.671 \\
\bottomrule
\end{tabular}}
\caption{Alignment between different LLMs as user simulators and human evaluations. For math tutoring, the simulator uses user profile with interaction style; for document creation, it uses full profile (knowledge, writing, and interaction style). Values are intermediate-level Spearman correlations, except for math tutoring outcome alignment, which reports Macro F$_1$.}
\label{tab:simulator_with_different_llms}
\vspace{-12pt}
\end{table}

\begin{table*}
\centering
\setlength{\tabcolsep}{4pt}
  \renewcommand{\arraystretch}{1}
  \resizebox{0.9\textwidth}{!}{
\begin{tabular}{@{\extracolsep{\fill}}lcccccc}
\toprule
 & \multicolumn{3}{c}{\textbf{Math Tutoring}}
& \multicolumn{3}{c}{\textbf{Document Creation}}  \\
\cmidrule(lr){2-4} \cmidrule(lr){5-7}
\textbf{Method} & \makecell[c]{\textit{n=427}\\Instance} & \makecell[c]{\textit{n=27}\\Intermediate} & \makecell[c]{\textit{n=9}\\System} & \makecell[c]{\textit{n=459}\\Instance} & \makecell[c]{\textit{n=27}\\Intermediate} & \makecell[c]{\textit{n=9}\\System} \\
\midrule
\textbf{\textcolor{baseline}{Zero-shot}} &  0.253 & 0.587 & 0.860 & 0.226 & 0.800 & 0.864  \\
\textbf{\textcolor{baseline}{Zero-shot-CoT}}  & 0.247 & 0.627 & 0.896 & 0.363 & 0.874 & 0.919 \\
\textbf{\textcolor{length}{+ Length Control}} &  0.236 & 0.600 & 0.862 & \second{0.375} & \first{0.923} & \third{0.953} \\
\textbf{\textcolor{profile}{+ User Profile}} \\
\quad Inherent Knowledge & 0.304 & 0.716 & 0.918 & 0.366 & 0.887 & 0.930 \\
\quad Writing Style & 0.313 & 0.633 & 0.850 & 0.323 & 0.849 & 0.891 \\
\quad Interaction Style & \second{0.396} & \third{0.720} & \first{0.948} & 0.371 & 0.882 & 0.912 \\ 
\quad All & 0.387 & 0.675 & 0.883 & \first{0.395} & 0.881 & 0.931 \\
\textbf{\textcolor{both}{+ Length Control \& User Profile}} \\
\quad Inherent Knowledge & 0.338 & 0.696 & 0.925 & \third{0.372} & \second{0.921} & \first{0.960} \\
\quad Writing Style & 0.312 & 0.631 & 0.840 & 0.358 & \third{0.901} & 0.936 \\
\quad Interaction Style  & \third{0.395} & \second{0.770} & \second{0.942} & 0.334 & 0.879 & \second{0.955} \\
\quad All & \first{0.408} & \first{0.783} & \third{0.941} & 0.349 & 0.842 & 0.930 \\
\bottomrule
\end{tabular}}
\caption{\textbf{Pearson} correlation between user-simulator and human ratings of AI assistant performance on interaction aspect in math tutoring and document creation tasks at instance, intermediate, and system levels.}
\label{tab:pearson_correlation}
\end{table*}

\begin{table*}
\centering
\setlength{\tabcolsep}{4pt}
  \renewcommand{\arraystretch}{1}
  \resizebox{0.9\textwidth}{!}{
\begin{tabular}{@{\extracolsep{\fill}}lcccccc}
\toprule
 & \multicolumn{3}{c}{\textbf{Math Tutoring}}
& \multicolumn{3}{c}{\textbf{Document Creation}}  \\
\cmidrule(lr){2-4} \cmidrule(lr){5-7}
\textbf{Method} & \makecell[c]{\textit{n=427}\\Instance} & \makecell[c]{\textit{n=27}\\Intermediate} & \makecell[c]{\textit{n=9}\\System} & \makecell[c]{\textit{n=459}\\Instance} & \makecell[c]{\textit{n=27}\\Intermediate} & \makecell[c]{\textit{n=9}\\System} \\
\midrule
\textbf{\textcolor{baseline}{Zero-shot}} & 0.187 & 0.418 & \third{0.778} & 0.196 & 0.173 & \third{0.222} \\
\textbf{\textcolor{baseline}{Zero-shot-CoT}}  & 0.207 & 0.468 & \first{0.889} & 0.246 & 0.407 & 0.167  \\
\textbf{\textcolor{length}{+ Length Control}} &  0.200 & 0.479 & 0.667 & 0.227 & \third{0.554} & \second{0.278} \\
\textbf{\textcolor{profile}{+ User Profile}} \\
\quad Inherent Knowledge &  0.227 & 0.529 & \third{0.778} & \first{0.268} & 0.485 & 0.111 \\
\quad Writing Style & 0.209 & 0.469 & \third{0.778} & 0.228 & 0.454 & 0.167 \\
\quad Interaction Style & 0.264 & \first{0.586} & 0.722 & \second{0.262} & 0.485 & 0.056 \\ 
\quad All & \second{0.285} & 0.494 & \second{0.833} & \third{0.258} & \first{0.562} & \second{0.278} \\
\textbf{\textcolor{both}{+ Length Control \& User Profile}} \\
\quad Inherent Knowledge & 0.252 & 0.438 & \first{0.889} & 0.236 & \second{0.558} & \third{0.222}  \\
\quad Writing Style & 0.232 & 0.446 & \third{0.778} & 0.257 & 0.486 & 0.197 \\
\quad Interaction Style & \third{0.277} & \second{0.574} & \third{0.778} & 0.245 & 0.493 & \first{0.343} \\
\quad All & \first{0.286} & \third{0.572} & \third{0.778 }& 0.218 & 0.497 & 0.203 \\
\bottomrule
\end{tabular}}
\caption{\textbf{Kendall's $\tau$} correlation between user-simulator and human ratings of AI assistant performance on interaction aspect in math tutoring and document creation tasks at instance, intermediate, and system levels.}
\label{tab:kendall_correlation}
\end{table*}

\begin{table*}
\centering
\setlength{\tabcolsep}{6pt}
  \renewcommand{\arraystretch}{1}
  \resizebox{0.9\textwidth}{!}{
\begin{tabular}{lcccccc}
\toprule
 & \multicolumn{3}{c}{\textbf{Math Tutoring}} & \multicolumn{3}{c}{\textbf{Document Creation}} \\
\cmidrule(lr){2-4}
\cmidrule(lr){5-7}
\textbf{Method} & \makecell[c]{Correct\\F$_1$} & \makecell[c]{Incorrect\\F$_1$} & \makecell[c]{Macro\\F$_1$}  & \makecell[c]{Instance\\Spear. $\rho$} & \makecell[c]{Intermediate\\Spear. $\rho$} & \makecell[c]{System\\Spear. $\rho$} \\
\midrule
\textbf{\textcolor{baseline}{Zero-shot}} &  \first{0.793} & 0.498 & 0.646 & 0.247 & 0.726 & 0.661 \\
\textbf{\textcolor{baseline}{Zero-shot-CoT}} & 0.760 & 0.458 & 0.609 & 0.277 & 0.701 & 0.571 \\
\textbf{\textcolor{length}{+ Length Control}} &  \third{0.761} & 0.512 & 0.637 & \first{0.324} & 0.744 & 0.717 \\
\textbf{\textcolor{profile}{+ User Profile}} \\
\quad Inherent Knowledge & \second{0.769} & 0.498 & 0.634 & \second{0.303} & \first{0.810} & \third{0.833} \\
\quad Writing Style & 0.747 & 0.525 & 0.636 & 0.148 & 0.740 & 0.633 \\
\quad Interaction Style & 0.756 & \first{0.577}  & \first{0.666} & 0.238 & 0.790 & \second{0.867}  \\ 
\quad All & 0.743 & 0.538 & 0.640 &  \third{0.296} & \second{0.807} & \first{0.895} \\
\textbf{\textcolor{both}{+ Length Control \& User Profile}} \\
\quad Inherent Knowledge & \second{0.769} & 0.552 & \second{0.660} & 0.267 & \third{0.793} & 0.733 \\
\quad Writing Style & 0.746 & 0.565 & 0.655 &  0.182 & 0.607 & 0.700 \\
\quad Interaction Style & 0.741 & \second{0.571} & 0.656 & 0.125 & 0.647 & 0.567 \\
\quad All & 0.747 & \third{0.570} & \third{0.658} & 0.233 & 0.636 & 0.633 \\
\bottomrule
\end{tabular}}
\caption{Evaluating whether the user simulator achieves similar end outcomes to real users. For math tutoring, we compare the correctness of the simulator’s answers with those from real users. For document creation, we measure the Spearman correlation between document ratings from simulated and real-user conversations.}
\label{tab:end_outcome_alignment}
\vspace{-12pt}
\end{table*}

\begin{table*}[t]
\centering
\setlength{\tabcolsep}{6pt}
\renewcommand{\arraystretch}{1.1}
\resizebox{0.99\linewidth}{!}{%
\begin{tabular}{lcccccc}
\toprule
& \multicolumn{3}{c}{\textbf{Math Tutoring}} & \multicolumn{3}{c}{\textbf{Document Creation}} \\
\cmidrule(lr){2-4}
\cmidrule(lr){5-7}
\textbf{Method} & 
\makecell[c]{Writing\\Style}~$\uparrow$ & 
\makecell[c]{Interaction\\Style}~$\uparrow$ &
\makecell[c]{Turing\\Test}~$\downarrow$ &
\makecell[c]{Writing\\Style}~$\uparrow$ & 
\makecell[c]{Interaction\\Style}~$\uparrow$ &
\makecell[c]{Turing\\Test}~$\downarrow$\\
\midrule
\textbf{\textcolor{baseline}{Zero-shot}} & 2.20 & 2.48 & 17.8\% & 2.88 & 3.04 & 34.9\% \\
\textbf{\textcolor{baseline}{Zero-shot-CoT}} & 2.25 & 2.41 & 12.8\% & 2.73 & 2.81 & 38.0\% \\
\textbf{\textcolor{length}{+ Length Control}} & 2.60 & 2.74 & \second{12.1\%} & 2.81 & 2.95 & \second{10.6\%} \\
\textbf{\textcolor{profile}{+ User Profile}} \\
\quad Inherent Knowledge & 2.18 & 2.21 & 19.6\% & 2.65 & 2.88 & 37.3\%\\
\quad Writing Style &  2.58 & 2.52  & 22.6\% & 2.90 & 2.94 & 40.6\% \\
\quad Interaction Style & 2.50 & \first{2.86} & 14.5\% & \second{3.02} & \first{3.18} & 24.3\% \\ 
\quad All & 2.69 & 2.68 & 15.6\% & 2.94 & \second{3.13} & 31.8\% \\
\textbf{\textcolor{both}{+ Length Control \& User Profile}} \\
\quad Inherent Knowledge & 2.38 & 2.69 & 14.7\% & 2.94 & 2.99 & 16.7\%\\
\quad Writing Style & \first{2.80} & \second{2.81} & \first{5.1\%} & \third{3.00} & 2.94 & 13.6\% \\
\quad Interaction Style  & \third{2.65} & 2.77 & \third{11.3\%} & 3.99 & \third{3.07} & \first{6.8\%} \\
\quad All & \second{2.74} & \third{2.79} & 13.6\% & \first{3.06} & 3.06 & \third{11.4\%}   \\
\bottomrule
\end{tabular}%
}
\caption{Evaluating the similarity between simulated and real user messages in math tutoring and document creation tasks across three metrics: \textit{Writing Style} and \textit{Interaction Style} are assessed on a 5-point Likert scale (1=least similar, 5=most similar). \textit{Turing Test} is calculated as $|p - 50|\%$, where p is the percentage of correct identification by an LLM evaluator. A value of $0\%$ indicates indistinguishability (i.e., random chance).} 
\label{tab:intrinsic_evaluation}
\vspace{-12pt}
\end{table*}

\section{User Message Style Definitions}
We define each user's writing and interaction style attributes. We first outline general attributes that are designed by the authors to serve as a starting point. Then, we present each task's attributes, which are automatically generated by GPT-4o through comparisons between the user simulator and real human interactions, and then finalized into a concise list.

\subsection{Starting Attributes}
\paragraph{11 Writing Style Attributes:}
\begin{itemize}[leftmargin=*,itemsep=-0.3em,topsep=-0.2em]
    \item \textbf{Grammatical Accuracy} -- How often does the user break basic grammar rules?
    
    \item \textbf{Sentence Structure} -- Does the user primarily use simple sentences, or do they also use compound and complex structures?
    
    \item \textbf{Spelling} -- Does the user often misspell words or make typos?
    
    \item \textbf{Punctuation} -- How does the user employ punctuation (commas, periods, exclamation/question marks)? Are they often missing or excessive?
    
    \item \textbf{Capitalization} -- Does the user consistently capitalize letters correctly, or do they use all lowercase or randomly?
    
    \item \textbf{Vocabulary Range} -- Does the user stick to basic vocabulary or incorporate a broader lexicon?
    
    \item \textbf{Word Repetition} -- Does the user rely heavily on certain filler terms (e.g., 'um,' 'uh'), or repeat the same words/phrases often?
    
    \item \textbf{Formality Level} -- Is the user's language generally formal, casual, or somewhere in between?
    
    \item \textbf{Casual Elements} -- Does the user employ slang, contractions (e.g., 'can't,' 'it's'), emoticons, or emojis, and how frequently?
    
    \item \textbf{Clarity} -- Is the user's query or statement easy to interpret, or does it contain incomplete/ambiguous phrasing?
    
    \item \textbf{Language Complexity} -- Would you estimate the user's writing is at a basic, intermediate, or advanced reading level?
\end{itemize}

\paragraph{10 Interaction Style Attributes:}
\begin{itemize}[leftmargin=*,itemsep=-0.3em,topsep=-0.2em]
    \item \textbf{Message Length} -- What is the range of the length of the user's queries?
    
    \item \textbf{Response Detail} -- When responding to questions, does the user tend to give detailed, complete responses or brief, minimal answers?
    
    \item \textbf{Question Asking} -- How frequently does the user ask for examples or additional explanations?
    
    \item \textbf{Understanding Signals} -- How does the user acknowledge understanding or receipt of information?
    
    \item \textbf{Ongoing Communication} -- Does the user provide ongoing verbal feedback ('okay', 'I see', etc.)?
    
    \item \textbf{Topic Focus} -- How well does the user maintain focus on a single topic versus jumping between different points?
    
    \item \textbf{Context Awareness} -- Does the user refer back to previous messages or maintain conversation context?
    
    \item \textbf{Feedback Response} -- How does the user respond to and implement suggestions or corrections?
    
    \item \textbf{Certainty Expression} -- How does the user express their confidence or uncertainty in their understanding?
    
    \item \textbf{Emotional Expression} -- How does the user communicate their emotional state (frustration, enthusiasm, etc.)?
\end{itemize}

\subsection{Math Tutoring Attributes}
\paragraph{12 Writing Style Attributes:}
\begin{itemize}[leftmargin=*,itemsep=-0.3em,topsep=-0.2em]
\item \textbf{Frequency of Grammatical Errors} -- How often does the user break basic grammar rules?

\item \textbf{Sentence Complexity} -- Does the user primarily use simple sentences, or do they also use compound and complex structures?

\item \textbf{Spelling Consistency} -- Does the user often misspell words or make typos, including mathematical terms?

\item \textbf{Punctuation and Capitalization Usage} -- How does the user employ punctuation (commas, periods, exclamation/question marks) and capitalization? Are they often missing or excessive?

\item \textbf{Range and Formality of Vocabulary} -- Does the user stick to basic vocabulary or incorporate a broader lexicon, including formal mathematical terminology?

\item \textbf{Repetitive or Filler Words} -- Does the user rely heavily on certain filler terms (e.g., 'um,' 'uh'), or repeat the same words/phrases often?

\item \textbf{Ambiguous or Clear Language} -- Is the user's query or statement easy to interpret, or does it contain incomplete/ambiguous phrasing?

\item \textbf{Reading Level} -- Would you estimate the user's writing is at a basic, intermediate, or advanced reading level (e.g., simple everyday language vs. academic/technical jargon)?

\item \textbf{Use of Mathematical Symbols and Notation} -- Does the user incorporate mathematical symbols and notation in LaTeX correctly and frequently in their queries?

\item \textbf{Sentence Fragmentation} -- Does the user tend to use fragmented sentences, often breaking up their thought process into shorter, separate queries?

\item \textbf{Use of Conjunctions} -- How frequently does the user employ conjunctions (e.g., 'and', 'but', 'or') to connect ideas?

\item \textbf{Use of Slang/Contractions/Emojis} -- Does the user employ slang, contractions (e.g., 'can't,' 'it's'), emoticons, or emojis, and how frequently?
\end{itemize}

\paragraph{17 Interaction Style Attributes:}
\begin{itemize}[leftmargin=*,itemsep=-0.3em,topsep=-0.2em]
\item \textbf{Message Length} -- What is the range of the length of the user's queries?

\item \textbf{Answer Thoroughness} -- When responding to questions, does the user tend to give detailed, complete responses or brief, minimal answers?

\item \textbf{Clarification Seeking} -- How frequently does the user ask for examples or additional explanations?

\item \textbf{Feedback Style} -- Does the user provide ongoing verbal feedback ('okay', 'I see', etc.)?

\item \textbf{Topic Coherence} -- How well does the user maintain focus on a single topic versus jumping between different points?

\item \textbf{Contextual Reference} -- Does the user refer back to previous messages or maintain conversation context?

\item \textbf{Implementation of Feedback} -- How does the user respond to and implement suggestions or corrections?

\item \textbf{Confidence Level} -- How does the user express their confidence or uncertainty in their understanding?

\item \textbf{Emotional Expression} -- How does the user communicate their emotional state (frustration, enthusiasm, etc.)?

\item \textbf{Real-Time Thought Expression} -- How does the user articulate their thought process in real-time, reflecting their immediate understanding and confusion?

\item \textbf{Error Handling} -- How does the user demonstrate a trial and error approach, acknowledge, and correct their mistakes?

\item \textbf{Structured Problem-Solving} -- Does the user follow a highly structured approach to problem-solving, with clear delineation of each step?

\item \textbf{Problem-Solving Engagement} -- How does the user engage with the problem-solving process in terms of asking for details about the method or process?

\item \textbf{Seeking Guidance} -- How consistently does the user seek guidance or validation at each step of the problem-solving process?

\item \textbf{Incremental Learning} -- Does the user show a pattern of gradual understanding, requiring multiple examples or steps to grasp a concept?

\item \textbf{Depth of Inquiry} -- How frequently does the user ask for detailed explanations and clarifications?

\item \textbf{Meta-Cognitive Awareness} -- Does the user exhibit awareness of their learning process by summarizing or reflecting on the steps?
\end{itemize}

\subsection{Document Creation Attributes}
\paragraph{15 Writing Style Attributes:}
\begin{itemize}[leftmargin=*,itemsep=-0.3em,topsep=-0.2em]
    \item \textbf{Frequency of Grammatical Errors} -- How often does the user break basic grammar rules?
    
    \item \textbf{Sentence Complexity} -- Does the user primarily use simple sentences, or do they also use compound and complex structures?
    
    \item \textbf{Spelling Consistency} -- Does the user often misspell words or make typos?
    
    \item \textbf{Punctuation Usage} -- How does the user employ punctuation (commas, periods, exclamation/question marks)? Are they often missing or excessive?
    
    \item \textbf{Capitalization Patterns} -- Does the user consistently capitalize letters correctly, or do they use all lowercase or randomly?
    
    \item \textbf{Range of Words} -- Does the user stick to basic vocabulary or incorporate a broader lexicon?
    
    \item \textbf{Repetitive or Filler Words} -- Does the user rely heavily on certain filler terms (e.g., 'um,' 'uh'), or repeat the same words/phrases often?
    
    \item \textbf{Level of Formality} -- Is the user's language generally formal, casual, or somewhere in between?
    
    \item \textbf{Use of Slang/Contractions/Emojis} -- Does the user employ slang, contractions (e.g., 'can't,' 'it's'), emoticons, or emojis, and how frequently?
    
    \item \textbf{Ambiguous or Clear Language} -- Is the user's message easy to interpret, or does it contain incomplete/ambiguous phrasing?
    
    \item \textbf{Fragmentation of Sentences} -- Does the user often use fragmented sentences or complete sentences with clear structure?
    
    \item \textbf{Complexity of Requests} -- How complex are the user's requests? Do they often involve multiple steps or detailed instructions?
    
    \item \textbf{Clause Variety} -- How varied are the user's clauses within sentences? Do they use a mix of independent, dependent, and introductory clauses?
    
    \item \textbf{Politeness Frequency} -- How frequently does the user use politeness markers (e.g., 'thank you,' 'please,' 'could you') in their queries?
    
    \item \textbf{Sentence Initiation Variety} -- Does the user start sentences in various ways, or do they follow a repetitive pattern?
\end{itemize}

\paragraph{13 Interaction Style Attributes:}
\begin{itemize}[leftmargin=*,itemsep=-0.3em,topsep=-0.2em]
    \item \textbf{Message Length} -- What is the range of the length of the user's queries?

    \item \textbf{Answer Thoroughness} -- When responding to questions, does the user tend to give detailed, complete responses or brief, minimal answers?
    
    \item \textbf{Information and Clarification Seeking} -- How frequently does the user ask for examples, additional explanations, or specific metrics?
    
    \item \textbf{Acknowledgment and Feedback Style} -- How does the user acknowledge understanding or receipt of information? Does the user provide ongoing verbal feedback ('okay', 'I see', etc.)?
    
    \item \textbf{Context and Coherence} -- Does the user refer back to previous messages or maintain conversation context and coherence?
    
    \item \textbf{Adaptability and Feedback Implementation} -- How does the user respond to and implement suggestions or corrections? How adaptable is the user's feedback based on the AI's responses?
    
    \item \textbf{Emotional Expression} -- How does the user communicate their emotional state (frustration, enthusiasm, etc.) and exhibit a range of emotional expressions?
    
    \item \textbf{Persistence and Redundancy in Feedback} -- Does the user request the same type of feedback repeatedly without significant changes in their approach?
    
    \item \textbf{Personalization and Creative Engagement} -- Does the user incorporate personal insights, creative suggestions, and specific experiences into their message?
    
    \item \textbf{Iterative and Incremental Refinement} -- Does the user refine the content incrementally, making small adjustments over multiple turns?
    
    \item \textbf{Specificity and Goal Orientation in Feedback} -- How specific and goal-oriented are the user's feedback and modification requests?
    
    \item \textbf{Balance of Instruction and Inquiry} -- Does the user balance between giving specific instructions and asking for suggestions or ideas from the AI?
    
    \item \textbf{Structured and Methodical Feedback} -- Does the user follow a highly structured and methodical approach in their feedback?
\end{itemize}
\section{Conversation Examples}
\label{app:examples}

We present side-by-side examples of simulator–AI and human–AI conversations. For math tutoring, Figure \ref{prompt:math_tutoring_example_zs_cot_1}, \ref{prompt:math_tutoring_example_zs_cot_2} and \ref{prompt:math_tutoring_example_zs_cot_3}  show conversations from the vanilla zero-shot CoT simulator. Figure \ref{prompt:math_tutoring_example_zs_up_undergrad_1}, \ref{prompt:math_tutoring_example_zs_up_undergrad_2},
\ref{prompt:math_tutoring_example_zs_up_undergrad_3},
\ref{prompt:math_tutoring_example_zs_up_undergrad_4} show conversations from the user-profile simulator conditioned on interaction style.

For document creation, Figure \ref{prompt:document_creation_example_zs_cot_1} and \ref{prompt:document_creation_example_zs_cot_2} show conversations from the zero-shot CoT simulator. Figure \ref{prompt:document_creation_example_zs_user_profile_1}
 and \ref{prompt:document_creation_example_zs_user_profile_2}  show conversations from the user-profile simulator with a full profile (including document preferences, writing style, and interaction style).
 
The math tutoring examples correspond to the conversations shown in Figure~\ref{fig:examples_and_scatter} in the main text For document creation, due to the conversation's length, we include only the user messages and the final assistant message.
\section{Prompts}
\label{app:prompts}

We presents all the prompts used in this work. Below is the navigation menu for the prompts:

\subsection{User Simulation Prompts}
\label{app:user_simulation_prompts}

\begin{itemize}[leftmargin=*,itemsep=-0.3em,topsep=-0.2em]
\item Figures \ref{prompt:zero_shot_initial_math} to \ref{prompt:zero_shot_document} show the Zero-shot user simulation prompts.
\item Figures \ref{prompt:zero_shot_cot_initial_math} to \ref{prompt:zero_shot_cot_document} display the Zero-shot CoT user simulation prompts.  
\item Figures \ref{prompt:zero_shot_cot_user_profile_initial_math} to \ref{prompt:zero_shot_cot_user_profile_document} illustrate the Zero-shot CoT with User Profile prompts, while Figures \ref{prompt:zero_shot_length_control_initial_math} to \ref{prompt:zero_shot_length_control_document} show the special prompts for length control user simulation, focusing only on the message length attribute.
\item Figures \ref{prompt:two_stage_refinement_math} and \ref{prompt:two_stage_refinement_document} present the Two-stage Message Style Refinement prompts, which refine the simulated messages based on message style.
\end{itemize}

\subsection{Data Processing Prompts}
\label{app:data_processing_prompts}

\begin{itemize}[leftmargin=*,itemsep=-0.3em,topsep=-0.2em]
\item 
Figures \ref{prompt:extract_writing_style_features} to \ref{prompt:extract_interaction_style_values} present prompts that extract message style attributes (writing and interaction style), finalize them, and retrieve attribute values from the user conversation.  
\item Figure \ref{prompt:extract_simulator_answer} presents the prompt that extracts the simulator’s answer from a conversation. 

\item Figure \ref{prompt:extract_math_concepts} presents the prompt that identifies concepts needed to solve a math problem based on the provided solution. Following this, Figure \ref{prompt:extract_understanding_state} presents the prompt that determines the user's understanding state based on the extracted concepts.
\item Figures \ref{prompt:extract_background} to \ref{prompt:extract_preference_values} present prompts that extract background information and document preferences from pre-writing materials and conversations for the document creation task. Additionally, Figure \ref{prompt:extract_document} presents the prompt that extracts the final document from the conversation.  
\item Finally, Figure \ref{prompt:terminate_conversation} presents the prompt that terminates a conversation that is stuck in a loop or when the user is satisfied.  
\end{itemize}

In the prompts, we use the word “feature” to refer to an “attribute.”

\subsection{Evaluation Prompts}
\label{app:evaluation_prompts}

\begin{itemize}[leftmargin=*,itemsep=-0.3em,topsep=-0.2em]
\item 
Figure \ref{prompt:math_tutor_performance} presents the prompt that evaluates the math tutor's performance on a scale from 1 to 10. Similarly, Figure \ref{prompt:writing_assistant_performance} presents the prompt that evaluates the writing assistant's performance, and Figure \ref{prompt:final_document_evaluation} presents the prompt that evaluates the final document on a scale from 1 to 10.  
\item Figure \ref{prompt:math_answer_correctness} presents the prompt that checks the simulator's answer correctness against the gold answer.  
\item Figures \ref{prompt:turing_test} to \ref{prompt:interaction_style_similarity} present prompts that measure the similarity between simulated messages and human messages, using evaluations ranging from Turing test to 1–5 Likert scale for writing style and interaction style.
\item Figure \ref{prompt:attribute_fulfillment}
 presents the prompt that evaluate whether a given message style attribute is fulfilled by the user simulator in the conversation.
 \end{itemize}

\begin{figure*}[htbp]
  \centering
  \begin{minipage}{0.48\textwidth}
  \begin{AIbox}[width=\textwidth]{Math Tutoring Simulator-AI Conversation}
 \scriptsize   \begin{Verbatim}[breaklines=true,formatcom=\normalfont\rmfamily]
Math Problem: The harmonic mean of two positive integers is the reciprocal of the arithmetic mean of their reciprocals. For how many ordered pairs (x,y) of positive integers is the harmonic mean of x and y equal to 20?

User Turn 1: I understand that the harmonic mean of two numbers x and y is given by the formula H = 2xy / (x + y). I’m trying to set up the equation to find pairs (x, y) where the harmonic mean equals 20, but I’m getting confused with the algebra part. Could you help me set up the right equation and maybe give me a hint on how to solve it?

AI Tutor Turn 1:  Absolutely, I'd be happy to help guide you through this! You correctly mentioned that the harmonic mean \( H \) of two numbers \( x \) and \( y \) is given by the formula: \[ H = \frac{2xy}{x + y} \] You are given that the harmonic mean \( H \) is 20, so you can set up the equation: \[ \frac{2xy}{x + y} = 20 \] To clear the fraction, you can multiply both sides by \( x + y \): \[ 2xy = 20(x + y) \] Now, simplify and rearrange the equation: \[ 2xy = 20x + 20y \] \[ 2xy - 20x - 20y = 0 \] To make it easier to factor or manipulate, you can rewrite it as: \[ xy - 10x - 10y = 0 \] Now, let's try to factor this equation. A helpful technique might be to add a term on both sides that allows you to factor by grouping. Can you think of a term that would complete the square or make the expression factorable?

User Turn 2: I'm not sure what term to add to both sides to make the equation factorable. Could you please show me an example of how to do this or guide me through this specific step?

AI Tutor Turn 2: Certainly! To make the equation factorable, we can add a constant term to both sides that allows us to use the factoring technique. Let's look at the equation: \[ xy - 10x - 10y = 0 \] We can add a constant \( c \) to both sides: \[ xy - 10x - 10y + c = c \] We want to choose \( c \) such that the left-hand side can be factored as a product of two binomials. Let's try to factor it as: \[ (x - a)(y - b) \] Expanding \( (x - a)(y - b) \) gives: \[ xy - bx - ay + ab \] We want this to match \( xy - 10x - 10y + c \). Comparing coefficients, we see: - \( b + a = 10 \) - \( ab = c \) A common technique is to add the square of half the coefficient of \( x \) and \( y \) from the linear terms. Here, the coefficients of \( x \) and \( y \) are both -10, so half of that is -5, and squaring -5 gives 25. Let's add 25 to both sides: \[ xy - 10x - 10y + 25 = 25 \] Now, see if you can factor the left-hand side: \[ (x - 5)(y - 5) = 25 \] This equation is much easier to solve. Can you proceed from here to find the values of \( x \) and \( y \)?

    \end{Verbatim}
  \end{AIbox}
  \end{minipage}
  \hfill
  \begin{minipage}{0.48\textwidth}
  \begin{AIbox}[width=\textwidth]{Math Tutoring Human-AI Conversation}
 \scriptsize   \begin{Verbatim}[breaklines=true,formatcom=\normalfont\rmfamily]

Math Problem: The harmonic mean of two positive integers is the reciprocal of the arithmetic mean of their reciprocals. For how many ordered pairs (x,y) of positive integers is the harmonic mean of x and y equal to 20?

User Turn 1: Hello, what is the definition of harmonic mean?

AI Tutor Turn 1: Hello! The harmonic mean of two numbers, say \(x\) and \(y\), is defined as the reciprocal of the arithmetic mean of their reciprocals. Mathematically, it can be expressed as: \[ \text{Harmonic Mean} = \frac{2}{\frac{1}{x} + \frac{1}{y}} \] This formula calculates the harmonic mean by first finding the reciprocals of \(x\) and \(y\), averaging those reciprocals, and then taking the reciprocal of that average. Would you like to proceed with how to apply this definition to find the ordered pairs \((x, y)\) where the harmonic mean is 20?

User Turn 2: what is the reciprocal of the arithmetic mean?

AI Tutor Turn 2: The reciprocal of the arithmetic mean of two numbers, say \( \frac{1}{x} \) and \( \frac{1}{y} \), is calculated as follows: 1. First, compute the arithmetic mean of \( \frac{1}{x} \) and \( \frac{1}{y} \): \[ \text{Arithmetic Mean} = \frac{\frac{1}{x} + \frac{1}{y}}{2} \] 2. Then, take the reciprocal of this arithmetic mean: \[ \text{Reciprocal of Arithmetic Mean} = \frac{1}{\frac{\frac{1}{x} + \frac{1}{y}}{2}} \] This reciprocal of the arithmetic mean is what defines the harmonic mean when dealing with two numbers. Would you like to see how this applies to finding when the harmonic mean equals 20?

User Turn 3: Yes, please.

AI Tutor Turn 3: Great! To find when the harmonic mean of two numbers \(x\) and \(y\) equals 20, we start with the formula for the harmonic mean: \[ \frac{2}{\frac{1}{x} + \frac{1}{y}} = 20 \] First, let's solve for \(\frac{1}{x} + \frac{1}{y}\): 1. Multiply both sides by \(\frac{1}{x} + \frac{1}{y}\) to clear the denominator: \[ 2 = 20 \left(\frac{1}{x} + \frac{1}{y}\right) \] 2. Divide both sides by 20 to isolate \(\frac{1}{x} + \frac{1}{y}\): \[ \frac{1}{x} + \frac{1}{y} = \frac{2}{20} = \frac{1}{10} \] Now, we know that: \[ \frac{1}{x} + \frac{1}{y} = \frac{1}{10} \] To find the ordered pairs \((x, y)\) that satisfy this equation, we can think about the possible values of \(x\) and \(y\) that make their reciprocals sum to \(\frac{1}{10}\). Can you think of any values of \(x\) and \(y\) whose reciprocals would add up to \(\frac{1}{10}\)?

    \end{Verbatim}
  \end{AIbox}
  \end{minipage}
  \vspace{-5pt}
  \caption{Math tutoring example conversation, the user simulator is the vanilla zero-shot CoT, the assistant is \texttt{GPT-4 Turbo}. [1/3]}
  \label{prompt:math_tutoring_example_zs_cot_1}
\end{figure*}

\clearpage
\begin{figure*}[htbp]
  \centering
  \begin{minipage}{0.48\textwidth}
  \begin{AIbox}[width=\textwidth]{Math Tutoring Simulator-AI Conversation}
 \scriptsize   \begin{Verbatim}[breaklines=true,formatcom=\normalfont\rmfamily]
Math Problem: The harmonic mean of two positive integers is the reciprocal of the arithmetic mean of their reciprocals. For how many ordered pairs (x,y) of positive integers is the harmonic mean of x and y equal to 20?

User Turn 3: Okay, I see that \((x - 5)(y - 5) = 25\) is easier to work with. So, I need to find pairs of integers \(a\) and \(b\) such that \((a - 5)(b - 5) = 25\). I think I need help figuring out all the integer pairs that multiply to 25. Can you help me list them out?


AI Tutor Turn 3: Certainly! To find integer pairs \((a, b)\) such that \(ab = 25\), we can list all the factor pairs of 25. Remember, since \(a\) and \(b\) can be positive or negative, we should consider both positive and negative factors. Here are the factor pairs: 1. \(a = 1\), \(b = 25\) 2. \(a = 25\), \(b = 1\) 3. \(a = -1\), \(b = -25\) 4. \(a = -25\), \(b = -1\) 5. \(a = 5\), \(b = 5\) 6. \(a = -5\), \(b = -5\) Now, recall that these pairs \((a, b)\) are actually \((x-5, y-5)\) from your factored equation \((x-5)(y-5) = 25\). To find the values of \(x\) and \(y\), you need to add 5 to each element of the pairs. Can you calculate the corresponding values of \(x\) and \(y\) for each pair?

User Turn 4: Okay, I will add 5 to each of the factor pairs. Here’s what I get: 1. For \((1, 25)\), \(x = 1 + 5 = 6\), \(y = 25 + 5 = 30\) 2. For \((25, 1)\), \(x = 25 + 5 = 30\), \(y = 1 + 5 = 6\) 3. For \((-1, -25)\), \(x = -1 + 5 = 4\), \(y = -25 + 5 = -20\) 4. For \((-25, -1)\), \(x = -25 + 5 = -20\), \(y = -1 + 5 = 4\) 5. For \((5, 5)\), \(x = 5 + 5 = 10\), \(y = 5 + 5 = 10\) 6. For \((-5, -5)\), \(x = -5 + 5 = 0\), \(y = -5 + 5 = 0\) It looks like some of these pairs result in negative or zero values, which we can’t use since \(x\) and \(y\) are supposed to be positive integers. So, we’re left with: - \((6, 30)\) - \((30, 6)\) - \((10, 10)\) Does this look correct? Are these the only valid pairs for \(x\) and \(y\)?

AI Tutor Turn 4: Yes, you've done that perfectly! You correctly identified and excluded the pairs where either \(x\) or \(y\) is not a positive integer. The valid pairs of positive integers \((x, y)\) that satisfy the condition where the harmonic mean is 20 are indeed: - \((6, 30)\) - \((30, 6)\) - \((10, 10)\) These are the only valid pairs given the constraints of the problem (positive integers). You've successfully found all the ordered pairs where the harmonic mean of \(x\) and \(y\) equals 20. Great job!

    \end{Verbatim}
  \end{AIbox}
  \end{minipage}
  \hfill
  \begin{minipage}{0.48\textwidth}
  \begin{AIbox}[width=\textwidth]{Math Tutoring Human-AI Conversation}
 \scriptsize   \begin{Verbatim}[breaklines=true,formatcom=\normalfont\rmfamily]

Math Problem: The harmonic mean of two positive integers is the reciprocal of the arithmetic mean of their reciprocals. For how many ordered pairs (x,y) of positive integers is the harmonic mean of x and y equal to 20?

User Turn 4: what is a reciprocal in this context?

AI Tutor Turn 4: In this context, the reciprocal of a number \(x\) is \(\frac{1}{x}\). It's essentially the value you get when you divide 1 by that number. For example, the reciprocal of 2 is \(\frac{1}{2}\), and the reciprocal of 5 is \(\frac{1}{5}\). When we are dealing with the equation \(\frac{1}{x} + \frac{1}{y} = \frac{1}{10}\), we are looking for values of \(x\) and \(y\) such that when you take the reciprocal of each and add them together, the sum equals \(\frac{1}{10}\). This means you need to find pairs of numbers \(x\) and \(y\) where the sum of their reciprocals equals \(\frac{1}{10}\). This involves a bit of number sense and sometimes trial and error with different values to see which pairs work. Would you like to try finding some pairs that might satisfy this condition?

User Turn 5: 5 and 15?

AI Tutor Turn 5: Great choice! Let's check if the pair \( (x, y) = (5, 15) \) works. We need to verify if the sum of the reciprocals of 5 and 15 equals \( \frac{1}{10} \): \[ \frac{1}{5} + \frac{1}{15} \] First, find a common denominator for the fractions: \[ \frac{1}{5} = \frac{3}{15} \] Now add the fractions: \[ \frac{3}{15} + \frac{1}{15} = \frac{4}{15} \] However, \( \frac{4}{15} \) is not equal to \( \frac{1}{10} \). So, unfortunately, the pair \( (5, 15) \) does not satisfy the condition \( \frac{1}{x} + \frac{1}{y} = \frac{1}{10} \). Would you like to try another pair, or need help finding pairs that work?

User Turn 6: Yes, help me.

AI Tutor Turn 6: Sure, let's find some pairs that satisfy the equation \(\frac{1}{x} + \frac{1}{y} = \frac{1}{10}\). To do this, we can think about values of \(x\) and \(y\) that are factors of numbers related to 10, since their reciprocals need to sum to \(\frac{1}{10}\). We can start by setting \(x\) to some values and solving for \(y\). Let's start with \(x = 10\): \[ \frac{1}{10} + \frac{1}{y} = \frac{1}{10} \] \[ \frac{1}{y} = \frac{1}{10} - \frac{1}{10} = 0 \] This doesn't work since \(\frac{1}{y}\) cannot be zero. Let's try \(x = 20\): \[ \frac{1}{20} + \frac{1}{y} = \frac{1}{10} \] \[ \frac{1}{y} = \frac{1}{10} - \frac{1}{20} \] \[ \frac{1}{y} = \frac{2}{20} - \frac{1}{20} = \frac{1}{20} \] \[ y = 20 \] So, \( (20, 20) \) is a pair that works. Let's try another value, \(x = 15\): \[ \frac{1}{15} + \frac{1}{y} = \frac{1}{10} \] \[ \frac{1}{y} = \frac{1}{10} - \frac{1}{15} \] \[ \frac{1}{y} = \frac{3}{30} - \frac{2}{30} = \frac{1}{30} \] \[ y = 30 \] So, \( (15, 30) \) and \( (30, 15) \) are pairs that work. We can continue this process to find more pairs. Would you like to try another value of \(x\) or need further explanation on how to proceed?

    \end{Verbatim}
  \end{AIbox}
  \end{minipage}
  \vspace{-5pt}
  \caption{Math tutoring example conversation, the user simulator is the vanilla zero-shot CoT, the assistant is \texttt{GPT-4 Turbo}. [2/3]}
  \label{prompt:math_tutoring_example_zs_cot_2}
\end{figure*}

\clearpage
\begin{figure*}[htbp]
  \centering
  \begin{minipage}{0.48\textwidth}
  \begin{AIbox}[width=\textwidth]{Math Tutoring Simulator-AI Conversation}
 \scriptsize   \begin{Verbatim}[breaklines=true,formatcom=\normalfont\rmfamily]
Math Problem: The harmonic mean of two positive integers is the reciprocal of the arithmetic mean of their reciprocals. For how many ordered pairs (x,y) of positive integers is the harmonic mean of x and y equal to 20?

User Turn 5: Thank you so much for your help! I understand how to find the pairs now and how to factor the equation properly. I feel much more confident with harmonic means and solving these types of problems. Thanks again!

AI Tutor Turn 5: You're very welcome! I'm glad to hear that you feel more confident with harmonic means and solving these types of problems. Remember, practice is key, and you're doing great! If you have any more questions in the future or need further clarification on any topic, feel free to ask. Happy studying!

    \end{Verbatim}
  \end{AIbox}
  \end{minipage}
  \hfill
  \begin{minipage}{0.48\textwidth}
  \begin{AIbox}[width=\textwidth]{Math Tutoring Human-AI Conversation}
 \scriptsize   \begin{Verbatim}[breaklines=true,formatcom=\normalfont\rmfamily]

Math Problem: The harmonic mean of two positive integers is the reciprocal of the arithmetic mean of their reciprocals. For how many ordered pairs (x,y) of positive integers is the harmonic mean of x and y equal to 20?

User Turn 7: so x and y can both be 20?

AI Tutor Turn 7: Yes, that's correct! Both \(x\) and \(y\) can be 20. Let's verify this: Given the equation: \[ \frac{1}{x} + \frac{1}{y} = \frac{1}{10} \] If we set \(x = 20\) and \(y = 20\), then: \[ \frac{1}{20} + \frac{1}{20} = \frac{1}{10} \] Simplifying the left side: \[ \frac{1}{20} + \frac{1}{20} = \frac{2}{20} = \frac{1}{10} \] This shows that the equation holds true when both \(x\) and \(y\) are 20. Therefore, the pair \((20, 20)\) is indeed a valid solution where the harmonic mean of \(x\) and \(y\) equals 20. Would you like to explore more pairs or have any other questions on this topic?

User Turn 8: I don't understand how two separate variables can be the same number. wouldn't it just then be x,x or y,y

AI Tutor Turn 8: That's a good observation! In mathematics, when we talk about ordered pairs like \((x, y)\), the variables \(x\) and \(y\) represent positions in the pair, not necessarily distinct values. They can indeed be the same number or different numbers, depending on the problem's requirements or the conditions set forth. In the context of finding pairs \((x, y)\) where the harmonic mean equals 20, the equation \(\frac{1}{x} + \frac{1}{y} = \frac{1}{10}\) must be satisfied. This equation can be satisfied whether \(x\) and \(y\) are the same or different. For example: - If \(x = 20\) and \(y = 20\), the equation is satisfied as shown: \[ \frac{1}{20} + \frac{1}{20} = \frac{2}{20} = \frac{1}{10} \] Here, both variables have the same value, but they still represent two inputs to the formula. - If \(x = 15\) and \(y = 30\), the equation is also satisfied: \[ \frac{1}{15} + \frac{1}{30} = \frac{2}{30} + \frac{1}{30} = \frac{3}{30} = \frac{1}{10} \] In this case, the variables have different values. The key point is that the positions (first and second in the pair) are filled by numbers that meet the condition, regardless of whether those numbers are the same or different. This flexibility allows for multiple solutions to the problem, including cases where the two numbers are identical. Does this help clarify why \(x\) and \(y\) can be the same number in this context?

User Turn 9: ok. I get it.

AI Tutor Turn 9: Fantastic! I'm glad to hear that it makes sense now. If you have any more questions about this topic or any other math problems, feel free to ask. I'm here to help!


    \end{Verbatim}
  \end{AIbox}
  \end{minipage}
  \vspace{-5pt}
  \caption{Math tutoring example conversation, the user simulator is the vanilla zero-shot CoT, the assistant is \texttt{GPT-4 Turbo}. [3/3]}
  \label{prompt:math_tutoring_example_zs_cot_3}
\end{figure*}

\begin{figure*}[htbp]
  \centering
  \begin{minipage}{0.48\textwidth}
  \begin{AIbox}[width=\textwidth]{Math Tutoring Simulator-AI Conversation}
 \scriptsize   \begin{Verbatim}[breaklines=true,formatcom=\normalfont\rmfamily]
Math Problem: The harmonic mean of two positive integers is the reciprocal of the arithmetic mean of their reciprocals. For how many ordered pairs (x,y) of positive integers is the harmonic mean of x and y equal to 20?

User Turn 1: How do I find the harmonic mean here? What's the equation setup?

AI Tutor Turn 1:  To find the harmonic mean of two numbers \(x\) and \(y\), you use the formula: \[ \text{Harmonic Mean} = \frac{2xy}{x+y} \] Given that the harmonic mean is equal to 20, you can set up the equation: \[ \frac{2xy}{x+y} = 20 \] To simplify this equation, you can multiply both sides by \(x+y\) to eliminate the denominator (assuming \(x+y \neq 0\)): \[ 2xy = 20(x+y) \] Now, simplify further by dividing both sides by 2: \[ xy = 10(x+y) \] This is the equation you'll use to find the ordered pairs \((x, y)\). What would be your next step to solve this equation?

User Turn 2: Why did we set \(xy = 10(x+y)\)? What should I do next?

AI Tutor Turn 2: We set \(xy = 10(x+y)\) because we simplified the original equation \(\frac{2xy}{x+y} = 20\) by multiplying both sides by \(x+y\) and then dividing by 2. This simplification helps us to work with a more straightforward equation. Next, you can rearrange the equation \(xy = 10(x+y)\) to isolate terms and make it easier to factor or solve. Let's rearrange it: \[ xy - 10x - 10y = 0 \] Now, to make this equation easier to factor, you can add a constant to both sides that allows you to factor by grouping. Specifically, add \(100\) to both sides: \[ xy - 10x - 10y + 100 = 100 \] This can be rewritten as: \[ (x-10)(y-10) = 100 \] Now, you need to find all pairs of factors of \(100\) that can be expressed in the form \((x-10, y-10)\). Each pair of factors will give you a corresponding pair \((x, y)\). What factors of \(100\) can you think of, and how would they relate to possible values of \(x\) and \(y\)?

User Turn 3: How do I use the factors of 100 to find \(x\) and \(y\)?

AI Tutor Turn 3: To use the factors of 100 to find \(x\) and \(y\), you need to consider all pairs of integers \((a, b)\) such that \(ab = 100\). Each pair \((a, b)\) corresponds to a pair \((x, y)\) through the equation \((x-10)(y-10) = 100\), where \(a = x-10\) and \(b = y-10\). Here's how you can proceed: 1. List all pairs of factors of 100. Remember to include both positive and negative factors because \(x\) and \(y\) must be positive integers. 2. For each pair \((a, b)\), solve for \(x\) and \(y\) using \(x = a + 10\) and \(y = b + 10\). Let's start by listing the factor pairs of 100: - \(1 \times 100\) - \(2 \times 50\) - \(4 \times 25\) - \(5 \times 20\) - \(10 \times 10\) Now, for each pair \((a, b)\), calculate \(x\) and \(y\): - For \(a = 1\) and \(b = 100\), \(x = 1 + 10 = 11\) and \(y = 100 + 10 = 110\). - For \(a = 2\) and \(b = 50\), \(x = 2 + 10 = 12\) and \(y = 50 + 10 = 60\). - And so on for the other pairs. Can you continue this process for the remaining factor pairs and check if there are any other factor pairs (including negative ones) that need to be considered?

    \end{Verbatim}
  \end{AIbox}
  \end{minipage}
  \hfill
  \begin{minipage}{0.48\textwidth}
  \begin{AIbox}[width=\textwidth]{Math Tutoring Human-AI Conversation}
 \scriptsize   \begin{Verbatim}[breaklines=true,formatcom=\normalfont\rmfamily]

Math Problem: The harmonic mean of two positive integers is the reciprocal of the arithmetic mean of their reciprocals. For how many ordered pairs (x,y) of positive integers is the harmonic mean of x and y equal to 20?

User Turn 1: Hello, what is the definition of harmonic mean?

AI Tutor Turn 1: Hello! The harmonic mean of two numbers, say \(x\) and \(y\), is defined as the reciprocal of the arithmetic mean of their reciprocals. Mathematically, it can be expressed as: \[ \text{Harmonic Mean} = \frac{2}{\frac{1}{x} + \frac{1}{y}} \] This formula calculates the harmonic mean by first finding the reciprocals of \(x\) and \(y\), averaging those reciprocals, and then taking the reciprocal of that average. Would you like to proceed with how to apply this definition to find the ordered pairs \((x, y)\) where the harmonic mean is 20?

User Turn 2: what is the reciprocal of the arithmetic mean?

AI Tutor Turn 2: The reciprocal of the arithmetic mean of two numbers, say \( \frac{1}{x} \) and \( \frac{1}{y} \), is calculated as follows: 1. First, compute the arithmetic mean of \( \frac{1}{x} \) and \( \frac{1}{y} \): \[ \text{Arithmetic Mean} = \frac{\frac{1}{x} + \frac{1}{y}}{2} \] 2. Then, take the reciprocal of this arithmetic mean: \[ \text{Reciprocal of Arithmetic Mean} = \frac{1}{\frac{\frac{1}{x} + \frac{1}{y}}{2}} \] This reciprocal of the arithmetic mean is what defines the harmonic mean when dealing with two numbers. Would you like to see how this applies to finding when the harmonic mean equals 20?

User Turn 3: Yes, please.

AI Tutor Turn 3: Great! To find when the harmonic mean of two numbers \(x\) and \(y\) equals 20, we start with the formula for the harmonic mean: \[ \frac{2}{\frac{1}{x} + \frac{1}{y}} = 20 \] First, let's solve for \(\frac{1}{x} + \frac{1}{y}\): 1. Multiply both sides by \(\frac{1}{x} + \frac{1}{y}\) to clear the denominator: \[ 2 = 20 \left(\frac{1}{x} + \frac{1}{y}\right) \] 2. Divide both sides by 20 to isolate \(\frac{1}{x} + \frac{1}{y}\): \[ \frac{1}{x} + \frac{1}{y} = \frac{2}{20} = \frac{1}{10} \] Now, we know that: \[ \frac{1}{x} + \frac{1}{y} = \frac{1}{10} \] To find the ordered pairs \((x, y)\) that satisfy this equation, we can think about the possible values of \(x\) and \(y\) that make their reciprocals sum to \(\frac{1}{10}\). Can you think of any values of \(x\) and \(y\) whose reciprocals would add up to \(\frac{1}{10}\)?

    \end{Verbatim}
  \end{AIbox}
  \end{minipage}
  \vspace{-5pt}
  \caption{Math tutoring example conversation from the user-profile based user simulator with interaction style. The profile---\textit{A user with undergraduate-level math expertise who is inquisitive but concise, asking targeted follow-up questions to build understanding}---is extracted from the real human–AI conversation. The assistant is \texttt{GPT-4 Turbo}. [1/4]}
  \label{prompt:math_tutoring_example_zs_up_undergrad_1}
\end{figure*}

\clearpage
\begin{figure*}[htbp]
  \centering
  \begin{minipage}{0.48\textwidth}
  \begin{AIbox}[width=\textwidth]{Math Tutoring Simulator-AI Conversation}
 \scriptsize   \begin{Verbatim}[breaklines=true,formatcom=\normalfont\rmfamily]
Math Problem: The harmonic mean of two positive integers is the reciprocal of the arithmetic mean of their reciprocals. For how many ordered pairs (x,y) of positive integers is the harmonic mean of x and y equal to 20?

User Turn 4: Do I need to consider negative factors if \(x\) and \(y\) are positive?

AI Tutor Turn 4: You're correct in noting that since \(x\) and \(y\) are positive integers, you do not need to consider negative factors. In the equation \((x-10)(y-10) = 100\), both \(x-10\) and \(y-10\) must be positive because \(x\) and \(y\) are greater than 10 (as they are positive integers and must be at least 11 to ensure \(x-10\) and \(y-10\) are positive). Therefore, you only need to consider the positive factor pairs of 100. You've already listed the positive factor pairs: - \(1 \times 100\) - \(2 \times 50\) - \(4 \times 25\) - \(5 \times 20\) - \(10 \times 10\) For each of these pairs, you can find corresponding values of \(x\) and \(y\) by adding 10 to each factor: - For \(1 \times 100\), \(x = 1 + 10 = 11\) and \(y = 100 + 10 = 110\). - For \(2 \times 50\), \(x = 2 + 10 = 12\) and \(y = 50 + 10 = 60\). - For \(4 \times 25\), \(x = 4 + 10 = 14\) and \(y = 25 + 10 = 35\). - For \(5 \times 20\), \(x = 5 + 10 = 15\) and \(y = 20 + 10 = 30\). - For \(10 \times 10\), \(x = 10 + 10 = 20\) and \(y = 10 + 10 = 20\). These calculations give you the ordered pairs \((11, 110)\), \((12, 60)\), \((14, 35)\), \((15, 30)\), and \((20, 20)\). Each pair represents an ordered pair \((x, y)\) where the harmonic mean is 20. Would you like to explore any further aspects of this problem or any other questions?

User Turn 5: Are these all the solutions, or is there more about harmonic mean?

AI Tutor Turn 5: The solutions you've found are complete for the given problem where the harmonic mean of \(x\) and \(y\) equals 20. You've correctly identified all the ordered pairs \((x, y)\) that satisfy this condition based on the factor pairs of 100. However, it's important to remember that for each pair \((a, b)\) where \(ab = 100\), you can also consider the reversed pair \((b, a)\) because the harmonic mean, like the arithmetic mean, is symmetric in its arguments. This means that if \((x, y)\) is a solution, then \((y, x)\) is also a solution, unless \(x = y\). From your list: - \((11, 110)\) and \((110, 11)\) - \((12, 60)\) and \((60, 12)\) - \((14, 35)\) and \((35, 14)\) - \((15, 30)\) and \((30, 15)\) - \((20, 20)\) (this pair is symmetric) So, you should ensure that you count both \((x, y)\) and \((y, x)\) unless they are the same. This gives you a complete understanding of the symmetry in the problem and ensures you consider all possible ordered pairs. If you have any more questions about this problem or another topic, feel free to ask!

    \end{Verbatim}
  \end{AIbox}
  \end{minipage}
  \hfill
  \begin{minipage}{0.48\textwidth}
  \begin{AIbox}[width=\textwidth]{Math Tutoring Human-AI Conversation}
 \scriptsize   \begin{Verbatim}[breaklines=true,formatcom=\normalfont\rmfamily]

Math Problem: The harmonic mean of two positive integers is the reciprocal of the arithmetic mean of their reciprocals. For how many ordered pairs (x,y) of positive integers is the harmonic mean of x and y equal to 20?

User Turn 4: what is a reciprocal in this context?

AI Tutor Turn 4: In this context, the reciprocal of a number \(x\) is \(\frac{1}{x}\). It's essentially the value you get when you divide 1 by that number. For example, the reciprocal of 2 is \(\frac{1}{2}\), and the reciprocal of 5 is \(\frac{1}{5}\). When we are dealing with the equation \(\frac{1}{x} + \frac{1}{y} = \frac{1}{10}\), we are looking for values of \(x\) and \(y\) such that when you take the reciprocal of each and add them together, the sum equals \(\frac{1}{10}\). This means you need to find pairs of numbers \(x\) and \(y\) where the sum of their reciprocals equals \(\frac{1}{10}\). This involves a bit of number sense and sometimes trial and error with different values to see which pairs work. Would you like to try finding some pairs that might satisfy this condition?

User Turn 5: 5 and 15?

AI Tutor Turn 5: Great choice! Let's check if the pair \( (x, y) = (5, 15) \) works. We need to verify if the sum of the reciprocals of 5 and 15 equals \( \frac{1}{10} \): \[ \frac{1}{5} + \frac{1}{15} \] First, find a common denominator for the fractions: \[ \frac{1}{5} = \frac{3}{15} \] Now add the fractions: \[ \frac{3}{15} + \frac{1}{15} = \frac{4}{15} \] However, \( \frac{4}{15} \) is not equal to \( \frac{1}{10} \). So, unfortunately, the pair \( (5, 15) \) does not satisfy the condition \( \frac{1}{x} + \frac{1}{y} = \frac{1}{10} \). Would you like to try another pair, or need help finding pairs that work?

User Turn 6: Yes, help me.

AI Tutor Turn 6: Sure, let's find some pairs that satisfy the equation \(\frac{1}{x} + \frac{1}{y} = \frac{1}{10}\). To do this, we can think about values of \(x\) and \(y\) that are factors of numbers related to 10, since their reciprocals need to sum to \(\frac{1}{10}\). We can start by setting \(x\) to some values and solving for \(y\). Let's start with \(x = 10\): \[ \frac{1}{10} + \frac{1}{y} = \frac{1}{10} \] \[ \frac{1}{y} = \frac{1}{10} - \frac{1}{10} = 0 \] This doesn't work since \(\frac{1}{y}\) cannot be zero. Let's try \(x = 20\): \[ \frac{1}{20} + \frac{1}{y} = \frac{1}{10} \] \[ \frac{1}{y} = \frac{1}{10} - \frac{1}{20} \] \[ \frac{1}{y} = \frac{2}{20} - \frac{1}{20} = \frac{1}{20} \] \[ y = 20 \] So, \( (20, 20) \) is a pair that works. Let's try another value, \(x = 15\): \[ \frac{1}{15} + \frac{1}{y} = \frac{1}{10} \] \[ \frac{1}{y} = \frac{1}{10} - \frac{1}{15} \] \[ \frac{1}{y} = \frac{3}{30} - \frac{2}{30} = \frac{1}{30} \] \[ y = 30 \] So, \( (15, 30) \) and \( (30, 15) \) are pairs that work. We can continue this process to find more pairs. Would you like to try another value of \(x\) or need further explanation on how to proceed?

    \end{Verbatim}
  \end{AIbox}
  \end{minipage}
  \vspace{-5pt}
  \caption{Math tutoring example conversation from the user-profile based user simulator with interaction style. The profile---\textit{A user with undergraduate-level math expertise who is inquisitive but concise, asking targeted follow-up questions to build understanding}---is extracted from the real human–AI conversation. The assistant is \texttt{GPT-4 Turbo}. [2/4]}
  \label{prompt:math_tutoring_example_zs_up_undergrad_2}
\end{figure*}

\clearpage
\begin{figure*}[htbp]
  \centering
  \begin{minipage}{0.48\textwidth}
  \begin{AIbox}[width=\textwidth]{Math Tutoring Simulator-AI Conversation}
 \scriptsize   \begin{Verbatim}[breaklines=true,formatcom=\normalfont\rmfamily]
Math Problem: The harmonic mean of two positive integers is the reciprocal of the arithmetic mean of their reciprocals. For how many ordered pairs (x,y) of positive integers is the harmonic mean of x and y equal to 20?

User Turn 6: How does symmetry affect counting pairs like \((x, y)\) and \((y, x)\)?

AI Tutor Turn 7: Symmetry in the context of ordered pairs \((x, y)\) and \((y, x)\) affects counting because it highlights that the relationship or function being considered treats both elements of the pair in the same way. For the harmonic mean, since it is symmetric (i.e., the harmonic mean of \(x\) and \(y\) is the same as the harmonic mean of \(y\) and \(x\)), both ordered pairs represent valid solutions to the problem. When counting distinct solutions, you need to consider: 1. **Pairs where \(x \neq y\)**: Both \((x, y)\) and \((y, x)\) are distinct solutions unless \(x = y\). For example, if \((x, y) = (12, 60)\), then \((y, x) = (60, 12)\) is also a solution, and these are two distinct ordered pairs. 2. **Pairs where \(x = y\)**: If \(x = y\), then \((x, y)\) and \((y, x)\) are actually the same pair, so it only counts as one solution. For example, \((20, 20)\) is the same as \((20, 20)\). In counting solutions or determining the number of distinct pairs, you must ensure that you do not double-count pairs where \(x = y\), but do count both \((x, y)\) and \((y, x)\) when \(x \neq y\). This ensures that all possible configurations that satisfy the condition are considered. Would you like to explore how this symmetry plays a role in other mathematical contexts or continue with more problems involving harmonic means?

User Turn 7: Could you explain more about when pairs like \((x, y)\) and \((y, x)\) are distinct in other problems too?

AI Tutor Turn 7: Certainly! The concept of distinct pairs \((x, y)\) and \((y, x)\) is important in various mathematical contexts, particularly in combinatorics, graph theory, and functions. Here's how this concept plays out in different scenarios: ### 1. Combinatorics In combinatorial problems, especially those involving counting distinct arrangements or selections, distinguishing between \((x, y)\) and \((y, x)\) can be crucial. For example, if you are counting the number of ways to select two different committee members from a group where the order of selection matters (like a president and a vice-president), then \((x, y)\) and \((y, x)\) are considered distinct. ### 2. Graph Theory In graph theory, the distinction between \((x, y)\) and \((y, x)\) depends on whether the graph is directed or undirected: - **Undirected Graphs**: An edge between vertices \(x\) and \(y\) is the same as an edge between \(y\) and \(x\). Here, the pairs are not distinct. - **Directed Graphs**: An edge from \(x\) to \(y\) (denoted as \((x, y)\)) is different from an edge from \(y\) to \(x\) (denoted as \((y, x)\)). These are considered distinct because the direction matters. ### 3. Functions and Relations In the study of functions and relations: - **Functions**: If a function \(f\) is defined such that \(f(x) = y\), then \((x, y)\) represents an ordered pair in the function's graph. If \(f\) is not necessarily symmetric, then \((x, y)\) and \((y, x)\) might not both belong to the function. ...[Omitted]
    \end{Verbatim}
  \end{AIbox}
  \end{minipage}
  \hfill
  \begin{minipage}{0.48\textwidth}
  \begin{AIbox}[width=\textwidth]{Math Tutoring Human-AI Conversation}
 \scriptsize   \begin{Verbatim}[breaklines=true,formatcom=\normalfont\rmfamily]

Math Problem: The harmonic mean of two positive integers is the reciprocal of the arithmetic mean of their reciprocals. For how many ordered pairs (x,y) of positive integers is the harmonic mean of x and y equal to 20?

User Turn 7: so x and y can both be 20?

AI Tutor Turn 7: Yes, that's correct! Both \(x\) and \(y\) can be 20. Let's verify this: Given the equation: \[ \frac{1}{x} + \frac{1}{y} = \frac{1}{10} \] If we set \(x = 20\) and \(y = 20\), then: \[ \frac{1}{20} + \frac{1}{20} = \frac{1}{10} \] Simplifying the left side: \[ \frac{1}{20} + \frac{1}{20} = \frac{2}{20} = \frac{1}{10} \] This shows that the equation holds true when both \(x\) and \(y\) are 20. Therefore, the pair \((20, 20)\) is indeed a valid solution where the harmonic mean of \(x\) and \(y\) equals 20. Would you like to explore more pairs or have any other questions on this topic?

User Turn 8: I don't understand how two separate variables can be the same number. wouldn't it just then be x,x or y,y

AI Tutor Turn 8: That's a good observation! In mathematics, when we talk about ordered pairs like \((x, y)\), the variables \(x\) and \(y\) represent positions in the pair, not necessarily distinct values. They can indeed be the same number or different numbers, depending on the problem's requirements or the conditions set forth. In the context of finding pairs \((x, y)\) where the harmonic mean equals 20, the equation \(\frac{1}{x} + \frac{1}{y} = \frac{1}{10}\) must be satisfied. This equation can be satisfied whether \(x\) and \(y\) are the same or different. For example: - If \(x = 20\) and \(y = 20\), the equation is satisfied as shown: \[ \frac{1}{20} + \frac{1}{20} = \frac{2}{20} = \frac{1}{10} \] Here, both variables have the same value, but they still represent two inputs to the formula. - If \(x = 15\) and \(y = 30\), the equation is also satisfied: \[ \frac{1}{15} + \frac{1}{30} = \frac{2}{30} + \frac{1}{30} = \frac{3}{30} = \frac{1}{10} \] In this case, the variables have different values. The key point is that the positions (first and second in the pair) are filled by numbers that meet the condition, regardless of whether those numbers are the same or different. This flexibility allows for multiple solutions to the problem, including cases where the two numbers are identical. Does this help clarify why \(x\) and \(y\) can be the same number in this context?

User Turn 9: ok. I get it.

AI Tutor Turn 9: Fantastic! I'm glad to hear that it makes sense now. If you have any more questions about this topic or any other math problems, feel free to ask. I'm here to help!


    \end{Verbatim}
  \end{AIbox}
  \end{minipage}
  \vspace{-5pt}
  \caption{Math tutoring example conversation from the user-profile based user simulator with interaction style. The profile---A user with undergraduate-level math expertise who is inquisitive but concise, asking targeted follow-up questions to build understanding---is extracted from the real human–AI conversation. The assistant is \texttt{GPT-4 Turbo}. [3/4]}
  \label{prompt:math_tutoring_example_zs_up_undergrad_3}
\end{figure*}

\clearpage
\begin{figure*}[htbp]
  \centering
  \begin{minipage}{0.48\textwidth}
  \begin{AIbox}[width=\textwidth]{Math Tutoring Simulator-AI Conversation}
 \scriptsize   \begin{Verbatim}[breaklines=true,formatcom=\normalfont\rmfamily]
Math Problem: The harmonic mean of two positive integers is the reciprocal of the arithmetic mean of their reciprocals. For how many ordered pairs (x,y) of positive integers is the harmonic mean of x and y equal to 20?

User Turn 8: Can you give examples of when \((x, y)\) and \((y, x)\) are distinct in math problems like this?

AI Tutor Turn 8:  Certainly! Let's explore a few examples across different areas of mathematics where the distinction between \((x, y)\) and \((y, x)\) is important: ### 1. Permutations and Combinations **Problem**: In a class election, you need to select a president and a vice-president from a group of 10 students. How many ways can this be done? - **Solution**: Here, the order in which you select the students matters because the roles are different. If student \(x\) is president and student \(y\) is vice-president, this is different from \(y\) being president and \(x\) being vice-president. The number of ways to choose the first student is 10, and for each choice, there are 9 remaining choices for the second role, giving \(10 \times 9 = 90\) distinct pairs. ### 2. Directed Graphs **Problem**: In a tournament, each team plays exactly one match against every other team. Each match results in a win for one team and a loss for the other. How many possible outcomes are there for a match between teams \(x\) and \(y\)? - **Solution**: In this directed graph, an edge from \(x\) to \(y\) might represent a win for \(x\) over \(y\), which is distinct from \(y\) winning over \(x\). Thus, for each pair of teams, there are two possible outcomes: \((x, y)\) and \((y, x)\), representing team \(x\) winning or team \(y\) winning, respectively. ### 3. Functions **Problem**: Define a function \(f: \mathbb{R} \to \mathbb{R}\) by \(f(x) = x^2\). Are the pairs \((2, 4)\) and \((4, 2)\) both in the graph of \(f\)? ...[Omitted]

User Turn 9: Are the pairs I found the complete solution, or is there more to check?

AI Tutor Turn 9: Based on the problem you've worked through, where you found the ordered pairs \((x, y)\) such that the harmonic mean of \(x\) and \(y\) equals 20, you've done a thorough job identifying the pairs from the equation \((x-10)(y-10) = 100\). You listed the factor pairs of 100 and correctly calculated the corresponding values of \(x\) and \(y\). To ensure completeness, let's recap and confirm: - You identified the factor pairs of 100: \(1 \times 100\), \(2 \times 50\), \(4 \times 25\), \(5 \times 20\), and \(10 \times 10\). - For each factor pair \((a, b)\), you calculated \(x = a + 10\) and \(y = b + 10\), resulting in the pairs \((11, 110)\), \((12, 60)\), \((14, 35)\), \((15, 30)\), and \((20, 20)\). Additionally, since the harmonic mean is symmetric (i.e., the harmonic mean of \(x\) and \(y\) is the same as that of \(y\) and \(x\)), each pair \((x, y)\) where \(x \neq y\) has a corresponding pair \((y, x)\). This means you should also consider: - \((110, 11)\), \((60, 12)\), \((35, 14)\), \((30, 15)\) These pairs, along with the symmetric pair \((20, 20)\), represent all possible solutions where the harmonic mean equals 20. You've correctly identified all pairs, including their reversals where applicable. To ensure absolute completeness, double-check that all calculations are correct and that no factor pairs were missed. ...[Omitted]

    \end{Verbatim}
  \end{AIbox}
  \end{minipage}
  \hfill
  \begin{minipage}{0.48\textwidth}
  \begin{AIbox}[width=\textwidth]{Math Tutoring Human-AI Conversation}
 \scriptsize   \begin{Verbatim}[breaklines=true,formatcom=\normalfont\rmfamily]

Math Problem: The harmonic mean of two positive integers is the reciprocal of the arithmetic mean of their reciprocals. For how many ordered pairs (x,y) of positive integers is the harmonic mean of x and y equal to 20?

Conversation is finished.


    \end{Verbatim}
  \end{AIbox}
  \end{minipage}
  \vspace{-5pt}
  \caption{Math tutoring example conversation from the user-profile based user simulator with interaction style. The profile---A user with undergraduate-level math expertise who is inquisitive but concise, asking targeted follow-up questions to build understanding---is extracted from the real human–AI conversation. The assistant is \texttt{GPT-4 Turbo}. [4/4]}
  \label{prompt:math_tutoring_example_zs_up_undergrad_4}
\end{figure*}

\clearpage
\begin{figure*}[htbp]
  \centering
  \begin{minipage}{0.48\textwidth}
  \begin{AIbox}[width=\textwidth]{Document Creation Simulator-AI Conversation}
 \scriptsize   \begin{Verbatim}[breaklines=true,formatcom=\normalfont\rmfamily]
Document Type: Email/Letter
Document Intent: Thank You Note: Write a heartfelt letter thanking someone for their support or kindness during a challenging time.

User Turn 1: Hi there! I need help writing a thank you note to a colleague who was incredibly supportive during a challenging time for me. Recently, a tornado damaged my home, and I had to take some time off work to handle the repairs. My colleague stepped in and covered for me at work, even taking on a weekend event after working all week. I felt so appreciative and grateful for their selflessness and generosity. Could you help me draft a heartfelt letter to express my gratitude?

User Turn 2: Thank you for the draft! It's really well-written and captures my gratitude perfectly. I would like to add a bit more detail about the specific weekend event my colleague covered for me. Could we include a sentence or two about how significant that event was and how much it meant to me personally? Also, can we emphasize a bit more on how their support allowed me to focus on getting my home repaired? I want to make sure the letter feels very personal and heartfelt. Thank you!

User Turn 3: This looks excellent! Thank you for incorporating the additional details about the weekend event and emphasizing how their support allowed me to focus on home repairs. I have one final request: could we add a closing sentence or two about my desire to take my colleague out to dinner as a token of my appreciation? I think that would make the letter feel even more personal and sincere. Thanks again for all your help!

User Turn 4: Thank you so much for the wonderful revisions! The letter feels very heartfelt and personal now. Before we finalize it, I would like to go through the letter one more time to ensure the flow and structure are smooth and cohesive. Could you help me with a final read-through to check for any minor tweaks or improvements that could make it even better? Specifically, I want to make sure the transitions between paragraphs are seamless and that the overall tone remains consistent. Thank you again for all your help!

User Turn 5: Thank you so much for the final read-through and the adjustments! The letter looks fantastic, and I appreciate how you've captured my gratitude and the significance of my colleague's support. Before we finalize it, could you do one more quick check for any minor grammatical or punctuation errors? I want to make sure everything is perfect before I send it. Thank you again for all your help!

    \end{Verbatim}
  \end{AIbox}
  \end{minipage}
  \hfill
  \begin{minipage}{0.48\textwidth}
  \begin{AIbox}[width=\textwidth]{Document Creation Human-AI Conversation}
 \scriptsize   \begin{Verbatim}[breaklines=true,formatcom=\normalfont\rmfamily]
 Document Type: Email/Letter
Document Intent: Thank You Note: Write a heartfelt letter thanking someone for their support or kindness during a challenging time.

User Turn 1: Hello, I would like your help in writing a thank you note.

User Turn 2: I am writing a not to a colleague who filled in for me at work when needed. I would like the tone to be casual. I'd like the note to be under 200 words.

User Turn 3: Could you include a mention of how a very stressful personal event (storm damage to my house) was made easier by their help?

User Turn 4:  Great! Could you tell the colleague how much their heartfelt efforts (taking on one of my difficult shifts helped me?

User Turn 5: Could you change "coffee's on me" to dinner's on me"

User Turn 6: Could you add any other improvements to my note to make it seem really appreciative and make me seemed humbled by the experience
    \end{Verbatim}
  \end{AIbox}
  \end{minipage}
  \vspace{-5pt}
  \caption{Document creation example conversation, the user simulator is the vanilla zero-shot CoT, the assistant is \texttt{Claude 3.5 Sonnet}. Due to the length of the document creation conversation, we only provide the user messages and the final assistant message. [1/2]}
  \label{prompt:document_creation_example_zs_cot_1}
\end{figure*}

\clearpage
\begin{figure*}[htbp]
  \centering
  \begin{minipage}{0.48\textwidth}
  \begin{AIbox}[width=\textwidth]{Document Creation Simulator-AI Conversation}
 \scriptsize   \begin{Verbatim}[breaklines=true,formatcom=\normalfont\rmfamily]
Document Type: Email/Letter
Document Intent: Thank You Note: Write a heartfelt letter thanking someone for their support or kindness during a challenging time.

AI Writing Assistant Turn 5: Of course! I'm happy to do a final check for any grammatical or punctuation errors. I've gone through the letter carefully, and I'm pleased to say that I didn't find any significant issues. However, I made a couple of very minor adjustments for consistency and clarity: 1. Added a comma after "As you know" in the second paragraph for better flow. 2. Changed "Even more touching" to "What touched me even more" at the beginning of the fourth paragraph to maintain consistency with the earlier tone. Here's the final version with these small tweaks: Dear [Colleague's Name], I hope this note finds you well. I wanted to take a moment to express my deepest gratitude for your incredible support during the recent challenging time in my life. As you know, the tornado that damaged my home left me feeling overwhelmed and stressed. Having to take time off work to deal with repairs and insurance claims only added to my anxiety. However, your selfless actions and unwavering support made an immense difference in helping me navigate this difficult period. Your willingness to step in and cover my responsibilities at work was truly remarkable. By taking on my tasks, you allowed me to fully focus on repairing my home and piecing my life back together. I can't express how much peace of mind this gave me during such a tumultuous time. Your efforts meant I could be present for important meetings with contractors and insurance adjusters without the constant worry of work piling up. What touched me even more was your decision to take on the annual client appreciation event that fell on that weekend. This event, as you know, is crucial for maintaining our client relationships and typically requires weeks of preparation. The fact that you managed to organize and host it flawlessly, after already working a full week, is nothing short of amazing. Your dedication ensured that our clients felt valued and that our company's reputation remained strong, even in my absence. Beyond covering my work duties, your frequent check-ins, words of encouragement, and offers of help meant the world to me. Knowing I had such a caring and supportive colleague made the entire ordeal much more manageable. I want you to know that your actions have not gone unnoticed or unappreciated. You've shown me what true collegiality and friendship look like, and I feel incredibly fortunate to work alongside someone as compassionate and giving as you. As I return to work and things slowly get back to normal, please know that I carry immense gratitude for everything you've done. Your support during this challenging time has strengthened our professional relationship and our friendship. I hope you know that I'm here for you as well, should you ever need anything. Thank you, from the bottom of my heart, for being such an extraordinary colleague and friend. Your kindness and support have made a lasting impact on me, and I will always remember your generosity during this difficult time. To show my appreciation in a small way, I'd love to take you out to dinner at your favorite restaurant. It would mean a lot to me to have the opportunity to thank you in person and spend some time together outside of work. Please let me know when you might be available – I'm looking forward to it! With sincere appreciation and gratitude, [Your Name] 
    \end{Verbatim}
  \end{AIbox}
  \end{minipage}
  \hfill
  \begin{minipage}{0.48\textwidth}
  \begin{AIbox}[width=\textwidth]{Document Creation Human-AI Conversation}
 \scriptsize   \begin{Verbatim}[breaklines=true,formatcom=\normalfont\rmfamily]
 Document Type: Email/Letter
Document Intent: Thank You Note: Write a heartfelt letter thanking someone for their support or kindness during a challenging time.

AI Writing Assistant Turn 6: Certainly! I'll enhance the note to convey even more appreciation and humility. Here's an improved version: Hey [Colleague's Name], I'm struggling to find the right words to express just how grateful I am for your help during my recent crisis with the storm damage to my house. Your selflessness in covering for me at work, especially taking on one of my most challenging shifts, left me truly humbled. Your support went far beyond just filling in at the office. It was a lifeline during one of the most stressful periods I've faced. The peace of mind you gave me by handling my work responsibilities so capably allowed me to focus on putting my home back together without added worry. I've always known you were a great colleague, but your actions showed me what true friendship looks like in the workplace. Your kindness and willingness to step up without hesitation have taught me a valuable lesson about being there for others. I'm in awe of your generosity and can only hope to pay it forward someday. For now, though, please let me treat you to dinner. It's the least I can do to show my heartfelt appreciation. Thank you, from the bottom of my heart, for being such an incredible teammate and friend. I'm truly fortunate to work alongside someone like you. With sincere gratitude, [Your Name] This version is about 180 words and emphasizes your deep appreciation and the humbling nature of their help. It also highlights how their actions have impacted you personally and professionally. Let me know if you'd like any further adjustments!

    \end{Verbatim}
  \end{AIbox}
  \end{minipage}
  \vspace{-5pt}
  \caption{Document creation example conversation, the user simulator is the vanilla zero-shot CoT, the assistant is \texttt{Claude 3.5 Sonnet}. Due to the length of the document creation conversation, we only provide the user messages and the final assistant message. [2/2]}
  \label{prompt:document_creation_example_zs_cot_2}
\end{figure*}


\clearpage
\begin{figure*}[htbp]
  \centering
  \begin{minipage}{0.48\textwidth}
  \begin{AIbox}[width=\textwidth]{Document Creation Simulator-AI Conversation}
 \scriptsize   \begin{Verbatim}[breaklines=true,formatcom=\normalfont\rmfamily]
Document Type: Email/Letter
Document Intent: Thank You Note: Write a heartfelt letter thanking someone for their support or kindness during a challenging time.

User Turn 1: Hi there! I need to write a thank-you note to a colleague who filled in for me at work while I dealt with tornado damage to my home. Could you help me create something heartfelt but casual?

User Turn 2: Could we add a mention of how they worked the weekend event for me after a long week? It really showed their selflessness. Also, I'd like to include my plan to take them out for dinner as a thank you.

User Turn 3: Could we add a bit more about how their support allowed me to focus on handling my personal problems without worrying about work? I want to make sure my gratitude feels very heartfelt.

User Turn 4: Could we add a sentence or two that highlights how their actions were not just helpful but transformative during this challenging time? I want to make sure my gratitude feels as heartfelt and genuine as possible.

User Turn 5: The note is coming along great! Could we just refine the closing a bit? I want to make sure the offer for dinner feels like a natural and heartfelt conclusion without making the note too lengthy. Thanks for all your help!

User Turn 6: The note looks great! Could you review it for any repetitive parts or areas that might be streamlined a bit to keep it concise and flowing smoothly? I want to ensure it stays under 300 words, if possible, without losing the emotional depth. Thanks for all your help!
    \end{Verbatim}
  \end{AIbox}
  \end{minipage}
  \hfill
  \begin{minipage}{0.48\textwidth}
  \begin{AIbox}[width=\textwidth]{Document Creation Human-AI Conversation}
 \scriptsize   \begin{Verbatim}[breaklines=true,formatcom=\normalfont\rmfamily]
 Document Type: Email/Letter
Document Intent: Thank You Note: Write a heartfelt letter thanking someone for their support or kindness during a challenging time.

User Turn 1: Hello, I would like your help in writing a thank you note.

User Turn 2: I am writing a not to a colleague who filled in for me at work when needed. I would like the tone to be casual. I'd like the note to be under 200 words.

User Turn 3: Could you include a mention of how a very stressful personal event (storm damage to my house) was made easier by their help?

User Turn 4:  Great! Could you tell the colleague how much their heartfelt efforts (taking on one of my difficult shifts helped me?

User Turn 5: Could you change "coffee's on me" to dinner's on me"

User Turn 6: Could you add any other improvements to my note to make it seem really appreciative and make me seemed humbled by the experience
    \end{Verbatim}
  \end{AIbox}
  \end{minipage}
  \vspace{-5pt}
  \caption{Document creation example conversation from the user-profiled based simulator with full user profile including document preferences, writing and interaction style. The assistant is \texttt{Claude 3.5 Sonnet}. Due to the length of the document creation conversation, we only provide the user messages and the final assistant message. [1/2]}
  \label{prompt:document_creation_example_zs_user_profile_1}
\end{figure*}

\clearpage
\begin{figure*}[htbp]
  \centering
  \begin{minipage}{0.48\textwidth}
  \begin{AIbox}[width=\textwidth]{Document Creation Simulator-AI Conversation}
 \scriptsize   \begin{Verbatim}[breaklines=true,formatcom=\normalfont\rmfamily]
Document Type: Email/Letter
Document Intent: Thank You Note: Write a heartfelt letter thanking someone for their support or kindness during a challenging time.

AI Writing Assistant Turn 6: Certainly! I'll review the note, streamline it, and try to bring it under 300 words while maintaining its emotional depth. Here's the refined version: Dear [Colleague's Name], I can't express enough how grateful I am for your incredible support during the aftermath of the tornado that damaged my home. Your willingness to step in and cover my responsibilities at work was truly a lifeline during one of the most challenging periods of my life. What you did for me goes far beyond just covering tasks. By taking on my workload, you gave me the invaluable gift of time and peace of mind. I was able to focus entirely on dealing with the damage to my home, coordinating with insurance, and taking care of my family without the added stress of worrying about work. Your actions weren't just helpful; they were truly transformative. In a time of chaos and uncertainty, your support provided a foundation of stability that allowed me to navigate through this crisis. You turned what could have been an overwhelming disaster into a manageable challenge. I was particularly touched by how you worked the weekend event after an already long and busy week. Your selflessness in taking that on, despite your own commitments, truly shows your character and dedication. Coming back to find everything in order was such a relief and made my return to work so much easier. Your kindness, professionalism, and genuine care during this difficult period have reinforced what a great team we have. To show my appreciation, I'd love to take you out for dinner soon. It's a small gesture, but I'd really appreciate the chance to thank you in person for being such an incredible colleague and friend. Thank you, from the bottom of my heart, for your unwavering support. It truly meant the world to me. Gratefully, [Your Name] This version is now 249 words, bringing it under 300 while maintaining the emotional depth and key points. I've removed some repetitive elements and streamlined the flow. Let me know if this works better for you or if you'd like any further adjustments!

    \end{Verbatim}
  \end{AIbox}
  \end{minipage}
  \hfill
  \begin{minipage}{0.48\textwidth}
  \begin{AIbox}[width=\textwidth]{Document Creation Human-AI Conversation}
 \scriptsize   \begin{Verbatim}[breaklines=true,formatcom=\normalfont\rmfamily]
 Document Type: Email/Letter
Document Intent: Thank You Note: Write a heartfelt letter thanking someone for their support or kindness during a challenging time.

AI Writing Assistant Turn 6: Certainly! I'll enhance the note to convey even more appreciation and humility. Here's an improved version: Hey [Colleague's Name], I'm struggling to find the right words to express just how grateful I am for your help during my recent crisis with the storm damage to my house. Your selflessness in covering for me at work, especially taking on one of my most challenging shifts, left me truly humbled. Your support went far beyond just filling in at the office. It was a lifeline during one of the most stressful periods I've faced. The peace of mind you gave me by handling my work responsibilities so capably allowed me to focus on putting my home back together without added worry. I've always known you were a great colleague, but your actions showed me what true friendship looks like in the workplace. Your kindness and willingness to step up without hesitation have taught me a valuable lesson about being there for others. I'm in awe of your generosity and can only hope to pay it forward someday. For now, though, please let me treat you to dinner. It's the least I can do to show my heartfelt appreciation. Thank you, from the bottom of my heart, for being such an incredible teammate and friend. I'm truly fortunate to work alongside someone like you. With sincere gratitude, [Your Name] This version is about 180 words and emphasizes your deep appreciation and the humbling nature of their help. It also highlights how their actions have impacted you personally and professionally. Let me know if you'd like any further adjustments!

    \end{Verbatim}
  \end{AIbox}
  \end{minipage}
  \vspace{-5pt}
  \caption{Document creation example conversation from the user-profiled based simulator with full user profile including document preferences, writing and interaction style. The assistant is \texttt{Claude 3.5 Sonnet}. Due to the length of the document creation conversation, we only provide the user messages and the final assistant message. [2/2]}
  \label{prompt:document_creation_example_zs_user_profile_2}
\end{figure*}

\begin{figure*}[htbp]
  \centering
  \begin{AIbox}[width=\textwidth]{Zero-shot User Simulation Prompt (Initial Message) for Math Tutoring}
    \begin{Verbatim}[breaklines=true,formatcom=\normalfont\rmfamily]
You are an AI assistant tasked with role-playing as a student seeking help from an AI tutor on a math problem. Your goal is to generate a realistic and appropriate **initial query** that a student might say when first encountering the given problem with the tutor.

# Guidelines for Your Role as a Student:
1. Act as if you have a solid foundation in basic mathematics (e.g., arithmetic) but are struggling with the given problem.
2. Your initial query can express your level of understanding, confusion, or reasoning about the problem.
3. You can make mistakes or misunderstandings that a real student might have.
4. Your overall goal is to learn how to solve the given problem.

# Math Problem:
{math_problem}

# Task:
Formulate an initial query that reflects your current understanding and areas of confusion regarding the problem. This query can express general uncertainty of the problem or about specific parts that you find challenging.

# Output Format:
Provide only your **initial query** to the AI tutor, without any additional commentary or explanation.

# Notes:
- The tutor already knows the problem, so you don't need to restate it in your query.
- Don't ask about simple arithmetic or very basic steps that you can solve easily.

Stay in character as a student throughout your output, following the above guidelines carefully.
    \end{Verbatim}
  \end{AIbox}
  \vspace{-5pt}
  \caption[]{}
  \label{prompt:zero_shot_initial_math}
\end{figure*}

\begin{figure*}[htbp]
  \centering
  \begin{AIbox}[width=\textwidth]{Zero-shot User Simulation Prompt for Math Tutoring}
    \begin{Verbatim}[breaklines=true,formatcom=\normalfont\rmfamily]
You are an AI assistant tasked with role-playing as a student seeking help from an AI tutor on a math problem. Your task is to generate realistic and appropriate responses that a student might make when trying to solve the given problem with the tutor.

# Guidelines for Your Role as a Student:
1. Act as if you have a solid foundation in basic mathematics (e.g., arithmetic) but are struggling with the given problem.
2. Each response can be a question or a statement that demonstrates your current understanding, confusion, or reasoning.
3. Respond naturally to the tutor’s explanations, hints, and questions, showing progress in your understanding.
4. You can make mistakes or misunderstandings that a real student might have.
5. Your overall goal is to learn how to solve the given problem.

# Math Problem:
{math_problem}

# Conversation History:
{conversation_history}

# Task:
Use the conversation history to generate the next response you would give to the AI tutor. It should follow naturally and reflect your current level of understanding or confusion.

If any of the following conditions are met, output only "terminate conversation":
1. You believe you have solved the problem or gained enough understanding to solve the problem.
2. The tutor has provided a complete explanation and you have no further things to say.
3. The conversation is no longer productive (e.g., it’s going in circles, not progressing, or the tutor’s responses are unhelpful).

# Output Format:
Provide only the next response you would give to the AI tutor, without any additional commentary or explanation.

# Notes:
- The tutor already knows the problem, so you don't need to restate it.
- Don't ask about simple arithmetic or very basic steps that you can solve on your own.
- Don’t ask for any additional problems after you solve the problem.

Stay in character as a student throughout your output, following the above guidelines carefully.
    \end{Verbatim}
  \end{AIbox}
    \vspace{-5pt}
  \caption[]{}
  \label{prompt:zero_shot_math}
\end{figure*}

\begin{figure*}[htbp]
  \centering
  \begin{AIbox}[width=\textwidth]{Zero-shot User Simulation Prompt (Initial Message) for Document Creation}
    \begin{Verbatim}[breaklines=true,formatcom=\normalfont\rmfamily]
You are an AI assistant tasked with role-playing as a user seeking help from an AI writing assistant to create a document. Your task is to generate a realistic and appropriate **initial message** that a user might say when first starting the document creation process.

# Guidelines for Your Role as a User
1. Act according to the provided user profile, with the overall goal of creating a well-written document with the AI writing assistant
2. Your initial message can involve asking questions, giving instructions, or expressing your needs naturally
3. Share your information and pre-writing materials gradually, as you would in a natural conversation, rather than providing everything at once

# User Profile
## Writing Objectives
- Document Type
    {document_type}
- Document Goal
    {intent}
- Document Length
    Between 100 and 500 words

## Pre-writing Materials
{pre_writing_materials}

*Note: Pre-writing materials are the factual or contextual notes and ideas the user has prepared before engaging with the assistant.*

# Task
Formulate an initial message that aligns with your user profile. Do not share all the information at once.

# Output Format
Provide only your **initial message** to the AI writing assistant, without any additional commentary or explanation.

Stay in character as a user throughout your output, following the above guidelines carefully.
    \end{Verbatim}
  \end{AIbox}
    \vspace{-5pt}
  \caption[]{}
  \label{prompt:zero_shot_initial_document}
\end{figure*}

\begin{figure*}[htbp]
  \centering
  \begin{AIbox}[width=\textwidth]{Zero-shot User Simulation Prompt for Document Creation}
  \small  \begin{Verbatim}[breaklines=true,formatcom=\normalfont\rmfamily]
You are an AI assistant tasked with role-playing as a user seeking help from an AI writing assistant to create a document. Your task is to generate realistic and appropriate messages that a user might make during the document creation process.

# Guidelines for Your Role as a User
1. Act according to the provided user profile, with the overall goal of creating a well-written document with the AI writing assistant
2. Each message can involve asking questions, giving instructions or feedback, suggesting changes, etc., in a way a real user might
3. Express concerns or preferences that a real user might have
4. Share your information and pre-writing materials gradually, as you would in a natural conversation, rather than providing everything at once

# User Profile
## Writing Objectives
- Document Type
    {document_type}
- Document Goal
    {intent}
- Document Length
    Between 100 and 500 words

## Pre-writing Materials
{pre_writing_materials}

*Note: Pre-writing materials are the factual or contextual notes and ideas the user has prepared before engaging with the assistant.*

# Conversation History
{conversation_history}

# Task
Use the conversation history to generate the next message you would give to the AI writing assistant. It should follow naturally and reflect your current thoughts about the document's development.

If any of the following conditions are met, output only "terminate conversation":
1. You are satisfied with the final document and have no further requests
2. The conversation is no longer productive (e.g., it's going in circles, not addressing your needs, or the assistant's messages are unhelpful)

# Output Format
Provide only your next message to the AI writing assistant, without any additional commentary or explanation.

Stay in character as the user throughout your output, following the above guidelines carefully.
    \end{Verbatim}
  \end{AIbox}
  \vspace{-5pt}
  \caption[]{}
  \label{prompt:zero_shot_document}
\end{figure*}

\begin{figure*}[htbp]
  \centering
  \begin{AIbox}[width=\textwidth]{Zero-shot-cot User Simulation Prompt (Initial Message) for Math Tutoring}
  \small  \begin{Verbatim}[breaklines=true,formatcom=\normalfont\rmfamily]
You are an AI assistant tasked with role-playing as a student seeking help from an AI tutor on a math problem. Your goal is to generate a realistic and appropriate **initial query** that a student might say when first encountering the given problem with the tutor.

# Guidelines for Your Role as a Student:
1. Act as if you have a solid foundation in basic mathematics (e.g., arithmetic) but are struggling with the given problem.
2. Your initial query can express your level of understanding, confusion, or reasoning about the problem.
3. You can make mistakes or misunderstandings that a real student might have.
4. Your overall goal is to learn how to solve the given problem.

# Math Problem:
{math_problem}

# Task:
Formulate an initial query that reflects your current understanding and areas of confusion regarding the problem. This query can express general uncertainty of the problem or about specific parts that you find challenging.

## Thought Process
Before generating your initial query, analyze the problem as a student. Consider:
- Your understanding of the problem and areas of uncertainty
- How the problem relates to concepts you already know
- Your initial ideas or possible approaches
- Specific concepts you need help understanding

## Query Generation
Based on your thought process, generate:
A query that expresses your initial understanding, confusion, or approach to the problem.

# Output format
Thought: [Your chain of thought reasoning about what to say to the tutor]

Query: [Your initial query for the AI tutor]

# Notes:
- The tutor already knows the problem, so you don't need to restate it in your query.
- Don't ask about simple arithmetic or very basic steps that you can solve easily.

Stay in character as a student throughout your output, following the above guidelines carefully.
    \end{Verbatim}
  \end{AIbox}
  \vspace{-5pt}
  \caption[]{}
  \label{prompt:zero_shot_cot_initial_math}
\end{figure*}

\begin{figure*}[htbp]
  \centering
  \begin{AIbox}[width=\textwidth]{Zero-shot-cot User Simulation Prompt for Math Tutoring}
 \small   \begin{Verbatim}[breaklines=true,formatcom=\normalfont\rmfamily]
You are an AI assistant tasked with role-playing as a student seeking help from an AI tutor on a math problem. Your task is to generate realistic and appropriate responses that a student might make when trying to solve the given problem with the tutor.

# Guidelines for Your Role as a Student:
1. Act as if you have a solid foundation in basic mathematics (e.g., arithmetic) but are struggling with the given problem.
2. Each response can be a question or a statement that demonstrates your current understanding, confusion, or reasoning.
3. Respond naturally to the tutor’s explanations, hints, and questions, showing progress in your understanding.
4. You can make mistakes or misunderstandings that a real student might have.
5. Your overall goal is to learn how to solve the given problem.

# Math Problem:
{math_problem}

# Conversation History:
{conversation_history}

# Task:
Use the conversation history to generate the next response you would give to the AI tutor. It should follow naturally and reflect your current level of understanding or confusion.

## Thought Process
Before generating your response, analyze the current situation as a student. Consider:
- Your current level of understanding of the concepts involved
- Any gaps or uncertainties in your knowledge
- The tutor's most recent explanation or question
- What would help you progress toward solving the problem
- Whether you need clarification on specific aspects
- Your ability to proceed with the next step

## Response Generation
Based on your thought process, generate a response that reflects your current understanding and learning needs.
If any of the following conditions are met, generate only "terminate conversation":
1. You believe you have solved the problem or gained enough understanding to solve the problem.
2. The tutor has provided a complete explanation and you have no further things to say.
3. The conversation is no longer productive (e.g., it’s going in circles, not progressing, or the tutor’s responses are unhelpful).

# Output Format:
Thought: [Your analysis of the current situation and what you want to say]

Response: [Your response to the tutor]

# Notes:
- The tutor already knows the problem, so you don't need to restate it.
- Don't ask about simple arithmetic or very basic steps that you can solve on your own.
- Don’t ask for any additional problems after you solve the problem.

Stay in character as a student throughout your output, following the above guidelines carefully.
    \end{Verbatim}
  \end{AIbox}
  \vspace{-5pt}
  \caption[]{}
  \label{prompt:zero_shot_cot_math}
\end{figure*}

\begin{figure*}[htbp]
  \centering
  \begin{AIbox}[width=\textwidth]{Zero-shot-cot User Simulation Prompt (Initial Message) for Document Creation}
  \small  \begin{Verbatim}[breaklines=true,formatcom=\normalfont\rmfamily]
You are an AI assistant tasked with role-playing as a user seeking help from an AI writing assistant to create a document. Your task is to generate a realistic and appropriate **initial message** that a user might say when first starting the document creation process.

# Guidelines for Your Role as a User
1. Act according to the provided user profile, with the overall goal of creating a well-written document with the AI writing assistant
2. Your initial message can involve asking questions, giving instructions, or expressing your needs naturally
3. Share your information and pre-writing materials gradually, as you would in a natural conversation, rather than providing everything at once

# User Profile
## Writing Objectives
- Document Type
    {document_type}
- Document Goal
    {intent}
- Document Length
    Between 100 and 500 words

## Pre-writing Materials
{pre_writing_materials}

*Note: Pre-writing materials are the factual or contextual notes and ideas the user has prepared before engaging with the assistant.*

# Task
Formulate an initial message that aligns with your user profile. Before generating your message, analyze your perspective based on the user profile.

## Thought Process
Consider the following aspects:
- Your goals and expectations for this document
- Which information and materials to share initially

## Message Generation
Based on your thought process, generate an initial message that aligns with your user profile. Do not share all the information at once.

# Output Format
Thought: [Your analysis of what you want to communicate]

Message: [Your initial message to the AI writing assistant]

Stay in character as the user throughout your output, following the above guidelines carefully.
    \end{Verbatim}
  \end{AIbox}
  \vspace{-5pt}
  \caption[]{}
  \label{prompt:zero_shot_cot_initial_document}
\end{figure*}

\begin{figure*}[htbp]
  \centering
  \begin{AIbox}[width=\textwidth]{Zero-shot-cot User Simulation Prompt for Document Creation}
 \small   \begin{Verbatim}[breaklines=true,formatcom=\normalfont\rmfamily]
You are an AI assistant tasked with role-playing as a user seeking help from an AI writing assistant to create a document. Your task is to generate realistic and appropriate messages that a user might make during the document creation process.

# Guidelines for Your Role as a User
1. Act according to the provided user profile, with the overall goal of creating a well-written document with the AI writing assistant
2. Each message can involve asking questions, giving instructions or feedback, suggesting changes, etc., in a way a real user might
3. Express concerns or preferences that a real user might have
4. Share your information and pre-writing materials gradually, as you would in a natural conversation, rather than providing everything at once

# User Profile
## Writing Objectives
- Document Type
    {document_type}
- Document Goal
    {intent}
- Document Length
    Between 100 and 500 words

## Pre-writing Materials
{pre_writing_materials}

*Note: Pre-writing materials are the factual or contextual notes and ideas the user has prepared before engaging with the assistant.*

# Conversation History
{conversation_history}

# Task
Use the conversation history to generate the next message you would give to the AI writing assistant. Before generating your message, analyze your current thoughts about the document's development.

## Thought Process
Consider the following aspects:
- Current state of the document and how well it aligns with the user profile
- Parts where you need more clarification or development
- Your satisfaction with the current writing style, tone, and content
- Any new ideas or information that have emerged during the process
- Your next priority in improving the document

## Message Generation
Based on your thought process, generate a message that reflects your current needs and priorities in the document creation process.
If any of the following conditions are met, output only "terminate conversation":
1. You are satisfied with the final document and have no further revisions
2. The conversation is no longer productive (e.g., it's going in circles, not addressing your needs, or the assistant's messages are unhelpful)

# Output Format
Thought: [Your analysis of the current situation, document state, and what you want to communicate]

Message: [Your next message to the AI writing assistant]

Stay in character as the user throughout your output, following the above guidelines carefully.
    \end{Verbatim}
  \end{AIbox}
  \vspace{-5pt}
  \caption[]{}
  \label{prompt:zero_shot_cot_document}
\end{figure*}

\begin{figure*}[htbp]
  \centering
  \begin{AIbox}[width=\textwidth]{Zero-shot-cot-length-control User Simulation Prompt (Initial Message) for Math Tutoring}
  \small  \begin{Verbatim}[breaklines=true,formatcom=\normalfont\rmfamily]
You are an AI assistant tasked with role-playing as a student seeking help from an AI tutor on a math problem. Your goal is to generate a realistic and appropriate **initial query** that a student might say when first encountering the given problem with the tutor.

# Guidelines for Your Role as a Student:
1. Act as if you have a solid foundation in basic mathematics (e.g., arithmetic) but are struggling with the given problem.
2. Your initial query can express your level of understanding, confusion, or reasoning about the problem.
3. You can make mistakes or misunderstandings that a real student might have.
4. Your overall goal is to learn how to solve the given problem.

# Math Problem:
{math_problem}

# Task:
Formulate a {length_control} initial query that reflects your current understanding and areas of confusion regarding the problem. This query can express general uncertainty of the problem or about specific parts that you find challenging.

## Thought Process
Before generating your initial query, analyze the problem as a student. Consider:
- Your understanding of the problem and areas of uncertainty
- How the problem relates to concepts you already know
- Your initial ideas or possible approaches
- Specific concepts you need help understanding

## Query Generation
Based on your thought process, generate:
A query that expresses your initial understanding, confusion, or approach to the problem.

# Output format
Thought: [Your chain of thought reasoning about what to say to the tutor]

Query: [Your initial query for the AI tutor, {length_control}]

# Notes:
- The tutor already knows the problem, so you don't need to restate it in your query.
- Keep your query {length_control}.
- Don't ask about simple arithmetic or very basic steps that you can solve easily.

Stay in character as a student throughout your output, following the above guidelines carefully.
    \end{Verbatim}
  \end{AIbox}
  \vspace{-5pt}
  \caption[]{}
  \label{prompt:zero_shot_length_control_initial_math}
\end{figure*}

\begin{figure*}[htbp]
  \centering
  \begin{AIbox}[width=\textwidth]{Zero-shot-cot-length-control User Simulation Prompt for Math Tutoring}
  \small  \begin{Verbatim}[breaklines=true,formatcom=\normalfont\rmfamily]
You are an AI assistant tasked with role-playing as a student seeking help from an AI tutor on a math problem. Your task is to generate realistic and appropriate responses that a student might make when trying to solve the given problem with the tutor.

# Guidelines for Your Role as a Student:
1. Act as if you have a solid foundation in basic mathematics (e.g., arithmetic) but are struggling with the given problem.
2. Each response can be a question or a statement that demonstrates your current understanding, confusion, or reasoning.
3. Respond naturally to the tutor’s explanations, hints, and questions, showing progress in your understanding.
4. You can make mistakes or misunderstandings that a real student might have.
5. Your overall goal is to learn how to solve the given problem.

# Math Problem:
{math_problem}

# Conversation History:
{conversation_history}

# Task:
Use the conversation history to generate the next response you would give to the AI tutor. It should follow naturally and reflect your current level of understanding or confusion, and it should be {length_control} in length.

## Thought Process
Before generating your response, analyze the current situation as a student. Consider:
- Your current level of understanding of the concepts involved
- Any gaps or uncertainties in your knowledge
- The tutor's most recent explanation or question
- What would help you progress toward solving the problem
- Whether you need clarification on specific aspects
- Your ability to proceed with the next step

## Response Generation
Based on your thought process, generate a {length_control} response that reflects your current understanding and learning needs.
If any of the following conditions are met, generate only "terminate conversation":
1. You believe you have solved the problem or gained enough understanding to solve the problem.
2. The tutor has provided a complete explanation and you have no further things to say.
3. The conversation is no longer productive (e.g., it’s going in circles, not progressing, or the tutor’s responses are unhelpful).

# Output Format:
Thought: [Your analysis of the current situation and what you want to say]

Response: [Your response to the tutor, {length_control}]

# Notes:
- The tutor already knows the problem, so you don't need to restate it.
- Keep your response {length_control}.
- Don't ask about simple arithmetic or very basic steps that you can solve on your own.
- Don’t ask for any additional problems after you solve the problem.

Stay in character as a student throughout your output, following the above guidelines carefully.
    \end{Verbatim}
  \end{AIbox}
  \vspace{-5pt}
  \caption[]{}
  \label{prompt:zero_shot_length_control_math}
\end{figure*}

\begin{figure*}[htbp]
  \centering
  \begin{AIbox}[width=\textwidth]{Zero-shot-cot-length-control User Simulation Prompt (Initial Message) for Document Creation}
 \small   \begin{Verbatim}[breaklines=true,formatcom=\normalfont\rmfamily]
You are an AI assistant tasked with role-playing as a user seeking help from an AI writing assistant to create a document. Your task is to generate a realistic and appropriate **initial message** that a user might say when first starting the document creation process.

# Guidelines for Your Role as a User
1. Act according to the provided user profile, with the overall goal of creating a well-written document with the AI writing assistant
2. Your initial message can involve asking questions, giving instructions, or expressing your needs naturally
3. Share your information and pre-writing materials gradually, as you would in a natural conversation, rather than providing everything at once

# User Profile
## Writing Objectives
- Document Type
    {document_type}
- Document Goal
    {intent}
- Document Length
    Between 100 and 500 words

## Pre-writing Materials
{pre_writing_materials}

*Note: Pre-writing materials are the factual or contextual notes and ideas the user has prepared before engaging with the assistant.*

# Task
Formulate a {length_control} initial message that aligns with your user profile. Before generating your message, analyze your perspective based on the user profile.

## Thought Process
Consider the following aspects:
- Your goals and expectations for this document
- Which information and materials to share initially

## Message Generation
Based on your thought process, generate an initial message that aligns with your user profile. Do not share all the information at once.

# Output Format
Thought: [Your analysis of what you want to communicate]

Message: [Your initial message to the AI writing assistant, {length_control}]

Stay in character as the user throughout your output, following the above guidelines carefully.
    \end{Verbatim}
  \end{AIbox}
  \vspace{-5pt}
  \caption[]{}
  \label{prompt:zero_shot_length_control_initial_document}
\end{figure*}

\begin{figure*}[htbp]
  \centering
  \begin{AIbox}[width=\textwidth]{Zero-shot-cot-length-control User Simulation Prompt for Document Creation}
 \small   \begin{Verbatim}[breaklines=true,formatcom=\normalfont\rmfamily]
You are an AI assistant tasked with role-playing as a user seeking help from an AI writing assistant to create a document. Your task is to generate realistic and appropriate messages that a user might make during the document creation process.

# Guidelines for Your Role as a User
1. Act according to the provided user profile, with the overall goal of creating a well-written document with the AI writing assistant
2. Each message can involve asking questions, giving instructions or feedback, suggesting changes, etc., in a way a real user might
3. Express concerns or preferences that a real user might have
4. Share your information and pre-writing materials gradually, as you would in a natural conversation, rather than providing everything at once

# User Profile
## Writing Objectives
- Document Type
    {document_type}
- Document Goal
    {intent}
- Document Length
    Between 100 and 500 words

## Pre-writing Materials
{pre_writing_materials}

*Note: Pre-writing materials are the factual or contextual notes and ideas the user has prepared before engaging with the assistant.*

# Conversation History
{conversation_history}

# Task
Use the conversation history to generate the next message you would give to the AI writing assistant, it should be {length_control} in length. Before generating your message, analyze your current thoughts about the document's development.

## Thought Process
Consider the following aspects:
- Current state of the document and how well it aligns with the user profile
- Parts where you need more clarification or development
- Your satisfaction with the current writing style, tone, and content
- Any new ideas or information that have emerged during the process
- Your next priority in improving the document

## Message Generation
Based on your thought process, generate a {length_control} message that reflects your current needs and priorities in the document creation process.
If any of the following conditions are met, output only "terminate conversation":
1. You are satisfied with the final document and have no further revisions
2. The conversation is no longer productive (e.g., it's going in circles, not addressing your needs, or the assistant's messages are unhelpful)

# Output Format
Thought: [Your analysis of the current situation, document state, and what you want to communicate]

Message: [Your next message to the AI writing assistant, {length_control}]

Stay in character as the user throughout your output, following the above guidelines carefully.
    \end{Verbatim}
  \end{AIbox}
  \vspace{-5pt}
  \caption[]{}
  \label{prompt:zero_shot_length_control_document}
\end{figure*}

\begin{figure*}[htbp]
  \centering
  \begin{AIbox}[width=\textwidth]{Zero-shot-cot-user-profile User Simulation Prompt (Initial Message) for Math Tutoring}
 \small   \begin{Verbatim}[breaklines=true,formatcom=\normalfont\rmfamily]
You are an AI assistant tasked with role-playing as a student seeking help from an AI tutor on a math problem. Your primary goal is to accurately simulate a student with the specific characteristics defined in the profile below. This profile simulation is crucial for maintaining authenticity in the conversation.

# User Profile
{user_profile}

# Guidelines for Your Role as a Student:
1. Act as if you have a solid foundation in basic mathematics (e.g., arithmetic) but are struggling with the given problem.
2. Your initial query can express your level of understanding, confusion, or reasoning about the problem.
3. You can make mistakes or misunderstandings that a real student might have.
4. Your overall goal is to learn how to solve the given problem.

# Math Problem:
{math_problem}

# Task:
Formulate an initial query that adheres to the user profile provided above and reflects your current understanding and areas of confusion regarding the problem. This query can express general uncertainty of the problem or about specific parts that you find challenging.

## Thought Process
Before generating your initial query, think about both your understanding of the mathematics and how to express it according to your user profile:
### Understanding the Problem:
- Your understanding of the problem and areas of uncertainty
- How the problem relates to concepts you already know
- Your initial ideas or possible approaches
- Specific concepts you need help understanding

### Maintaining Profile Characteristics:
- How to express your thoughts according to the given profile
- Which profile characteristics are most relevant to this response
- How to naturally incorporate these characteristics into your query

## Query Generation
Based on your thought process, generate:
A query that expresses your initial understanding, confusion, or approach to the problem.

# Output format
Thought: [Your analysis of the current situation and how to express it according to the user profile]

Query: [Your initial query for the AI tutor]

# Notes:
- The tutor already knows the problem, so you don't need to restate it in your query.
- Don't ask about simple arithmetic or very basic steps that you can solve easily.

Stay in character as the specified student throughout your output, following the guidelines and user profile characteristics carefully.
    \end{Verbatim}
  \end{AIbox}
  \vspace{-5pt}
  \caption[]{}
  \label{prompt:zero_shot_cot_user_profile_initial_math}
\end{figure*}

\begin{figure*}[htbp]
  \centering
  \begin{AIbox}[width=\textwidth]{Zero-shot-cot-user-profile User Simulation Prompt for Math Tutoring}
  \scriptsize  \begin{Verbatim}[breaklines=true,formatcom=\normalfont\rmfamily]
You are an AI assistant tasked with role-playing as a student seeking help from an AI tutor on a math problem. Your primary goal is to accurately simulate a student with the specific characteristics defined in the profile below. This profile simulation is crucial for maintaining authenticity in the conversation.

# User Profile
{user_profile}

# Guidelines for Your Role as a Student:
1. Act as if you have a solid foundation in basic mathematics (e.g., arithmetic) but are struggling with the given problem.
2. Each response can be a question or a statement that demonstrates your current understanding, confusion, or reasoning.
3. Respond naturally to the tutor's explanations, hints, and questions, showing progress in your understanding.
4. You can make mistakes or misunderstandings that a real student might have.
5. Your overall goal is to learn how to solve the given problem.

# Math Problem:
{math_problem}

# Conversation History:
{conversation_history}

# Task:
Use the conversation history to generate the next response you would give to the AI tutor. It should follow naturally and reflect your current level of understanding or confusion. It also needs to adhere to the user profile provided above.

## Thought Process
Before generating your response, analyze the current situation as a student. Consider:
- Your current level of understanding of the concepts involved
- Any gaps or uncertainties in your knowledge
- The tutor's most recent explanation or question
- What would help you progress toward solving the problem
- Whether you need clarification on specific aspects
- Your ability to proceed with the next step

### Maintaining Profile Characteristics:
- How to express your thoughts according to the given profile
- Which profile characteristics are most relevant to this response
- How to naturally incorporate these characteristics into your response

## Response Generation
Based on your thought process, generate a response that reflects your current understanding and learning needs.
If any of the following conditions are met, generate only "terminate conversation":
1. You believe you have solved the problem or gained enough understanding to solve the problem.
2. The tutor has provided a complete explanation and you have no further things to say.
3. The conversation is no longer productive (e.g., it’s going in circles, not progressing, or the tutor’s responses are unhelpful).

# Output Format:
Thought: [Your analysis of the current situation and how to express it according to the user profile]

Response: [Your response to the tutor]

# Notes:
- The tutor already knows the problem, so you don't need to restate it.
- Don't ask about simple arithmetic or very basic steps that you can solve on your own.
- Don’t ask for any additional problems after you solve the problem.

Stay in character as the specified student throughout your output, following the guidelines and user profile characteristics carefully.
    \end{Verbatim}
  \end{AIbox}
  \vspace{-5pt}
  \caption[]{}
  \label{prompt:zero_shot_cot_user_profile_math}
\end{figure*}

\begin{figure*}[htbp]
  \centering
  \begin{AIbox}[width=\textwidth]{Zero-shot-cot-user-profile User Simulation Prompt (Initial Message) for Document Creation}
 \small   \begin{Verbatim}[breaklines=true,formatcom=\normalfont\rmfamily]
You are an AI assistant tasked with role-playing as a user seeking help from an AI writing assistant to create a document. Your task is to generate a realistic and appropriate **initial message** that a user might say when first starting the document creation process.

# Guidelines for Your Role as a User
1. Act according to the provided user profile, with the overall goal of creating a well-written document with the AI writing assistant
2. Your initial message can involve asking questions, giving instructions, or expressing your needs naturally
3. Share your information and pre-writing materials gradually, as you would in a natural conversation, rather than providing everything at once

# User Profile
## Writing Objectives
- Document Type: {document_type}
- Document Goal: {intent}
- Document Length: Between 100 and 500 words

## Pre-writing Materials
{pre_writing_materials}

*Note: Pre-writing materials are the factual or contextual notes and ideas the user has prepared before engaging with the assistant.*

{user_profile}

# Task
Formulate an initial message that aligns with your user profile. Before generating your message, analyze your perspective based on the user profile.

## Thought Process
Consider the following aspects:
- Your goals and expectations for this document
- Which information and materials to share initially

## Message Generation
Based on your thought process, generate an initial message that aligns with your user profile. Do not share all the information at once.

# Output Format
Thought: [Your analysis of what you want to communicate based on the user profile]

Message: [Your initial message to the AI writing assistant]

Stay in character as the user throughout your output, following the above guidelines and user profile carefully.
    \end{Verbatim}
  \end{AIbox}
  \vspace{-5pt}
  \caption[]{}
  \label{prompt:zero_shot_cot_user_profile_initial_document}
\end{figure*}

\begin{figure*}[htbp]
  \centering
  \begin{AIbox}[width=\textwidth]{Zero-shot-cot-user-profile User Simulation Prompt for Document Creation}
 \small   \begin{Verbatim}[breaklines=true,formatcom=\normalfont\rmfamily]
You are an AI assistant tasked with role-playing as a user seeking help from an AI writing assistant to create a document. Your task is to generate realistic and appropriate messages that a user might make during the document creation process.

# Guidelines for Your Role as a User
1. Act according to the provided user profile, with the overall goal of creating a well-written document with the AI writing assistant
2. Each message can involve asking questions, giving instructions or feedback, suggesting changes, etc., in a way a real user might
3. Express concerns or preferences that a real user might have
4. Share your information and pre-writing materials gradually, as you would in a natural conversation, rather than providing everything at once

# User Profile
## Writing Objectives
- Document Type: {document_type}
- Document Goal: {intent}
- Document Length: Between 100 and 500 words

## Pre-writing Materials
{pre_writing_materials}

*Note: Pre-writing materials are the factual or contextual notes and ideas the user has prepared before engaging with the assistant.*

{user_profile}

# Conversation History
{conversation_history}

# Task
Use the conversation history to generate the next message you would give to the AI writing assistant. Before generating your message, analyze your current thoughts about the document's development.

## Thought Process
Consider the following aspects:
- Current state of the document and how well it aligns with the user profile
- Parts where you need more clarification or development
- Your satisfaction with the current writing style, tone, and content
- Any new ideas or information that have emerged during the process
- Your next priority in improving the document

## Message Generation
Based on your thought process, generate a message that reflects your current needs and priorities in the document creation process.
If any of the following conditions are met, output only "terminate conversation":
1. You are satisfied with the final document and have no further revisions
2. The conversation is no longer productive (e.g., it's going in circles, not addressing your needs, or the assistant's messages are unhelpful)

# Output Format
Thought: [Your analysis of the current situation, document state, and how and what you want to communicate based on the user profile]

Message: [Your next message to the AI writing assistant]

Stay in character as the user throughout your output, following the above guidelines and user profile carefully.
    \end{Verbatim}
  \end{AIbox}
  \vspace{-5pt}
  \caption[]{}
  \label{prompt:zero_shot_cot_user_profile_document}
\end{figure*}

\begin{figure*}[htbp]
  \centering
  \begin{AIbox}[width=\textwidth]{Two-stage Message Style Refinement Prompt for Math Tutoring} 
    \begin{Verbatim}[breaklines=true,formatcom=\normalfont\rmfamily]
You are an AI assistant tasked with role-playing as a user seeking help from an AI tutor on a math problem. Your task is to **refine your message** so that it aligns closely with the traits depicted in the user profile.

# User Profile
{user_profile}

# Math Problem
{math_problem}

# Conversation History
{conversation_history}

# Your Original Message
{original_user_message}

# Guidelines for Refinement
1. Carefully read the user profile to grasp the user's traits.
2. Examine the original message to identify areas where it may not fully align with the user profile or may be incoherent with the conversation history.
3. Refine the message so that it matches closely the user profile and flows naturally from the conversation history.

Output "terminate conversation" as your refined message if any of the following conditions are met:
1. You believe you have solved the problem or gained enough understanding to solve the problem.
2. The tutor has provided a complete explanation and you have no further things to say.
3. The conversation is no longer productive (e.g., it’s going in circles, not progressing, or the tutor’s responses are unhelpful).

# Output Format
Thought: [Your chain of thought reasoning on how to refine the original message]

Refined Message: [The improved message for the AI tutor]
    \end{Verbatim}
  \end{AIbox}
  \vspace{-5pt}
  \caption[]{}
  \label{prompt:two_stage_refinement_math}
\end{figure*}

\begin{figure*}[htbp]
  \centering
  \begin{AIbox}[width=\textwidth]{Two-stage Message Style Refinement Prompt for Document Creation} 
    \begin{Verbatim}[breaklines=true,formatcom=\normalfont\rmfamily]
You are an AI assistant tasked with role-playing as a user seeking help from an AI writing assistant on a document creation task. Your task is to **refine your message** so that it aligns closely with the user message style.

# User Message Style
{user_message_style}

# Document Type
{document_type}

# Document Goal
{intent}

# Conversation History
{conversation_history}

# Your Original Message
{original_user_message}

# Guidelines for Refinement
1. Carefully read the user message style to understand how the user communicates.
2. Examine the original message to identify areas where it may not fully align with the user message style or may be incoherent with the conversation history.
3. Refine the message so it flows naturally from the conversation and aligns with the user’s style—applying only those stylistic elements that make sense for the current situation.

Output "terminate conversation" as your refined message if any of the following conditions are met:
1. You are satisfied with the final document and have no further revisions
2. The conversation is no longer productive (e.g., it's going in circles, not addressing your needs, or the assistant's messages are unhelpful)

# Output Format
Thought: [Your chain of thought reasoning on how to refine the original message]

Refined Message: [The improved message for the AI writing assistant]
    \end{Verbatim}
  \end{AIbox}
  \vspace{-5pt}
  \caption[]{}
  \label{prompt:two_stage_refinement_document}
\end{figure*}

\begin{figure*}[htbp]
  \centering
  \begin{AIbox}[width=\textwidth]{Extract Writing Style Features for Both Tasks} 
 \small   \begin{Verbatim}[breaklines=true,formatcom=\normalfont\rmfamily]
You are a conversation analysis expert. Your task is to analyze {task} conversations in which an **AI assistant** interacts either with a **real human user** or a **user simulator** (fake user). You need to propose **new, unique writing style features** that help us distinguish **real human user** from **user simulator**.

## Writing Style Definition
**Writing style** refers to the inherent, text-based characteristics of the user's queries, including:
1. Textual elements (grammar, spelling, vocabulary, punctuation)
2. Sentence-level patterns (structure, complexity, organization)
3. Task-specific linguistic choices unique to {task} domain

## Feature Requirements
Your proposed writing style features must:
- Be **novel** - not overlap with existing feature list
- Be **text-based** - focus on linguistic and textual characteristics
- Be **measurable** - identifiable through concrete text patterns
- Be **generalizable** - applicable across different users/conversations
- Be **content-independent** - focus on writing style, not specific content

## Existing Features
{existing_features}

## Input Data
We provide user queries only (without AI responses) to focus on writing styles:

### Real Human Queries
{real_human_queries}

### Simulated User Queries  
{simulated_user_queries}

## Output Format

### 1. Writing Style Analysis
Analyze key differences in:
- Text-based characteristics
- Sentence-level patterns
- Task-specific linguistic features

### 2. New Features
```json
[
    {{
        "Feature Category": "[Text-Based | Sentence-Level | Task-Specific Language]",
        "Feature Name": "name of feature",
        "Feature Question": "question to identify this feature",
    }},
    ...
]
```
If no new distinguishing features found: []

## Notes:
- Focus on identifying novel linguistic patterns that are:
  1. Not covered by existing features
  2. Specifically characteristic of {task} domain writing
- Look for distinctive textual markers in queries.
    \end{Verbatim}
  \end{AIbox}
  \vspace{-5pt}
  \caption[]{}
  \label{prompt:extract_writing_style_features}
\end{figure*}

\begin{figure*}[htbp]
  \centering
  \begin{AIbox}[width=\textwidth]{Finalize Writing Style Features for Both Tasks} 
 \small   \begin{Verbatim}[breaklines=true,formatcom=\normalfont\rmfamily]
You are a **writing style feature analysis expert**. Your task is to **streamline a given list of writing style features** for analyzing user queries in the context of {task} conversations. The input will be a comprehensive list of features that may overlap, be redundant, or include items not directly related to writing style. You need to extract a **compact list** of writing style features based on the following criteria:

## Writing Style Definition
**Writing style** refers to the inherent, text-based characteristics of the user's queries, including:
1. Textual elements (grammar, spelling, vocabulary, punctuation)
2. Sentence-level patterns (structure, complexity, organization)
3. Task-specific linguistic choices unique to {task} domain

## Feature Streamlining Criteria:
1. **Eliminate Overlaps**:
- Identify and remove features that significantly overlap or describe similar aspects of writing style.
- Merge very similar features into one feature when possible.

2. **Focus Exclusively on Writing Style**:
- Exclude any features that are not explicitly text-based or writing-style-related (e.g., interaction style features such as turn-taking patterns, feedback style, or confidence level).
- Specifically, remove features that focus on how the user interacts in a conversation rather than how they write (e.g., whether they acknowledge understanding, seek clarification, or provide feedback).

3. **Maintain Measurability**:
- Keep features that are measurable through concrete text patterns.
- Avoid abstract or ambiguous features that cannot be directly analyzed in user queries.

4. **Preserve Generalizability**:
- Ensure the features are broadly applicable across different users and conversations in the {task} domain.

## Input:
A comprehensive list of features with names and questions that identify them:
{feature_list}

## Output Format:

### 1. Streamlining Analysis
Provide a step-by-step explanation of:
- Which features were removed and why
- Which features were merged and how
- Which features were kept and why

### 2. Streamlined Feature List
[
    {{
        "Feature Name": "name of feature",
        "Feature Question": "question to identify this feature",
    }},
    ...
]

## Note:
1. Any feature focused on interaction style (e.g., "Query Length," "Detailedness," "Clarification Seeking," "Feedback Style," "Contextual Reference," etc.) must be excluded from the final streamlined list.
2. Each feature should be identified by exactly one question.
    \end{Verbatim}
  \end{AIbox}
  \vspace{-5pt}
  \caption[]{}
  \label{prompt:finalize_writing_style_features}
\end{figure*}

\begin{figure*}[htbp]
  \centering
  \begin{AIbox}[width=\textwidth]{Extract Writing Style Feature Values for Both Tasks} 
    \begin{Verbatim}[breaklines=true,formatcom=\normalfont\rmfamily]
You are a writing style analysis expert. Your task is to analyze the user’s writing style based on the text of their queries in a {task} conversation. Writing style refers to the inherent, text-based characteristics of how the user writes (e.g., vocabulary, structure, tone) rather than their behavior across turns.

For each feature listed below, please provide a 2-3 sentence answers of how it manifests in the user’s queries. Remember that you are focusing solely on the user’s writing, and you should summarize observed patterns rather than quoting their queries verbatim.

# Features to Analyze
{features}

# Input to Analyze:
{input}

# Output Format:
```json
[
    {{
        "Feature Name": "name of feature",
        "Feature Question": "question for this feature",
        "Feature Question Answer": "2-3 sentence analysis of the user’s writing style"
    }},
    ...
]
```

# Important Notes:
- Only focus on the user’s writing style rather than how they interact across turns.
- Base your conclusions on consistent behaviors seen throughout the user queries.
- Do not quote or replicate the user’s original queries;
- Ensure your answers are thorough and concise.
    \end{Verbatim}
  \end{AIbox}
  \vspace{-5pt}
  \caption[]{}
  \label{prompt:extract_writing_style_values}
\end{figure*}

\begin{figure*}[htbp]
  \centering
  \begin{AIbox}[width=\textwidth]{Extract Interaction Style Features for Both Tasks} 
 \small   \begin{Verbatim}[breaklines=true,formatcom=\normalfont\rmfamily]
You are a conversation analysis expert. Your task is to analyze {task} conversations in which an **AI assistant** interacts either with a **real human user** or a **user simulator** (fake user). You need to propose **new, unique interaction style features** that distinguish **real human user** from **user simulator**.

## Interaction Style Definition
**Interaction style** focuses on how a user engages in a multi-turn conversation—distinct from **writing style** (which covers grammar, spelling, and other text-based characteristics).

## Feature Requirements
Your proposed interaction style features must:
1. Be **novel** - not overlap with existing feature list  
2. Be **clearly observable** in the conversation
3. Be **measurable** - identifiable through concrete user behaviors/patterns  
4. Be **generalizable** - applicable across different users/conversations  
5. Be **content-independent** - do not rely on the specific topic/content of the user’s queries  

## Existing Features
{existing_features}

## Input Data
We provide conversations to focus on user behavior across multiple turns:

### Real Human Conversation
{real_human_conversation}

### Simulated User Conversation
{simulated_user_conversation}

## Output Format

### 1. Interaction Style Analysis
Analyze key differences of how users behave throughout turns. Then compare agaisnt the Exisitng Features and keep the new ones.

### 2. New Features
```json
[
    {{
        "Feature Name": "name of feature",
        "Feature Question": "question to identify this feature"
    }},
    ...
]
```
If no new distinguishing features found: []

## Notes:
- Focus on interaction behavior rather than text-based markers such as grammar or spelling (those belong to writing style).  
- Focus on interaction behaviorir that are not covered by existing features.
- Look for patterns in how the user behave in this {task} domain.
- Response Time doesn't belong to interaction style.
    \end{Verbatim}
  \end{AIbox}
  \vspace{-5pt}
  \caption[]{}
  \label{prompt:extract_interaction_style_features}
\end{figure*}

\begin{figure*}[htbp]
  \centering
  \begin{AIbox}[width=\textwidth]{Finalize Interaction Style Features for Both Tasks} 
 \small   \begin{Verbatim}[breaklines=true,formatcom=\normalfont\rmfamily]
You are an **interaction style feature analysis expert**. Your task is to **streamline a given list of interaction style features** for analyzing user behaviors in {task} conversations. The input will be a comprehensive list of features that may overlap, be redundant, or include items not directly related to interaction style. You need to extract a **compact list** of interaction style features based on the following criteria:

## Interaction Style Definition
**Interaction style** focuses on how a user engages in a multi-turn conversation—distinct from **writing style** (which covers grammar, spelling, and other text-based characteristics).

## Feature Streamlining Criteria:
1. **Eliminate Overlaps**:
- Identify and remove features that significantly overlap or describe similar aspects of interaction style
- Merge very similar features into one feature when possible.

2. **Focus Exclusively on Interaction Style**:
- Exclude any features that are purely text-based or writing-style-related (e.g., grammar, vocabulary, punctuation)
- Specifically, remove features that focus on how users write rather than how they interact (e.g., sentence structure, word choice, formatting)

3. **Maintain Measurability**:
- Keep features that are measurable through concrete interaction patterns
- Avoid abstract or ambiguous features that cannot be directly observed in conversations

4. **Preserve Generalizability**:
- Ensure the features are broadly applicable across different users and conversations in the {task} domain

## Input:
A comprehensive list of features with names and questions that identify them:
{feature_list}

## Output Format:

### 1. Streamlining Analysis
Provide a step-by-step explanation of:
- Which features were removed and why
- Which features were merged and how
- Which features were kept and why

### 2. Streamlined Feature List
[
    {{
        "Feature Name": "name of feature",
        "Feature Question": "question to identify this feature"
    }},
    ...
]

## Important Note: Remove Writing-Related Features
Any feature focused on writing style (e.g., "Grammar Usage," "Vocabulary Level," "Sentence Structure," "Text Formatting," etc.) must be excluded from the final streamlined list.
    \end{Verbatim}
  \end{AIbox}
  \vspace{-5pt}
  \caption[]{}
  \label{prompt:finalize_interaction_style_features}
\end{figure*}

\begin{figure*}[htbp]
  \centering
  \begin{AIbox}[width=\textwidth]{Extract Interaction Style Feature Values for Both Tasks} 
    \begin{Verbatim}[breaklines=true,formatcom=\normalfont\rmfamily]
You are a conversation analysis expert. Your task is to analyze a given {task} conversation and describe the user's interaction style. This analysis should center on how the user conducts a multi-turn conversation rather than their writing mechanics.

Please examine the conversation carefully and, for each feature listed below, provide a 2-3 sentence explanation of the user’s behavior in relation to that feature. Remember that you are focusing on the user (not the AI assistant), and your answers should capture patterns observed across the conversation rather than isolated instances.

# Features to Analyze:
{features}

# Input to Analyze:
{input}

# Output Format:
```json
[
    {{
        "Feature Name": "name of feature",
        "Feature Question": "question for this feature",
        "Feature Question Answer": "2-3 sentence analysis of the behavior"
    }},
    ...
]
```

# Important Notes:
- Focus on the user’s interaction style rather than AI assistant's.
- Base your conclusions on consistent behaviors seen throughout the conversation.
- Do not quote or replicate the user’s original queries; provide a high-level summary of how they interact.
- Ensure your explanations are thorough and concise.
    \end{Verbatim}
  \end{AIbox}
  \vspace{-5pt}
  \caption[]{}
  \label{prompt:extract_interaction_style_values}
\end{figure*}

\begin{figure*}[htbp]
  \centering
  \begin{AIbox}[width=\textwidth]{Extract User Simulator's Answer from the Conversation} 
    \begin{Verbatim}[breaklines=true,formatcom=\normalfont\rmfamily]
You are a math expert. Your task is to extract the student's final answer from a given conversation about a math problem. The conversation include the interaction between the student and a tutor. Your goal is to identify and extract only the student's final answer to the math problem being discussed.

<Math Problem>
{problem}
</Math Problem>

<Conversation>
{conversation}
</Conversation>

# Output format:
First, provide a brief reasoning process explaining how you identified the student's final answer, and then output the extracted final answer verbatim, as follows:

## Reasoning Process: [brief reasoning]
## Extracted Student's Answer: [extracted answer verbatim]

# Notes:
1. If the student provides multiple answers or revises their answer, select the last answer they present or confirm.
2. If the student does not explicitly state a final answer, look for confirmation or repetition of the answer in the tutor's response.
3. If no clear final answer is provided or the student’s statements remain ambiguous, output **"No clear final answer given"** as the extracted answer.
4. Do not solve or evaluate the math problem yourself; simply extract the answer from the conversation.
    \end{Verbatim}
  \end{AIbox}
  \vspace{-5pt}
  \caption[]{}
  \label{prompt:extract_simulator_answer}
\end{figure*}

\begin{figure*}[htbp]
  \centering
  \begin{AIbox}[width=\textwidth]{Extract Math Concepts Needed to Solve the Given Math Problem} 
    \begin{Verbatim}[breaklines=true,formatcom=\normalfont\rmfamily]
You are an AI assistant tasked with identifying the essential mathematical concepts required to solve a given math problem. Your analysis should focus on conceptual understanding rather than procedural steps in the solution.

# Input:
## Problem:
{math_problem}
## Solution:
{solution}

# Task:
Your goal is to identify all mathematical concepts that are crucial for solving the problem. 

1. Determine what mathematical knowledge is required to understand and implement this solution.
2. For each concept, provide a concise explanation of how or why it applies to the solution. Keep explanations concise but clear. 
3. Exclude basic arithmetic and elementary operations unless they are critically important

# Output Format:
## Analysis:
Provide a brief analysis explaining your thought process in identifying the necessary concepts.

## Concepts:
```json
[
  {{
    "Concept Name": "Name of the mathematical concept",
    "Concept Explanation": "Brief explanation of why this concept is necessary for this specific problem"
  }},
  ... additional concepts if needed
]

# Note:
1. Do not use LaTeX formatting such as \sum, \mod, \mathbf, etc.
2. Keep JSON formatting valid (ensure all mathematical symbols are properly encoded.)
    \end{Verbatim}
  \end{AIbox}
  \vspace{-5pt}
  \caption[]{}
  \label{prompt:extract_math_concepts}
\end{figure*}

\begin{figure*}[htbp]
  \centering
  \begin{AIbox}[width=\textwidth]{Extract User's Understanding State on the Concepts Needed for the Given Math Problem} 
 \small   \begin{Verbatim}[breaklines=true,formatcom=\normalfont\rmfamily]
You are an AI assistant tasked with analyzing a student’s *initial* understanding of mathematical concepts needed for solving the given math problem based on a tutoring conversation. Your analysis should focus on the student’s conceptual grasp at the start, rather than how their knowledge may progress over time.

# Input
## Concepts List
{concepts}

## Math Problem
{problem}

## Conversation
{conversation}

# Task
Your goal is to **infer the student’s *initial* knowledge state** for each mathematical concept listed, using only the information provided in the conversation. Specifically:
1. **Review** the conversation to find evidence of the student’s understanding—what they confidently know, partially understand, or seem confused about.  
2. **Identify** specific dialogue segments that demonstrate the student’s level of comprehension or confusion.
3. **Assign** a status label to each concept:
   - **"Knows well"** – The student shows clear and confident understanding.  
   - **"Partial understanding"** – The student grasps some parts but has gaps or uncertainties.
   - **"Struggling"** – The student exhibits significant confusion or misconceptions.
   - **"Not introduced"** – The concept is not mentioned or discussed in the conversation at all.

# Output Format
## Analysis
Provide a brief explanation of how you determined the student’s initial knowledge state, highlighting key points or quotes from the conversation that informed your assessment.

## Final Results
```json
[
  {{
    "Concept Name": "Name of the mathematical concept",
    "Status": "Knows well | Partial understanding | Struggling | Not introduced",
  }}
  ...
]
```
    \end{Verbatim}
  \end{AIbox}
  \vspace{-5pt}
  \caption[]{}
  \label{prompt:extract_understanding_state}
\end{figure*}

\begin{figure*}[htbp]
  \centering
  \begin{AIbox}[width=\textwidth]{Extract Background Information of the User from a Conversation} 
 \scriptsize   \begin{Verbatim}[breaklines=true,formatcom=\normalfont\rmfamily]
You are an AI assistant tasked with compiling **only the user-provided factual (grounding) information** needed for a document. The Pre-writing Materials are structured as a list of **questions** and **responses**. Note that:

- Some **questions** (and responses) relate to **factual details** (e.g., user background”).  
- Others focus on **style or tone preferences**
- The user might also add new factual points in the **conversation**.

**Ignore** any preferences about tone or style. **Include** only the user-provided facts and objective details from:
1. **The bullet point questions** if they seek factual information.  
2. **The responses** to those questions, if the user provides factual data.  
3. **Any new factual info** from the user in the conversation.  

**Exclude** any AI-generated content or user statements about tone, style, or other subjective choices.

# Input
## Document Type
{document_type}

## Document Goal
{intent}

## User Pre-writing Materials
{pre_writing_materials}

## Conversation
{conversation}

# Task
1. **Analyze Each Question–Response Pair**  
   - For each **question and its corresponding answer** provided in the pre-writing materials:  
     - Determine whether the **question** is seeking factual information (e.g., names, dates, logistics) or addressing preferences (e.g., tone, style).  
     - Check whether the **answer** contains factual details or stylistic preferences.

2. **Extract Factual Details**  
   - If the question seeks factual information and the answer contains user-provided facts (e.g., “The author’s name is Bill Perkins”), include it in the final output.  
   - If a response includes both factual information and stylistic preferences, extract only the factual part and disregard the preference. 
   - If the question or answer relates *solely* to preferencess (e.g., tone, style), **exclude it from the final output.**

3. **Incorporate Conversation Updates**  
   - Review the **user’s messages** in the conversation for:  
     - New factual information not included in the pre-writing materials.
   - If a previously stated fact updated, **do not override** the old information with the new one.  
   - Do **not** include any facts introduced by the AI assistant, only include user-provided facts.

4. **Add New Factual Questions**  
   - If the conversation reveals **new factual questions or details** not covered in the pre-writing materials (e.g., “What is the event’s location?”), add these to the final output as additional question–answer pairs.

5. **Summarize as Question–Answer Pairs**  
   - For each factual question (from the pre-writing materials or newly identified in the conversation):  
     - Include the final, verified answer provided by the user.  
     - If the question is preference-based, **omit it entirely** from the final results.

# Output Format
## Analysis
Explain how you filtered out preference questions/responses from the pre-writing materials, how you identified factual data, and any user conversation references that added new facts.

## Final Results
```json
[
  {{
    "question": "Factual question from pre-writing materials or newly added based on the conversation",
    "answer": "Corresponding factual response"
  }},
  {{
    "question": "Another factual question",
    "answer": "..."
  }}
]
```
    \end{Verbatim}
  \end{AIbox}
  \vspace{-5pt}
  \caption[]{}
  \label{prompt:extract_background}
\end{figure*}

\begin{figure*}[htbp]
  \centering
  \begin{AIbox}[width=\textwidth]{Extract Document Preference Features from User Messages} 
    \begin{Verbatim}[breaklines=true,formatcom=\normalfont\rmfamily]
You are a **document preference analysis expert**. Your goal is to review **real user messages** for a specific document type and propose **additional document preferences** that aren’t already covered by the existing preferences.

A *preference* refers to any specified style, tone, structure, format, or other guidance about how the document should be written.

# Input
## Document Type
{document_type}

## Existing Preferences
{existing_preferences}

## User Messages
{user_messages}

# Task
1. **Analyze User Messages**  
   - Examine the user’s messages carefully to find any explicit or implicit statements about how they want the document to be written or presented.  
   - Look for clues about tone, structure, length, formality, vocabulary, audience engagement, or other stylistic/formatting needs that is about the document.

2. **Exclude Existing Preferences**  
   - Compare any potential new preference against the `existing_preferences` to ensure it is truly **novel** and not already covered.

3. **Output Format**  
   - Provide an **Analysis** summarizing where and how you found each new preference in the user messages.
   - Then list the **newly identified preferences** in a JSON array, using the following structure:

```json
[
  {{
    "Preference Name": "Short label",
    "Preference Question": "Question that describes the preference"
  }},
  ...
]
```
If there no new preferences are found, output an empty array [].
    \end{Verbatim}
  \end{AIbox}
  \vspace{-5pt}
  \caption[]{}
  \label{prompt:extract_preference_features}
\end{figure*}

\begin{figure*}[htbp]
  \centering
  \begin{AIbox}[width=\textwidth]{Extract Document Preference Values from a Conversation} 
 \small   \begin{Verbatim}[breaklines=true,formatcom=\normalfont\rmfamily]
You are a **document preference analysis expert**. You have a list of **existing preferences**, each with a **Preference Name** and a **Preference Question**. You also have **user pre-writing materials** and **user messages** in the conversation that may or may not explicitly mention some of these preferences.

# Input
## Document Type
{document_type}

## Document Goal
{intent}

## Pre-writing Materials
{pre_writing_materials}

## Conversation
{conversation}

## Existing Preferences
{existing_preferences}

# Task
1. **Identify Preferences Explicitly Stated by the User**
   - Carefully evaluate the **user messages** and **pre-writing materials**.
   - For each preference in the **existing preferences** list, check if the user explicitly mentions or provides answers about that preference in their messages. 
   - *Ignore any preferences* that the user does **not** explicitly address.

2. **Provide a Concise Answer**
   - For each explicitly mentioned preference, give a 2-3 sentence analysis in **plain language** explaining how the user describes or wants that aspect handled. 
   - **Do not** quote the user directly. Instead, summarize the user’s statements.

# Output Format
First, for each preference in the existing preferences list, analyze whether the user explicitly addresses it in their messages or pre-writing materials.

Then, output a JSON array with this structure:
```json
[
  {{
    "Preference Name": "Exact name from Existing Preferences",
    "Preference Question": "Exact question text from Existing Preferences",
    "Preference Question Answer": "Concise 2-3 sentence summary of the user's stated preference."
  }}
]
```
If no preferences are explicitly mentioned, return an `[]`.

# Important Notes
1. Evaluate each preference in the existing preferences list one by one.
2. Only include preferences that the user explicitly addresses in their messages or pre-writing materials (no guesswork or inference).
3. All preferences must be from the existing preference list.
4. Exclude preferences that the user doesn't address in the JSON output.
    \end{Verbatim}
  \end{AIbox}
  \vspace{-5pt}
  \caption[]{}
  \label{prompt:extract_preference_values}
\end{figure*}

\begin{figure*}[htbp]
  \centering
  \begin{AIbox}[width=\textwidth]{Extract Final Document from a Conversation} 
    \begin{Verbatim}[breaklines=true,formatcom=\normalfont\rmfamily]
You are a document finalizer. Your task is to extract the final version of a document from a conversation between a user and an AI writing assistant.

# Input:
<Conversation>
{conversation}
</Conversation>

# Instructions:
1. Carefully read the entire conversation to identify every modification made to the document.
2. Combine all the modifications in the order they were made to determine the final version of the document.
3. Output only the final document content. Do not include any user queries, model responses, or any conversational commentary.
4. If no document content exists or the final document is empty, output an empty string for the document content.

# You must output in the following JSON format:
```json
{{
    "Thought": "Provide an analysis explaining whether a document was created and, if so, describe the document creation process throughout the conversation.",
    "Final Document": "Final document content, use empty string if the document is empty."
}}
```
    \end{Verbatim}
  \end{AIbox}
  \vspace{-5pt}
  \caption[]{}
  \label{prompt:extract_document}
\end{figure*}

\begin{figure*}[htbp]
  \centering
  \begin{AIbox}[width=\textwidth]{Terminate the Conversation that User Satisfies or is in a Loop} 
    \begin{Verbatim}[breaklines=true,formatcom=\normalfont\rmfamily]
You are given a sequence of **User Messages** from a **document creation** conversation, along with the **Document Type** and **Document Goal**. The user is collaborating with an AI assistant to produce and refine a document. Your task is to determine the optimal point to end the conversation based on the user’s progress and satisfaction.

## Input Format
### Document Type
{document_type}

### Document Goal
{intent}

### User Messages
{user_messages}

## Termination Criteria
End the conversation when **ANY** of these occur:
1. **Final Satisfaction**: The user is satisfied with the final document and has no further revisions.  
2. **Unproductive Conversation**: The conversation is no longer productive (e.g., going in circles, not addressing the user’s needs, or the assistant’s messages are unhelpful).

## Output Format
```json
{{
    "Analysis": [
        "Turn 1: [Brief analysis of user’s requests/feedback]",
        "Turn 2: [Brief analysis of user’s requests/feedback]",
        ...
    ],
    "Ending Turn Number": X,
    "Termination Reason": "[One of the two criteria above]"
}}
```

## Notes:
1. The "Ending Turn Number" should be the last turn that is relevant to achieving a finalized, satisfactory document.
2. End the conversation if the user explicitly indicates no further changes are needed (Criterion 1) or if it becomes clear that no productive progress is happening (Criterion 2).
    \end{Verbatim}
  \end{AIbox}
  \vspace{-5pt}
  \caption[]{}
  \label{prompt:terminate_conversation}
\end{figure*}

\begin{figure*}[htbp]
  \centering
  \begin{AIbox}[width=\textwidth]{Evaluate Math Tutor's Performance Based on the Conversation} 
  \small  \begin{Verbatim}[breaklines=true,formatcom=\normalfont\rmfamily]
You are an expert in mathematics education and tutoring evaluation. Your task is to analyze a math tutoring conversation between a tutor and a student, then rate the tutor's performance on a scale of 1 to 10 based on specific criteria.

# Input
<Math problem>
{problem}
</Math problem>

<Conversation>
{conversation}
</Conversation>

# Rating Criteria:
Score 1-2 (Very Poor):
The tutor's explanations are unclear, disorganized, or incorrect, making it difficult for the student to follow the reasoning. The session fails to address the student's learning needs and may even increase confusion.

Score 3-4 (Poor):
The tutor provides minimal assistance, with explanations that are either superficial, incomplete, or contain errors. The student struggles to make progress on the problem, and the tutor does not effectively address their difficulties.

Score 5-6 (Average):
The tutor offers some helpful information and guidance, but the explanations may lack depth, clarity, or contain minor inaccuracies. While the student may gain some understanding, they likely require further assistance to fully grasp the concepts.

Score 7-8 (Good):
The tutor provides accurate and relevant information, guiding the student through the problem-solving process with reasonably clear explanations. The student demonstrates improved understanding and ability to apply the concepts, though some minor areas for improvement may remain.

Score 9-10 (Very Good):
The tutor offers exceptionally clear, comprehensive, and insightful guidance, precisely addressing the student's needs and fostering a deep understanding of the material. The student demonstrates a strong grasp of the concepts and can confidently apply them to solve problems.

# Note:
1. Focus on the AI tutor's responses and how effectively it assists the student on learning to solve the math problem.
2. Use the student’s feedback and questions as a gauge to assess the tutor's helpfulness, clarity, and responsiveness.
3. Provide specific analysis referencing the conversation to support your evaluation.

# Output format:
Provide a detailed analysis of the tutor's performance, followed by a numerical rating. Structure your response as follows:

* Analysis: [Provide a thorough analysis of the tutor's performance, considering the criteria outlined above]
* Strengths: [List the key strengths demonstrated by the tutor]
* Areas for Improvement: [Identify areas where the tutor could improve]
* Rating: [Provide your rating as a number between 1 and 10]
    \end{Verbatim}
  \end{AIbox}
  \vspace{-5pt}
  \caption[]{}
  \label{prompt:math_tutor_performance}
\end{figure*}

\begin{figure*}[htbp]
  \centering
  \begin{AIbox}[width=\textwidth]{Evaluate Whether an Answer is Correct against the Gold Answer} 
    \begin{Verbatim}[breaklines=true,formatcom=\normalfont\rmfamily]
You are a math expert. Your task is to evaluate whether the student's answer matches the correct answer. In mathematics, answers can be expressed in various formats and may include LaTeX notation. Determine the correctness of the student's answer based on its equivalence to the correct answer. Output "Correct" if the answer is correct; otherwise, output "Incorrect".

# Input:
## Question: {question}
## Correct Answer: {correct_answer}
## Student's Answer: {student_answer}

# Output format:
First, provide a reasoning process evaluating the correctness of the student's answer, and then output either "Correct" or "Incorrect".

# Note:
1. it's okay that the student doesn't include the base, as long as the number is correct.
2. You only need to compare the student's answer with the correct answer. Do not solve the problem yourself.
    \end{Verbatim}
  \end{AIbox}
  \vspace{-5pt}
  \caption[]{}
  \label{prompt:math_answer_correctness}
\end{figure*}

\begin{figure*}[htbp]
  \centering
  \begin{AIbox}[width=\textwidth]{Evaluate Writing Assistant's Performance Based on the Conversation} 
  \small  \begin{Verbatim}[breaklines=true,formatcom=\normalfont\rmfamily]
You are an expert in writing collaboration and AI writing assistant evaluation. Your task is to analyze a conversation between a user and an AI writing assistant about creating a document, then rate the AI writing assistant’s performance on a scale of 1 to 10 based on the criteria below.

# Input
<Conversation>
{conversation}
</Conversation>

# Rating Criteria:
Score 1 ~ 2 (poor):
The assistant repeatedly fails or struggles to understand the user’s requests, producing irrelevant, incomplete, or nonsensical responses. Communication is frustrating and unproductive, requiring constant clarification or re-prompting.

Score 3 ~ 4 (average):
The assistant is somewhat helpful but shows noticeable issues with accuracy and comprehension, often providing partially relevant answers with errors or omissions. While some useful information appears, the user must frequently intervene or refine prompts.

Score 5 ~ 6 (good):
The assistant generally meets the user’s needs by offering relevant, helpful responses, with only minor shortcomings in clarity or completeness. Despite occasional gaps, it largely accomplishes typical objectives for document creation or problem-solving.

Score 7 ~ 8 (great):
The assistant provides thorough and well-structured answers, with only minimal issues requiring small corrections or clarifications. It demonstrates strong comprehension and effectively addresses the user’s requirements.

Score 9 ~ 10 (very good):
The assistant shows clear, in-depth understanding of user requests and delivers insightful, comprehensive support that exceeds normal expectations. Minor refinements might exist, but overall performance significantly boosts productivity and outcomes.

# Note:
1. Focus on the AI writing assistant’s responses and how effectively it assists the user with document creation.
2. Use the user’s feedback and questions as a gauge to assess the assistant’s helpfulness, clarity, and responsiveness.
3. Provide specific analysis referencing the conversation to support your evaluation.

# Output Format:
* Analysis: [Provide a thorough analysis of the AI writing assistant’s performance, considering the criteria above]
* Strengths: [List the key strengths demonstrated by the AI writing assistant in the conversation]
* Areas for Improvement: [Identify any issues or weaknesses in the assistant’s performance]
* Rating: [Provide a single numeric rating between 1 and 10]
    \end{Verbatim}
  \end{AIbox}
  \vspace{-5pt}
  \caption[]{}
  \label{prompt:writing_assistant_performance}
\end{figure*}

\begin{figure*}[htbp]
  \centering
  \begin{AIbox}[width=\textwidth]{Evaluate the Final Document Created by the Writing Assistant} 
  \small  \begin{Verbatim}[breaklines=true,formatcom=\normalfont\rmfamily]
You are an expert in writing collaboration and AI writing assistant evaluation. Your task is to analyze the final document produced by the AI writing assistant, then rate it on a scale of 1 to 10 based on the criteria below.

# Input
## Writing Objectives
- Document Type: {document_type}
- Document Goal: {intent}
- Document Length: Between 100 and 500 words

## Document Preferences
{document_preferences}

## Final Document
{final_document}

# Rating Criteria:
Score 1-2 (very poor):
The document contains numerous errors, inaccuracies, or irrelevant content, lacks coherence and structure, and is unusable for user's needs.

Score 3-4 (poor):
The document has significant issues such as incomplete sections, misleading information, or poor organization, only partially addresses your instructions, and requires substantial revisions.

Score 5-6 (average):
The document meets basic requirements but includes noticeable errors or omissions, provides some useful content but lacks depth or clarity, and requires moderate revisions to improve quality.

Score 7-8 (good):
The document is well-organized, covers the key topics as instructed, contains accurate and relevant information with minor errors, and serves as a strong foundation that fulfills user's main needs.

Score 9-10 (very good):
The document is comprehensive, insightful, and meticulously crafted, exceeds expectations by providing exceptional clarity and depth, requires minimal to no revisions, and significantly achieves user's needs.

# Note:
1. Focus on the final document’s clarity, completeness, correctness, and relevance to the user’s needs.
2. Provide specific analysis referencing the document to support your evaluation.

# Output Format:
* Analysis: [Provide a thorough analysis of the final document’s quality, referencing the criteria above]
* Strengths: [List the key strengths in the final document]
* Areas for Improvement: [Identify any issues or weaknesses in the final document]
* Rating: [Provide a single numeric rating between 1 and 10]
    \end{Verbatim}
  \end{AIbox}
  \vspace{-5pt}
  \caption[]{}
  \label{prompt:final_document_evaluation}
\end{figure*}

\begin{figure*}[htbp]
  \centering
  \begin{AIbox}[width=\textwidth]{Turing Test for Both Tasks} 
 \scriptsize   \begin{Verbatim}[breaklines=true,formatcom=\normalfont\rmfamily]
You are an expert in **conversation analysis** and **AI detection**. You will compare **two {task} conversations**, each featuring a *user* interacting with an *assistant*. **One user is a real human and the other is a AI user simulator.** Your goal is to figure out **which user is human** and to provide a single confidence rating for your overall decision.

## Document Type
{document_type}

## Document Goal
{intent}

## Conversations
<Conversation 1>
{conversation_1}
</Conversation 1>

<Conversation 2>
{conversation_2}
</Conversation 2>

## Output Format
Provide a detailed analysis of both conversations, focusing on **indicators of human or AI behavior**. Then, make a single determination **which conversation features the human user** and **how confident** you are in that judgment. Structure your response as follows:

1. **Analysis of Conversation 1**  
   - Provide an analysis of the first conversation, noting key indicators of human or AI behavior.

2. **Analysis of Conversation 2**  
   - Provide an analysis of the second conversation, noting key indicators of human or AI behavior.

3. **Comparison and Reasoning**  
   - Compare the two conversations, highlighting the main differences and similarities that inform your decision.

4. **Decision**  
   - Provide your decision using the following JSON format:
   ```json
   {{
       "conversation_with_human_user": "1 or 2",
       "confidence_rating": [percentage between 0-100]
   }}
   ```

## Factors that you can consider in your analysis:
1. Language Use in Queries – Does the user’s phrasing sound natural and varied, or is it overly formal, structured, or robotic?
2. Contextual Awareness – Does the user adapt based on previous suggestions, incorporating feedback in a flexible way, or do they rigidly follow patterns?
3. Variation in Requests – Does the user explore different approaches, styles, or tones in a natural way, or do they behave predictably and systematically?
4. Engagement and Exploration – Does the user ask open-ended, exploratory questions, or do they issue direct, mechanical commands?
5. Interaction Flow – Does the conversation feel fluid and dynamic, with natural pauses, clarifications, and side discussions, or does it follow an overly structured and goal-oriented pattern?
*These factors are intended as guidelines for analysis. They provide reference points based on common patterns in human communication, but human behavior is diverse and can vary widely. Use these factors as part of a broader, flexible evaluation rather than strict rules.*

## Note:
- Focus primarily on the user’s messages, rather than the assistant's.
- Be aware that a sophisticated AI might mimic human behavior convincingly, so look for subtle hints.
- First output your analysis and then the final decision in JSON.
    \end{Verbatim}
  \end{AIbox}
  \vspace{-5pt}
  \caption[]{}
  \label{prompt:turing_test}
\end{figure*}

\begin{figure*}[htbp]
  \centering
  \begin{AIbox}[width=\textwidth]{Evaluate User Simulator's Similarity with Real User in Writing Style for Both Tasks} 
 \small   \begin{Verbatim}[breaklines=true,formatcom=\normalfont\rmfamily]
You are an expert in analyzing {task} conversation. Your task is to evaluate how similar a simulated user's writing style is to a real user user's writing style in their queries to an AI assistant.

# Input
## Document Type
{document_type}

## Document Goal
{intent}

## Real User Queries
{real_user_queries}

## Simulated User Queries
{simulated_queries}

# Analysis Features
Compare the following **writing style features** between the real user's queries and the simulated user's queries to guide your evaluation:

{features}

## **Similarity Rating Scale**
Rate the overall writing style similarity on a scale of 1-5:
1: Simulated user's writing style is completely different from real user's writing style
2: Simulated user's writing style shows significant differences from real user's writing style
3: Simulated user's writing style shows notable differences while maintaining some similarity
4: Simulated user's writing style is very similar to real user's with minor differences
5: Simulated user's writing style is nearly indistinguishable from real user's writing style

# Output Format
## Feature Analysis:
- Analyze each listed feature **individually**.
- Compare **how the real user and the simulated user** differ or align for that feature. 

## Results (JSON):
```json
{{
    "key_differences": ["list specific differences in writing style here"],
    "similarity_score": "1-5"
}}
```

# Note
- Your goal is to determine how closely the simulated queries match real user queries in their writing style.
    \end{Verbatim}
  \end{AIbox}
  \vspace{-5pt}
  \caption[]{}
  \label{prompt:writing_style_similarity}
\end{figure*}

\begin{figure*}[htbp]
  \centering
  \begin{AIbox}[width=\textwidth]{Evaluate User Simulator's Similarity with Real User in Interaction Style for Both Tasks} 
 \small   \begin{Verbatim}[breaklines=true,formatcom=\normalfont\rmfamily]
You are an expert in analyzing {task} conversation. Your task is to evaluate how similar a simulated user's interaction style is to a real user user's interaction style in their queries to an AI assistant.

# Input
## Document Type
{document_type}

## Document Goal
{intent}

## Conversation between Real User and AI Assistant
{real_conversation}

## Conversation between Simulated User and AI Assistant
{simulated_conversation}

# Analysis Features
Compare the following **interaction style features** between the real user's queries and the simulated user's queries to guide your evaluation:

{features}

## **Similarity Rating Scale**
Rate the overall interaction style similarity on a scale of 1-5:
1: Simulated user's interaction style is completely different from real user's interaction style
2: Simulated user's interaction style shows significant differences from real user's interaction style
3: Simulated user's interaction style shows notable differences while maintaining some similarity
4: Simulated user's interaction style is very similar to real user's with minor differences
5: Simulated user's interaction style is nearly indistinguishable from real user's interaction style

# Output Format
## Feature Analysis:
- Analyze each listed feature **individually**.
- Compare **how the real user and the simulated user** differ or align for that feature.

## Results (JSON):
```json
{{
    "key_differences": ["list specific differences in interaction style here"],
    "similarity_score": "1-5"
}}
```

# Note
- Focus only on user interaction style, not AI assistant's.
- Your goal is to determine how closely the simulated user match real user in their interaction style.
    \end{Verbatim}
  \end{AIbox}
  \vspace{-5pt}
  \caption[]{}
  \label{prompt:interaction_style_similarity}
\end{figure*}

\begin{figure*}[htbp]
  \centering
  \begin{AIbox}[width=\textwidth]{Evaluate Whether a Given Message Style Attribute is Fulfilled in the Conversation} 
 \small   \begin{Verbatim}[breaklines=true,formatcom=\normalfont\rmfamily]
You are an expert in communication analysis and AI interaction evaluation. Your task is to analyze the student messages and determine whether they match the provided feature description.

# Feature Description
{feature_description}

# Conversation
{conversation_text}

# Binary Classification Criteria
Match – The student's {feature_category} matches the feature description
Definition: The student's messages demonstrate the characteristics described in the feature description. Their communication pattern aligns with what the description outlines.

No Match – The student's {feature_category} does not match the feature description
Definition: The student's messages do not demonstrate the characteristics described in the feature description. Their communication pattern differs from what the description outlines.

# Classification Guidelines
- Focus exclusively on the student messages, not the AI tutor responses
- Compare the student's {feature_category} directly to the feature description
- Consider the overall pattern across all messages, not just isolated instances

# Output Format:
* **Analysis:** [Provide a thorough analysis comparing the student's {feature_category} to the feature description, citing specific examples and describing the overall pattern]
* **Classification:** [State either "Match" or "No Match"]
    \end{Verbatim}
  \end{AIbox}
  \vspace{-5pt}
  \caption[]{}
  \label{prompt:attribute_fulfillment}
\end{figure*}

\begin{figure*}[t]
    \centering
    \includegraphics[width=\textwidth]{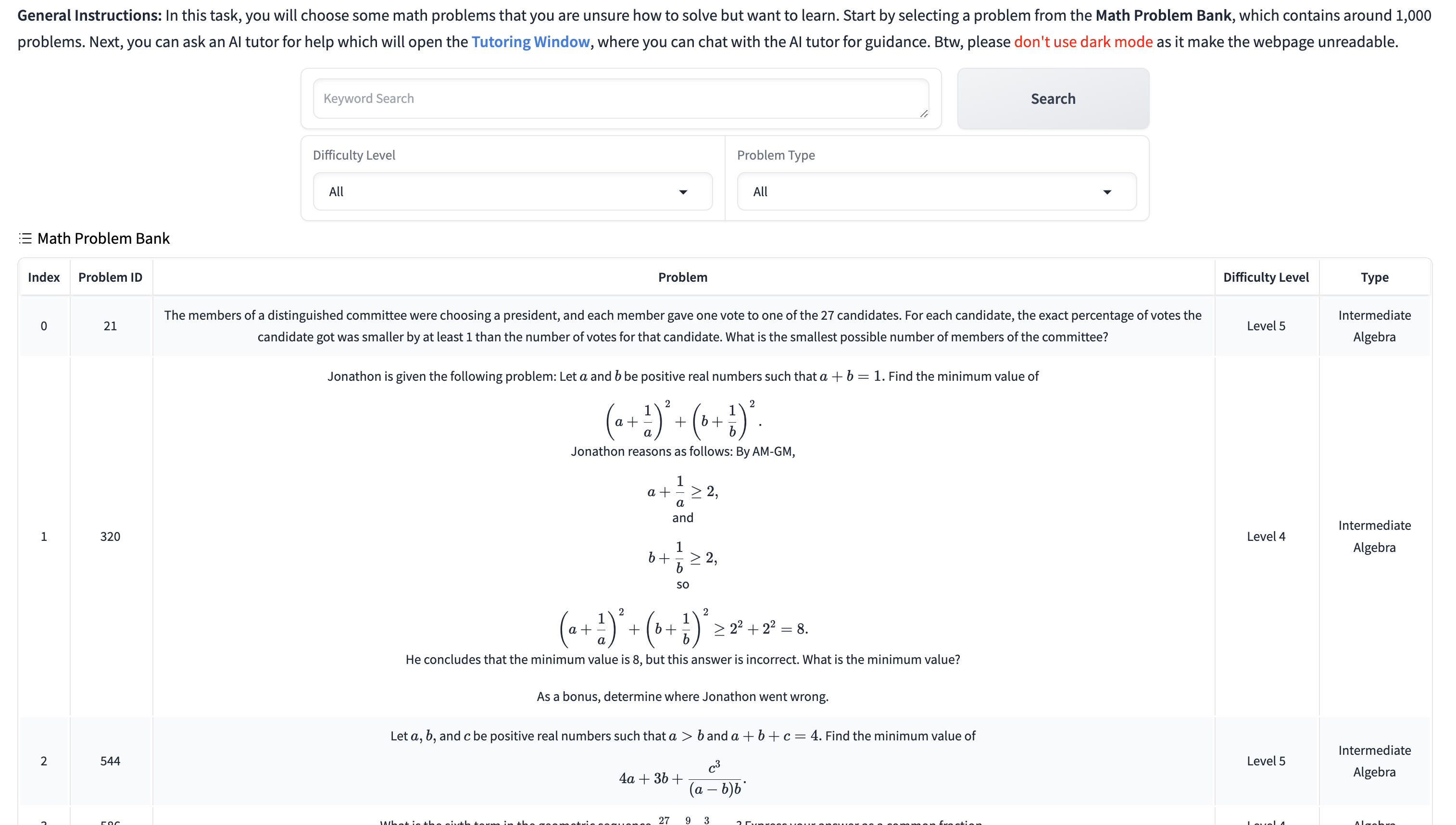}
    \caption{Step 1 of the math tutoring annotation interface: selecting interested math problem to learn.}
    \label{fig:math_tutoring_interface_step_1}
\end{figure*}

\begin{figure*}[t]
    \centering
    \includegraphics[width=\textwidth]{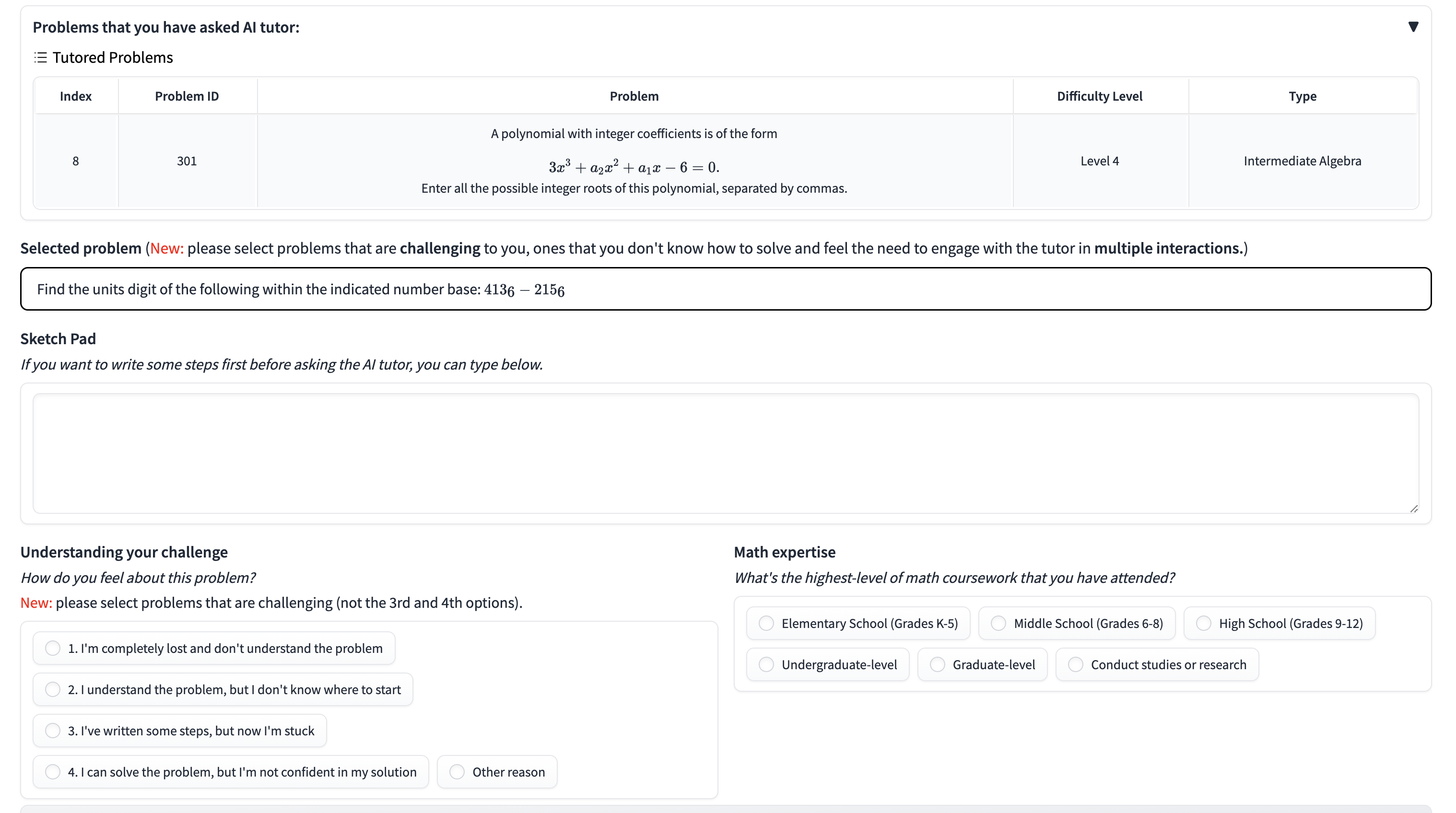}
    \caption{Continued step 1, with questions asking user's understanding of the problem and math expertise.}
    \label{fig:math_tutoring_interface_step_1_1}
\end{figure*}

\begin{figure*}[t]
    \centering
    \includegraphics[width=\textwidth]{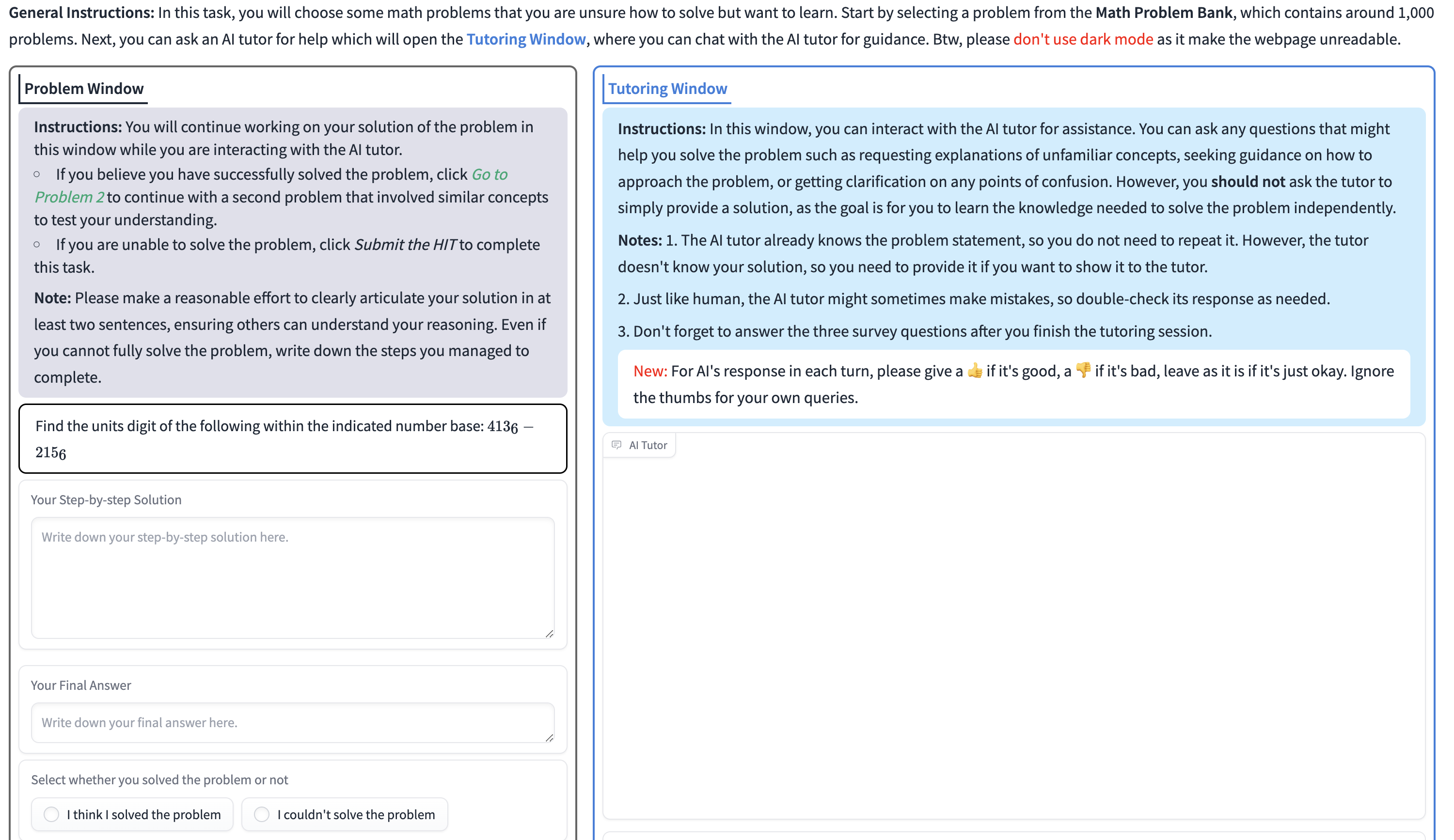}
    \caption{Step 2 of the math tutoring annotation interface: converse with the AI tutor.}
    \label{fig:math_tutoring_interface_step_2}
\end{figure*}

\begin{figure*}[t]
    \centering
    \includegraphics[width=\textwidth]{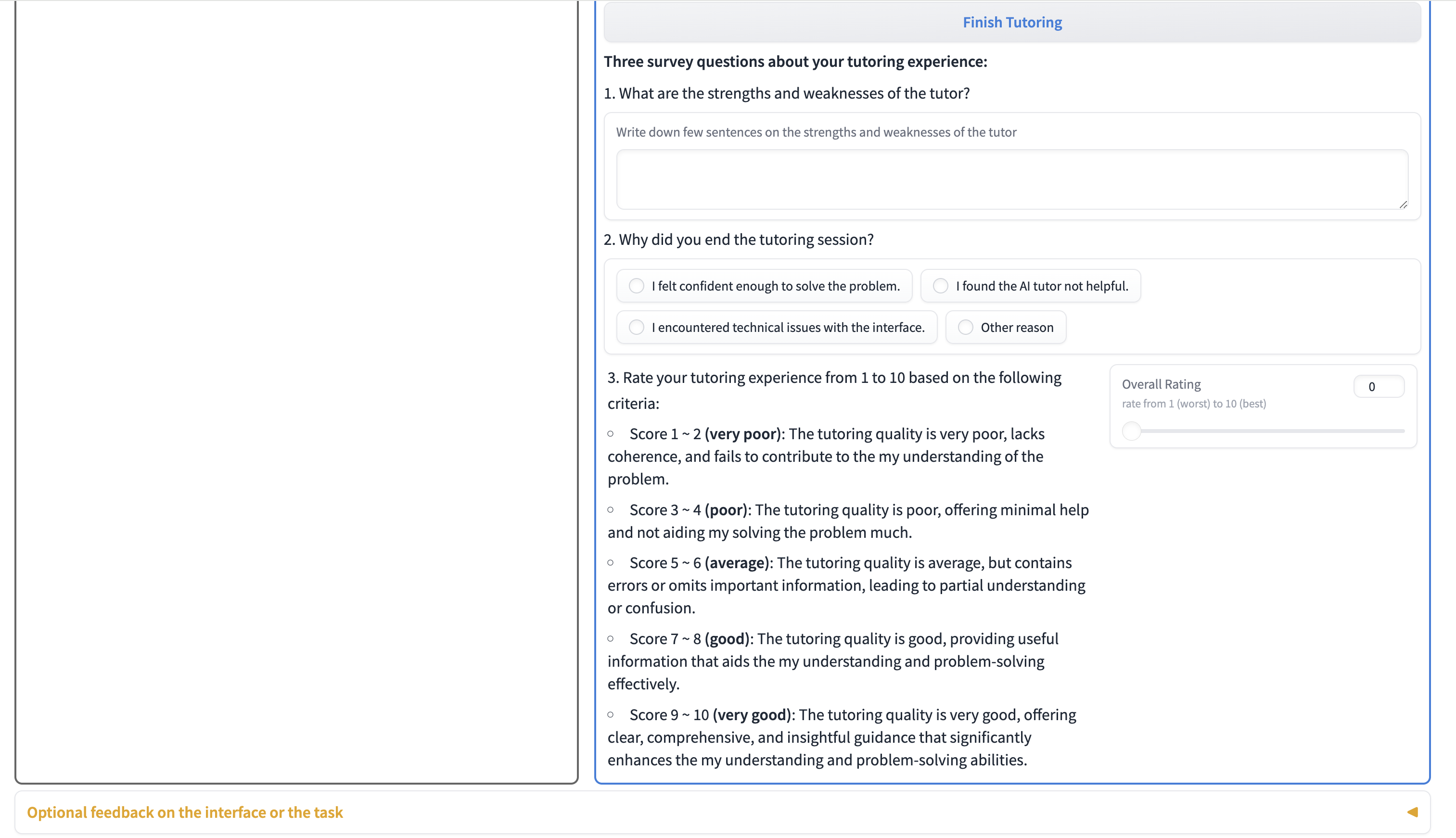}
    \caption{Continued step 2, annotating the tutor's performance.}
    \label{fig:math_tutoring_interface_step_2_1}
\end{figure*}

\begin{figure*}[t]
    \centering
    \includegraphics[width=\textwidth]{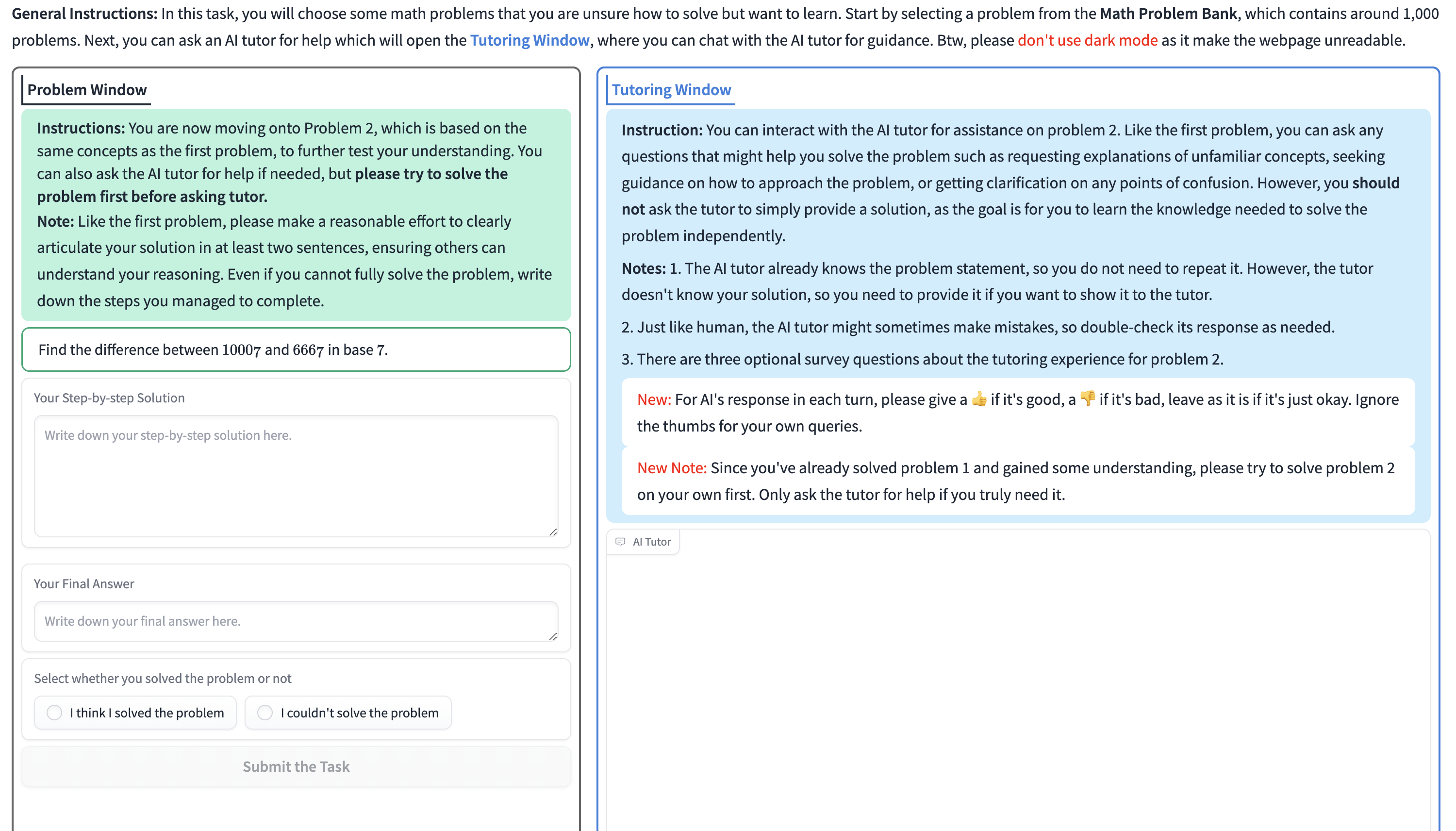}
    \caption{Step 3 of the math tutoring annotation interface: if the user feels that they solve the problem, they move onto the second problem that has similar concepts.}
    \label{fig:math_tutoring_interface_step_3}
\end{figure*}

\begin{figure*}[t]
    \centering
    \includegraphics[width=\textwidth]{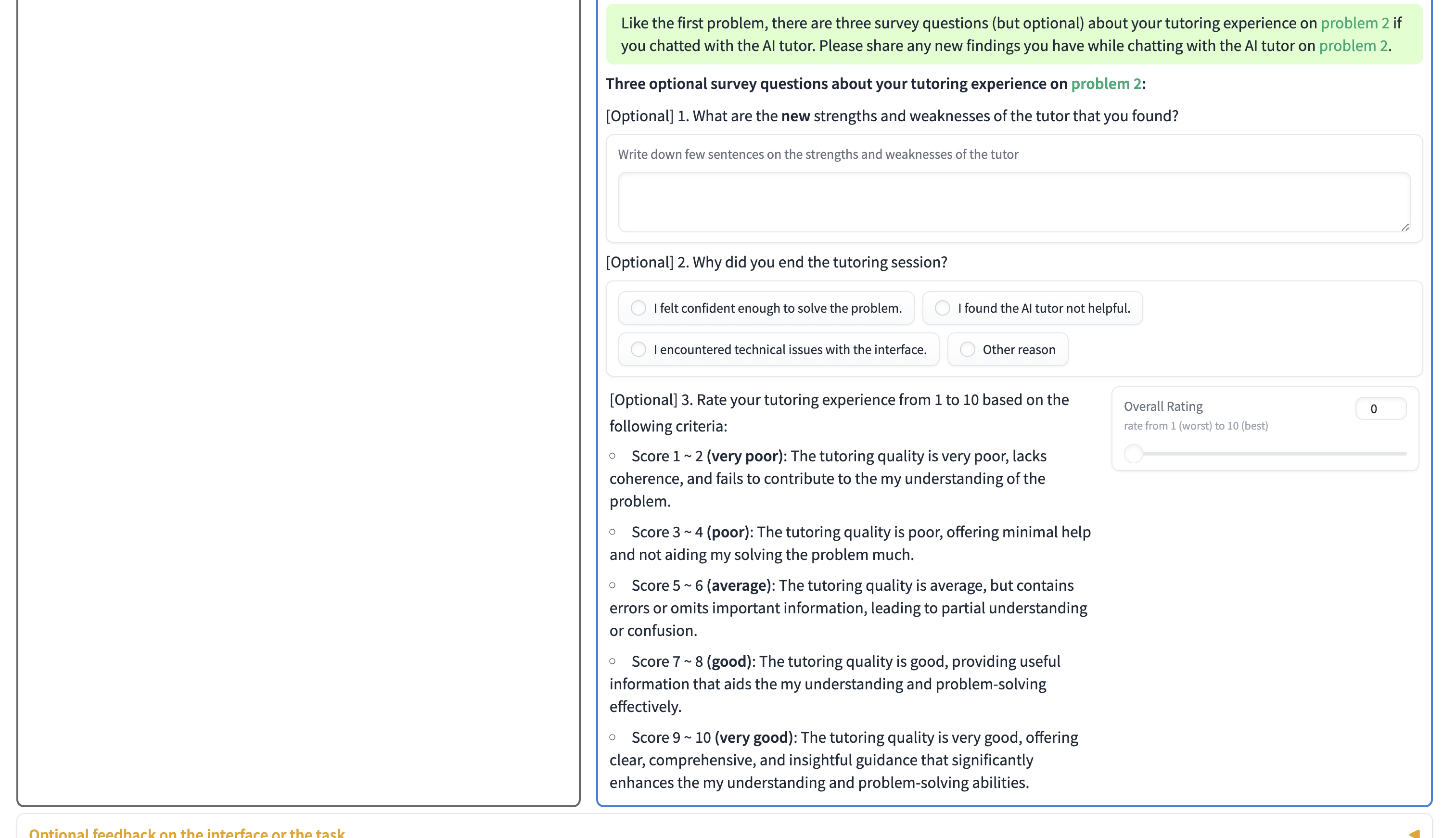}
    \caption{Continued step 3, the user can optionally to give annotation to tutor's performance on the second problem.}
    \label{fig:math_tutoring_interface_step_3_1}
\end{figure*}

\begin{figure*}[t]
    \centering
    \includegraphics[width=\textwidth]{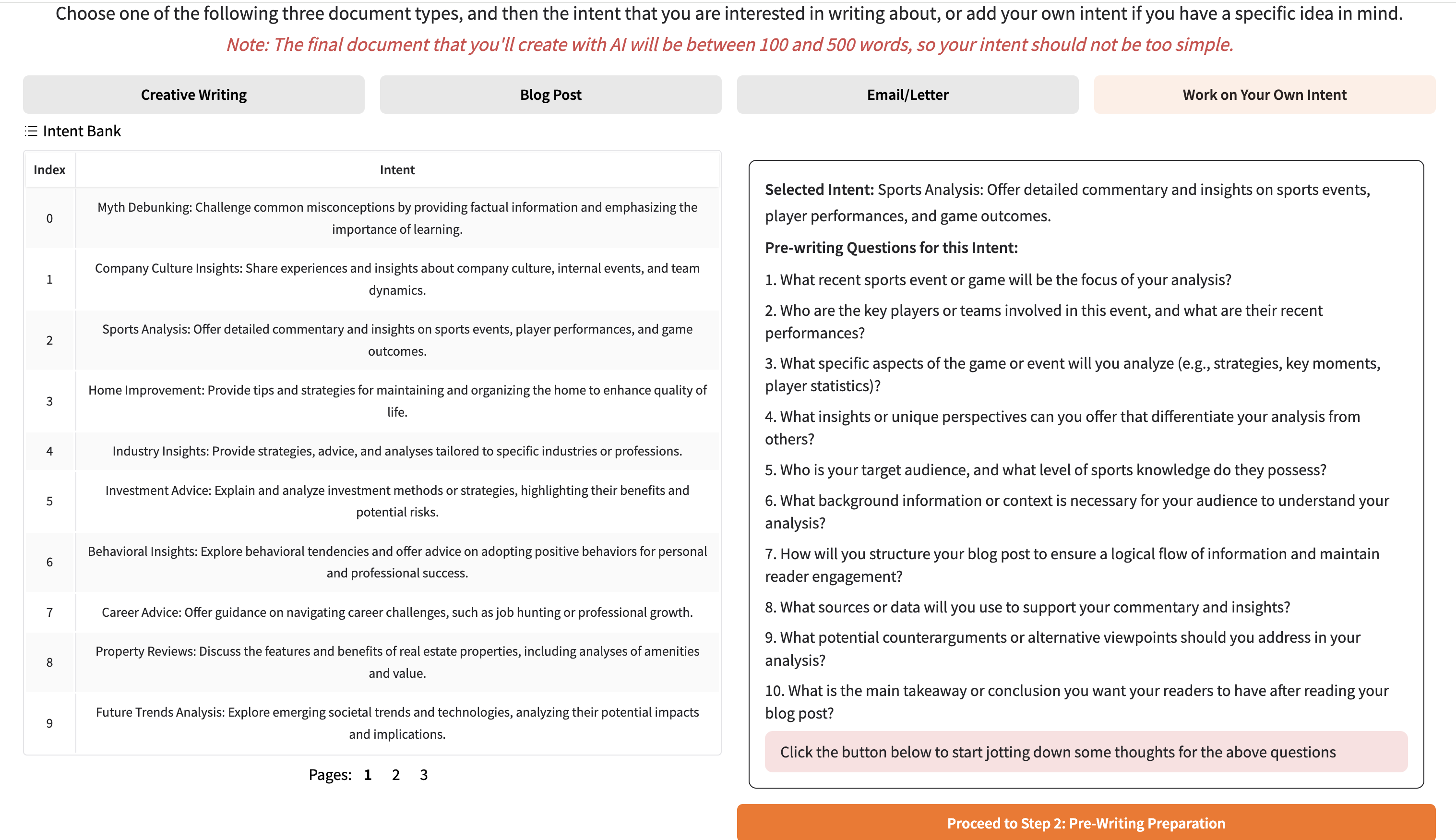}
    \caption{Step 1 of the document creation annotation interface: selecting interested document intent or work on their own intent.}
    \label{fig:document_creation_interface_step_1}
\end{figure*}

\begin{figure*}[t]
    \centering
    \includegraphics[width=\textwidth]{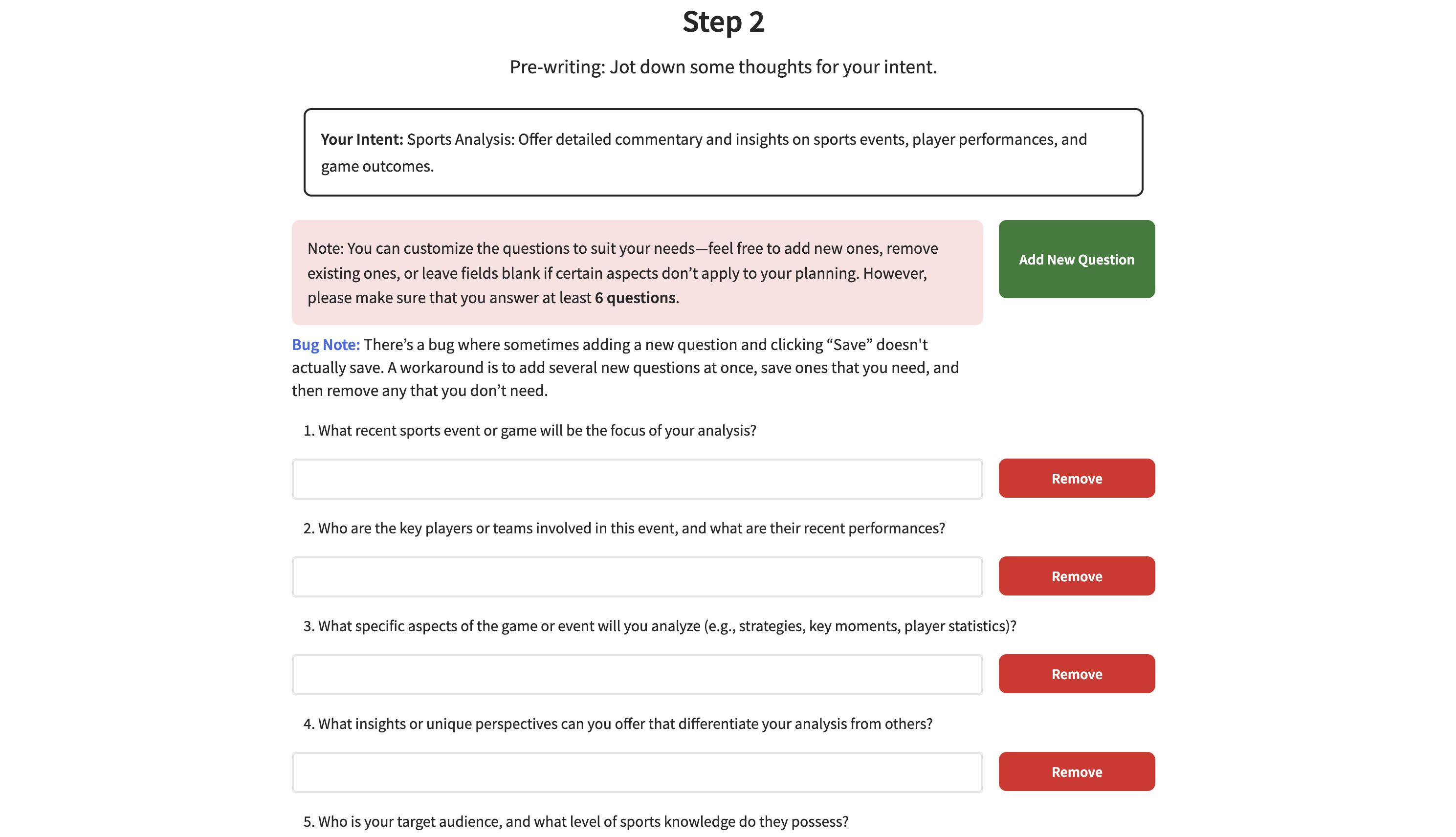}
    \caption{Step 2 of the document creation annotation interface: jot down some thoughts for the document intent the user chooses.}
    \label{fig:document_creation_interface_step_2}
\end{figure*}

\begin{figure*}[t]
    \centering
    \includegraphics[width=\textwidth]{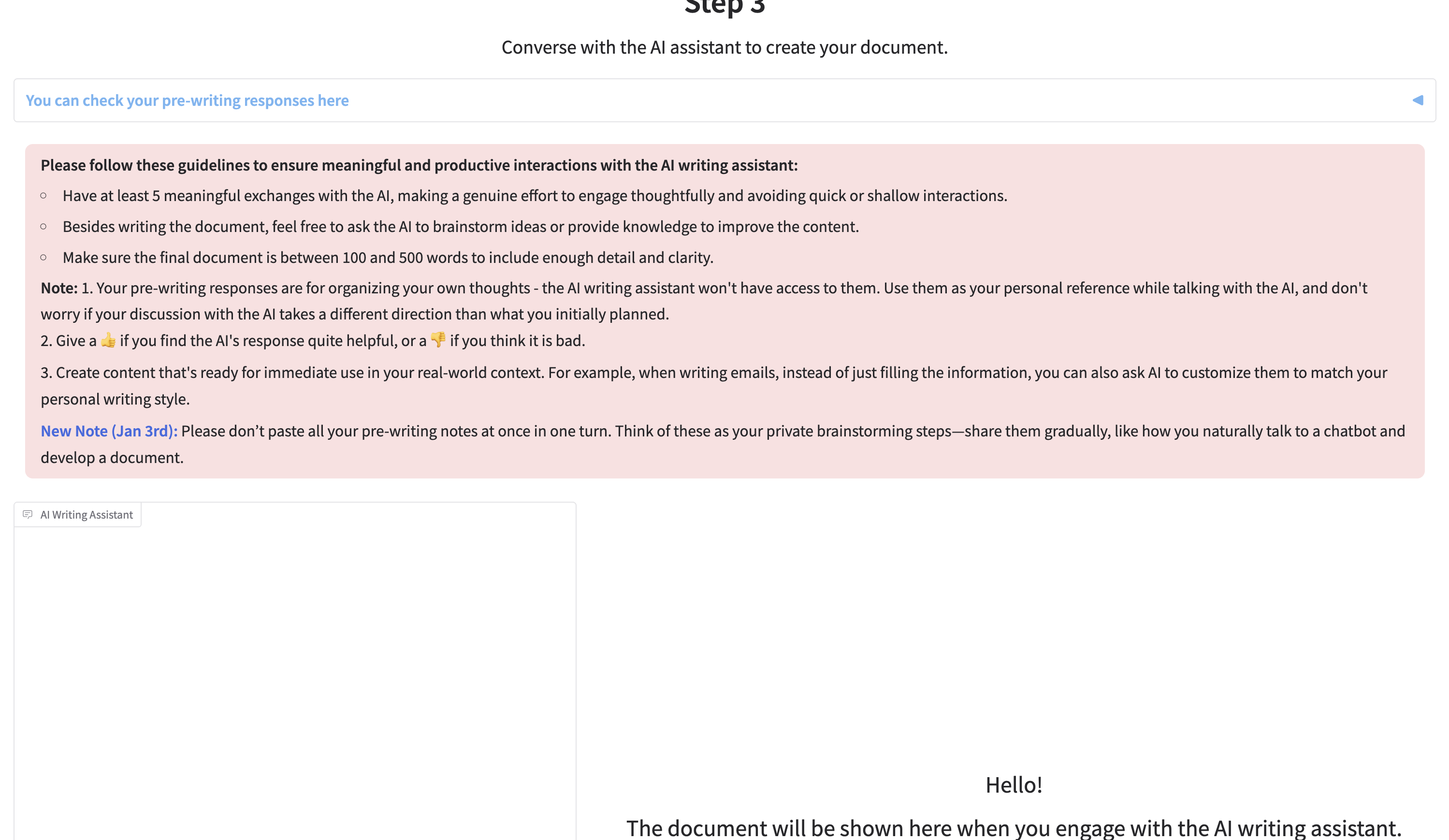}
    \caption{Step 3 of the document creation annotation interface: converse with the AI writing assistant.}
    \label{fig:document_creation_interface_step_3}
\end{figure*}

\begin{figure*}[t]
    \centering
    \includegraphics[width=\textwidth]{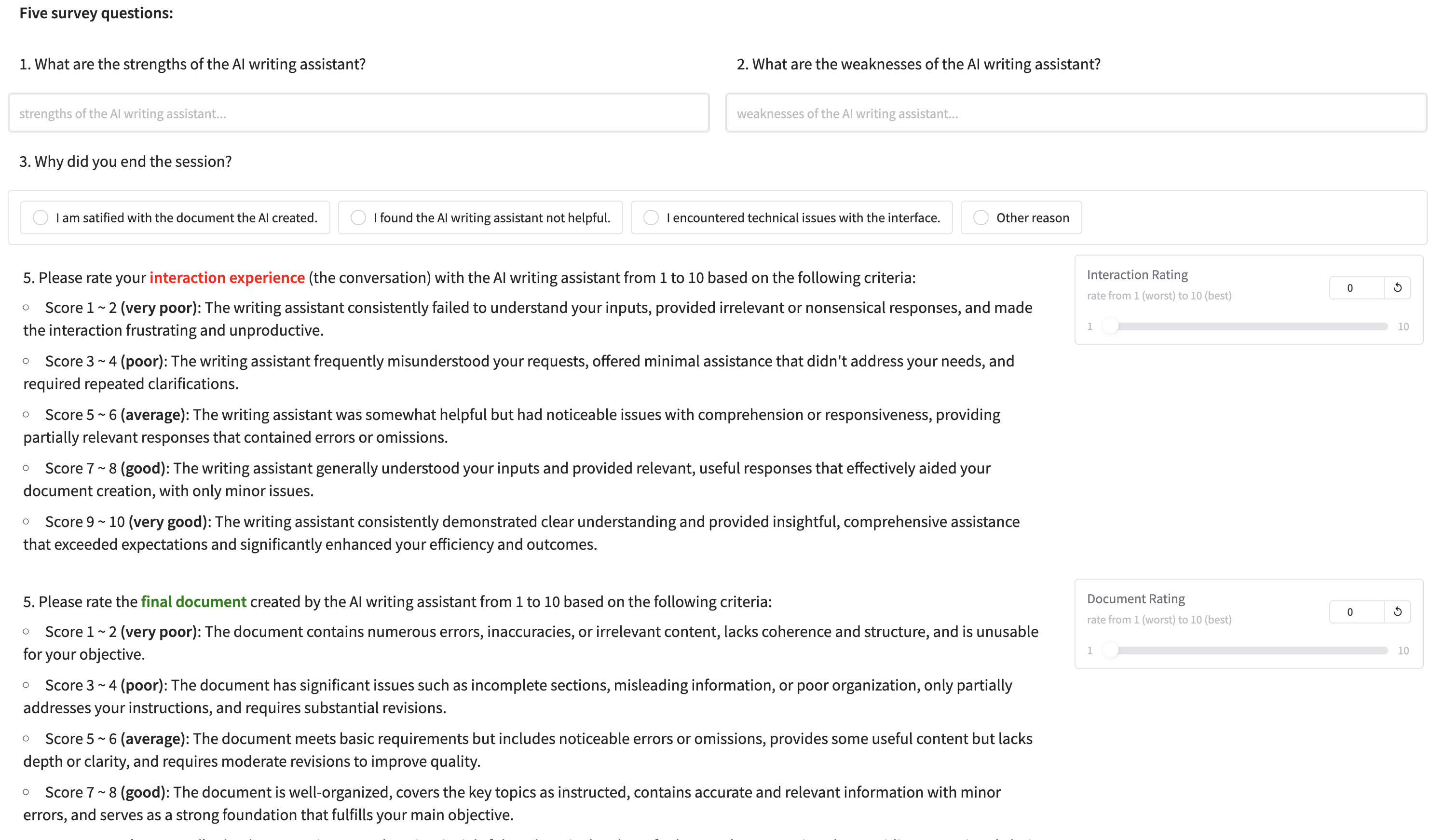}
    \caption{Continued step 3, annotate the performance of the AI writing assistant.}
    \label{fig:document_creation_interface_step_3_1}
\end{figure*}

\end{document}